\documentclass[letterpaper, 10 pt, conference]{ieeeconf}  

\IEEEoverridecommandlockouts                              

\usepackage{graphicx}
\usepackage{cuted}
\usepackage{capt-of}
\usepackage{booktabs}
\usepackage{amssymb}
\usepackage{amsmath}
\usepackage{array}
\usepackage{multirow}
\usepackage{booktabs}
\usepackage{xurl}
\usepackage{booktabs}
\usepackage{array}
\usepackage{amsmath}
\usepackage{amssymb}
\usepackage{dblfloatfix}

\usepackage{marvosym}
\usepackage{longtable}

\title{\LARGE \bf
DexJoCo: A Benchmark and Toolkit for Task-Oriented Dexterous Manipulation on MuJoCo
}

\author{
Hanwen Wang*\(^{1}\), Weizhi Zhao*\(^{1}\), Xiangyu Wang*\(^{1}\), Siyuan Huang*\(^{2}\), \\
He Lin\(^{1}\), Boyuan Zheng\(^{1}\), Rongtao Xu\(^{3}\), Gang Wang\(^{4}\), Yao Mu\(^{2}\), He Wang\(^{5}\), \\
Lue Fan\(^{1,\dagger,\textsuperscript{\Letter}}\), Hongsheng Li\(^{6}\), Zhaoxiang Zhang\(^{1,\textsuperscript{\Letter}}\), Tieniu Tan\(^{1}\) \\[0.3em]
\(^{1}\)NLPR \& MAIS, CASIA, \(^{2}\)SJTU, \(^{3}\)MBZUAI, \\
\(^{4}\)Beijing Institute of Basic Medical Sciences, \(^{5}\)PKU \& Galbot, \(^{6}\)CUHK
\vspace*{-6mm}
\thanks{* Equal contribution. \(\dagger\) Project lead. 
\textsuperscript{\Letter}~Corresponding authors: \{lue.fan, zhaoxiang.zhang\}@casia.ac.cn}
}

\makeatletter
\def\@makecaption#1#2{%
  \vskip\abovecaptionskip
  \setbox\@tempboxa\hbox{{\normalfont\footnotesize #1: #2}}%
  \ifdim \wd\@tempboxa >\hsize
    {\normalfont\footnotesize #1: #2\par}
  \else
    \hbox to\hsize{\hfil\box\@tempboxa\hfil}
  \fi
  \vskip\belowcaptionskip}
\makeatother

\begin{document}

\maketitle
\thispagestyle{empty}
\pagestyle{empty}


\begin{strip}
	\centering
	\includegraphics[width=0.98\textwidth]{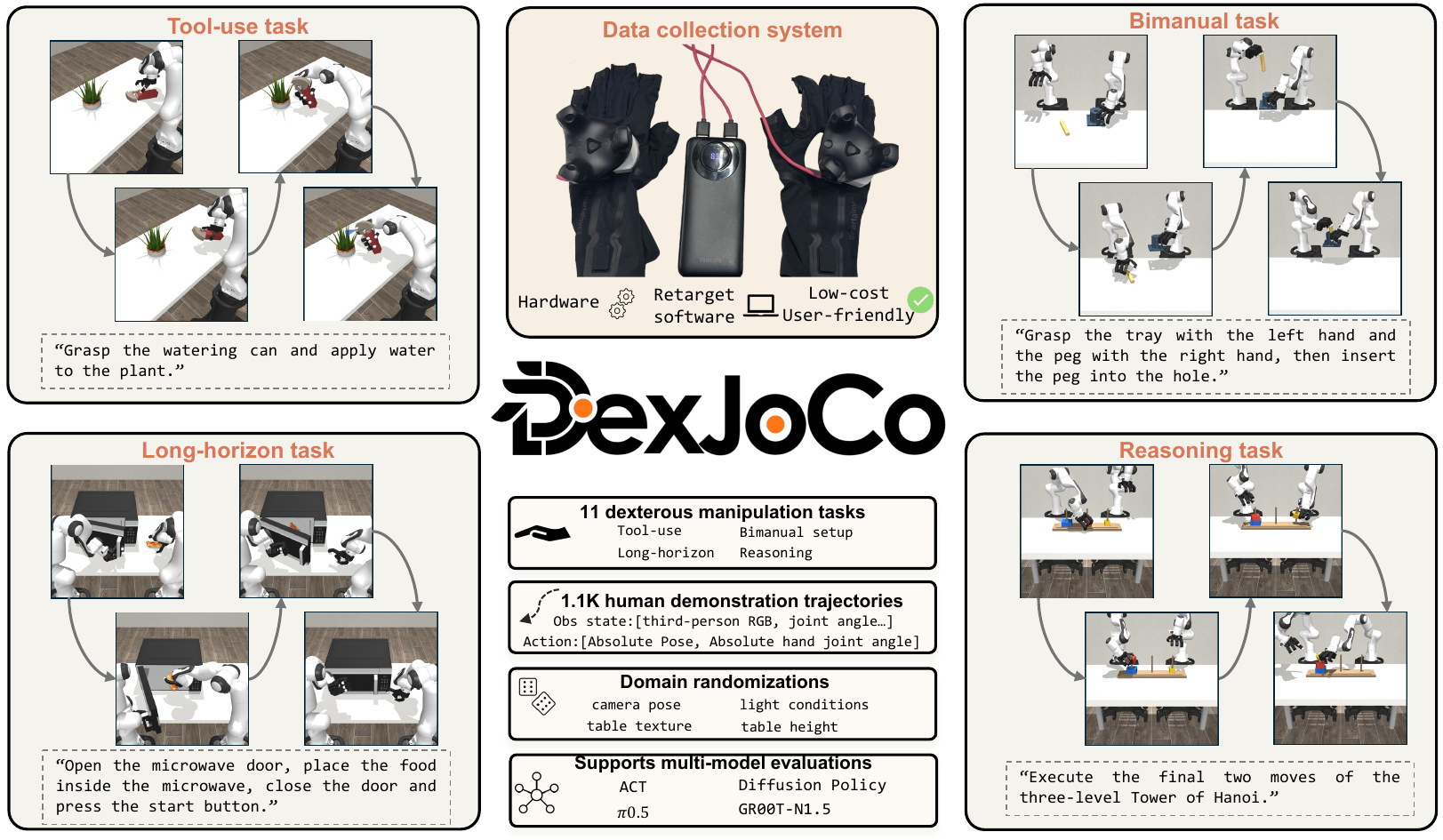}
	\captionof{figure}{
		\textbf{Overview of DexJoCo.} DexJoCo is a dexterous manipulation benchmark with a toolkit for data collection and policy evaluation, covering tool-use, bimanual coordination, long-horizon execution, and reasoning. It includes 11 tasks, 1.1K human demonstration trajectories, and supports trajectory replay under domain randomization for robustness evaluation.
	}
	\label{fig:overview}
	\vspace*{3mm}
\end{strip}

\begin{abstract}

	Achieving human-level manipulation requires dexterous robotic hands capable of complex object interactions. Advancing such capabilities further demands standardized benchmarks for systematic evaluation. However, prior work does not focus on task-oriented dexterous manipulation. Existing benchmarks also often lack high quality human demonstrations and an easy-to-use evaluation pipeline. In this paper, we present DexJoCo, a benchmark and toolkit for task-oriented dexterous manipulation, comprising 11 functionally grounded tasks that evaluate tool-use, bimanual coordination, long-horizon execution, and reasoning. We develop a low-cost data collection system and collect 1.1K trajectories across these tasks, with support for domain randomization to assess robustness. We benchmark modern models under diverse settings, including visual and dynamics randomization, multi-task training, and action-head adaptation. Through extensive empirical analysis, we identify several important insights and common limitations of current policies in dexterous manipulation, highlighting key challenges for future research in dexterous hand robot learning. Code is available at: \url{https://dexjoco.github.io}

\end{abstract}


\section{INTRODUCTION}

Learning from human demonstrations is an effective pathway toward generalist robot manipulation. In recent years, the robotics community has developed low-cost data collection pipelines~\cite{fu2024mobile,chi2024universal} and introduced a wide range of foundation models based on the VLA architecture~\cite{zitkovich2023rt,o2024open,kim2025openvla,black2025pi_,gr00tn1_2025}. However, most existing systems and datasets primarily focus on manipulator-gripper platforms. Human-level manipulation requires dexterous hands capable of fine-grained and contact-rich interactions, making dexterous manipulation learning increasingly important~\cite{shaw2023leaphand,christoph2025orca,romero2024eyesight,zheng2026egoscale,heng2025vitacformerlearningcrossmodalrepresentation}. Advancing dexterous manipulation learning also requires standardized evaluation benchmarks to systematically measure model capabilities and guide future research.

\begin{table*}[t]
	\centering
	\footnotesize
	\setlength{\tabcolsep}{8pt}
	\renewcommand{\arraystretch}{1.12}

	\begin{tabular}{lccccc}
		\toprule
		Benchmark & Hand Tool-Use       & Bimanual       & Reasoning      & Hand MoCap System & Trajectory Collection Method \\
		\midrule

		Adroit~\cite{Rajeswaran-RSS-18}
		          & \(\checkmark\)      &                &                & \(\checkmark\)    & Few Human + RL Policy        \\

		CALVIN~\cite{mees2022calvin}
		          &                     &                & \(\checkmark\) &                   & Motion Planning              \\

		DexArt~\cite{Bao_2023_CVPR}
		          & \(\checkmark\)      &                &                &                   & RL Policy                    \\

		LIBERO~\cite{liu2023libero}
		          &                     &                &                &                   & Human Demonstration          \\

		RoboTwin 2.0~\cite{Mu_2025_CVPR}
		          &                     & \(\checkmark\) &                &                   & Motion Planning              \\

		DexMimicGen~\cite{jiang2025dexmimicgen}
		          &                     & \(\checkmark\) &                & \(\checkmark\)    & Few Human + MimicGen         \\

		Bi-DexHands~\cite{chen2022towards}
		          & \(\checkmark\)      & \(\checkmark\) &                &                   & RL Policy                    \\

		\midrule

		DexJoCo (ours)
		          & \(\checkmark\)      & \(\checkmark\) & \(\checkmark\) & \(\checkmark\)
		          & Human Demonstration                                                                                      \\

		\bottomrule
	\end{tabular}

	\caption{\textbf{Comparison with existing manipulation benchmarks.} DexJoCo features more comprehensive evaluation task categories that highlight the capabilities of dexterous hands, together with an easy-to-use infrastructure for hand motion based data collection.}
	\label{tab:comparison}
\end{table*}

Due to differences in environmental setups and robot configurations across laboratories, evaluating dexterous manipulation algorithms requires a benchmark. Although evaluation benchmarks for manipulator-gripper robotic systems have become relatively mature, and several benchmark efforts have also been introduced for dexterous hand manipulation, existing approaches still suffer from the following limitations: (1) Many existing works omit the manipulator and consider hand-only setups to enlarge the effective workspace, resulting in benchmark trajectories that are difficult to realize in real-world scenarios. (2) Current benchmarks evaluate in-hand manipulation or simple pick-and-place tasks; however, in-hand manipulation tasks are limited in functional diversity, while simple pick-and-place tasks have been extensively studied in prior work. (3) Existing works lack reliable and user-friendly systems for collecting high-quality dexterous manipulation trajectories. Since complex dexterous hand behaviors are difficult to generate using conventional motion planning, most existing works rely on reinforcement learning or automated generation pipelines to obtain trajectories, which often produce behaviors that are inconsistent with natural human manipulation patterns. (4) Existing dexterous manipulation benchmarks lack standardized language instructions and unified data formats for modern VLA models, making systematic training and evaluation difficult.

The robot learning community still lacks a standardized benchmark for dexterous hand manipulation, highlighting the need for an evaluation framework. Therefore, we present DexJoCo, a benchmark and toolkit for task-oriented dexterous manipulation, with comparisons to existing manipulation benchmarks summarized in Table~\ref{tab:comparison}. In designing the tasks, we emphasize functionally grounded interactions that highlight the capabilities of dexterous hands, particularly in tool-use scenarios that require fine-grained finger coordination and complex object interactions. Furthermore, we introduce long-horizon tasks, bimanual coordination tasks, and reasoning tasks to evaluate policy performance across multiple dimensions. A comprehensive evaluation framework requires not only diverse and functionally meaningful task definitions, but also an efficient system for collecting manipulation trajectories. To this end, we develop a low-cost teleoperation hardware setup together with a retargeting module that reduces the embodiment gap between human hand motions and dexterous hand control. Using this system, we collect demonstration data across our task suite and evaluate several modern manipulation policies, leading to several insights into the limitations and challenges of current dexterous manipulation policies, which may facilitate future progress in robot learning. Our contributions are summarized as follows:

(1) \textbf{DexJoCo benchmark}: We introduce a dexterous manipulation benchmark featuring functionally grounded tasks that evaluate the capabilities of dexterous hands, including fine-grained manipulation, tool-use, bimanual coordination, long-horizon execution, and reasoning capabilities.

(2) \textbf{DexJoCo toolkit}: We develop a low-cost teleoperation system with a retargeting module for efficient collection of dexterous manipulation demonstrations.

(3) \textbf{DexJoCo datasets}: We collect 1.1K human demonstration trajectories in simulation and evaluate several modern policies, where dexterous hand trajectory data remains relatively limited in prior work.

\section{Related Works}

\begin{figure*}[t]
	\centering
	\includegraphics[width=\textwidth]{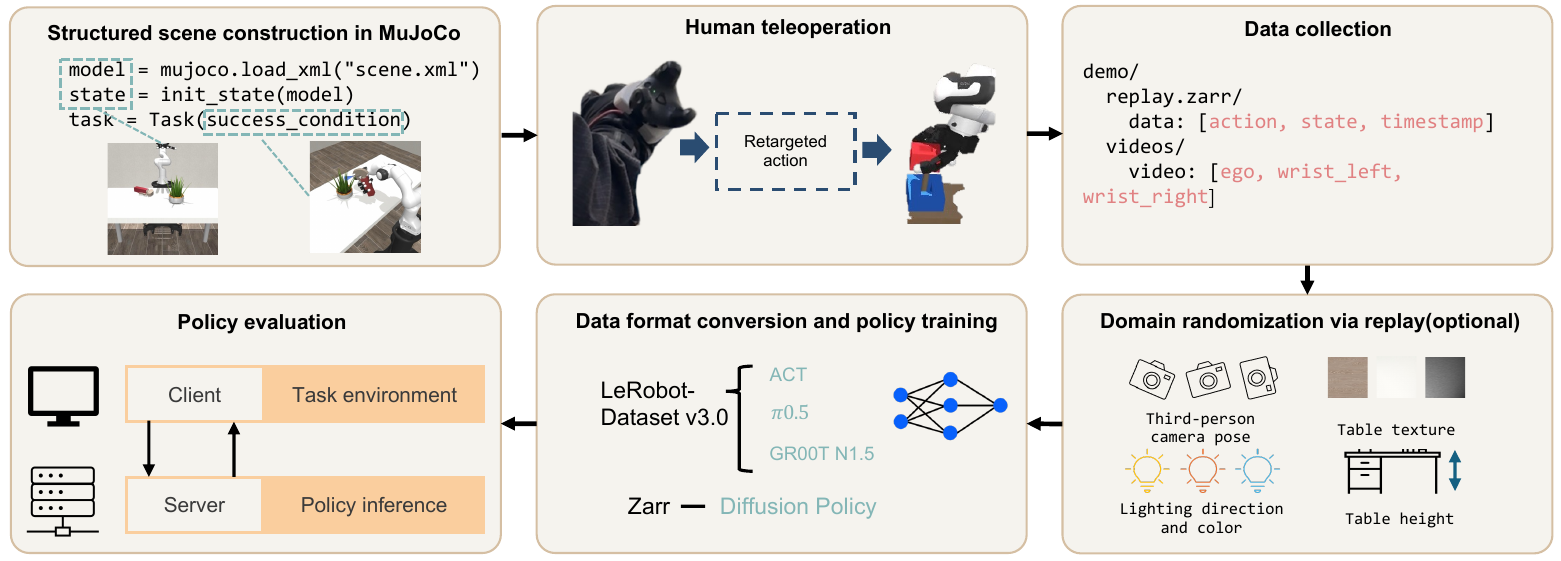}
	\caption{\textbf{DexJoCo pipeline.} 3D assets are first imported into MuJoCo, where structured success conditions are defined based on object poses, articulated joint states, contact conditions, and temporal constraints. Human demonstrations are collected through the teleoperation system, with actions directly recorded as robot position control commands. Replay-based visual augmentation can optionally be applied to the collected trajectories. The data can then be converted into mainstream formats such as LeRobot and DP Zarr through the provided interface. After training, policies are evaluated in the constructed task environments using a server--client framework.}
	\label{fig_pipe}
	\vspace{-3mm}
\end{figure*}

\textbf{Dexterous Manipulation Benchmark.} When designing benchmarks for manipulator–gripper robotic systems, the relatively low degrees of freedom of these robots make it possible to collect large amounts of trajectory data at low cost or through automated procedures~\cite{gu2023maniskill2,liu2023libero,mclean2025metaworld,mandlekar2023mimicgen,mees2022calvin,Mu_2025_CVPR,mu2maniskill,robocasa2024,robocasa365,robomimic2021,taomaniskill3}. However, achieving human-level manipulation requires dedicated benchmarks for manipulator–hand robotic systems. Several existing dexterous hand benchmarks~\cite{chen2022towards,xu2023unidexgrasp,wan2023unidexgrasp++,Rajeswaran-RSS-18,Bao_2023_CVPR,NEURIPS2023_8a84a434} are primarily designed for reinforcement learning and mainly focus on in-hand manipulation. While effective for evaluating low-level dexterous control, their task formulations often provide limited coverage of functional, task-oriented interactions with the environment. Moreover, without access to high-quality human demonstrations, reinforcement learning alone often struggles to generate reasonable and physically plausible manipulation trajectories. Some recent works have adopted human demonstrations or automatically generated trajectories to enable imitation learning for dexterous hand systems~\cite{zhudexflywheel,jiang2025dexmimicgen,luo2025human}. Nevertheless, the resulting task designs are often not sufficiently challenging or functionally rich to assess human-level dexterous manipulation.

\textbf{Dexterous Hand Trajectory Collection.} The technical pipeline for collecting trajectories on manipulator–gripper robotic systems has become increasingly mature~\cite{wu2024gello, liu2025factr, chi2024universal,fu2024mobile}. In practice, action recording only requires tracking the target 6D pose of the robot end-effector, while the gripper itself typically has only a single degree of freedom, eliminating the need for specialized hardware. Trajectory collection for dexterous hand systems is considerably more challenging due to their high degrees of freedom. In practice, specialized hardware is often required to capture the pose of each fingertip and retarget it to the robotic hand. Standard RGB camera-based solutions offer the lowest hardware cost~\cite{qin2023anyteleop,shaw2024learning}, but they frequently suffer from severe occlusion and inefficient hand pose estimation. VR headset-based systems can improve the efficiency of hand pose tracking~\cite{ding2025bunny,iyer2024open,zhang2026unidex}, yet they are often uncomfortable for prolonged use and still remain susceptible to partial occlusion. In contrast, motion-capture gloves or exoskeleton devices can largely eliminate occlusion issues and avoid the need for dedicated vision-based hand pose estimation algorithms~\cite{zhang2025doglove,yin2025geometric,xu2025dexumi,wen2025gr,wang2024dexcap,gao2025glovity,fang2025dexop,feng2025learning}, enabling the direct acquisition of high-frequency and high-precision hand motion data. Their main drawbacks, however, are the relatively high hardware cost, and in the case of exoskeletons, limited wearing comfort. Therefore, we aim to design a data collection system based on motion-capture gloves, together with an effective retargeting algorithm, to achieve both low cost and ease of use.

\section{DexJoCo Benchmark and Toolkit}

DexJoCo provides a benchmark and toolkit for dexterous manipulation, including task environments, human demonstration collection tools, policy training interfaces, and evaluation utilities. Fig.~\ref{fig_pipe} illustrates the overall DexJoCo pipeline, from task construction and trajectory collection to policy training and evaluation. In this section, we describe the robot setup and observation state, teleoperation system, task design, domain randomization settings, and policy evaluation.

\subsection{Robot Setup and Observation State}
\label{sec:robot_setup}
DexJoCo is developed on top of the MuJoCo physics simulator. The robotic system consists of three main components: a Rethink Robotics mount as the base, a Franka Panda manipulator, and an Allegro Hand for dexterous manipulation. These assets are mature, precisely modeled, and widely adopted in the robotics community. DexJoCo provides rich perceptual observations from the simulation environment, including third-person and wrist-mounted RGB and RGB-D images, object poses of the interactive entities in the scene, the robot’s motion states, the current end-effector pose, and the joint angles of the hand. The action space in the collected robot trajectories is defined as follows: manipulator actions are represented by the target absolute end-effector pose in the world coordinate frame, while hand actions are specified as target absolute joint angles.

\subsection{Human Demonstration Data Collection System}
\begin{figure}[t]
	\centering
	\includegraphics[width=\linewidth]{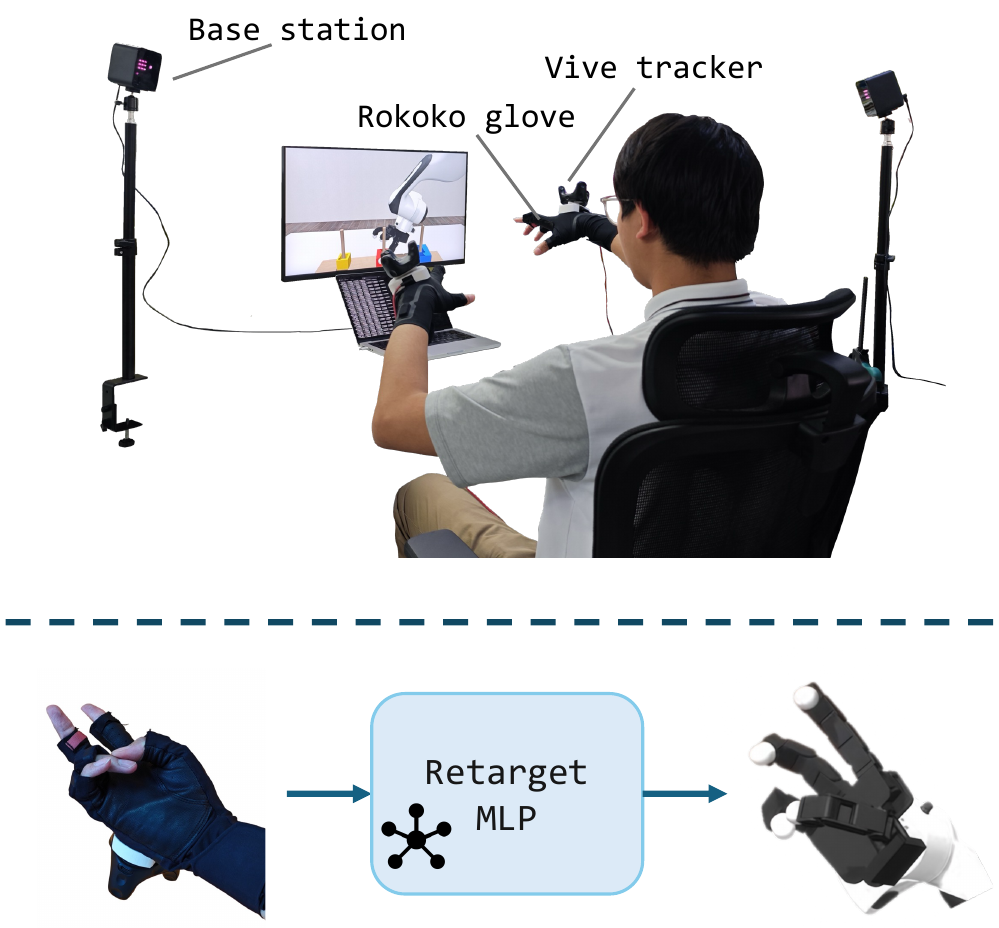}
	\caption{\textbf{Human demonstration data collection system.} The top figure shows the overall teleoperation system. A Rokoko glove is used to capture hand poses, while an HTC Vive tracker is employed to track the wrist pose. The bottom figure shows that a retargeting mapping is trained to convert human fingertip poses into joint configurations of the Allegro hand.}
	\label{fig:data_collection_system}
	\vspace{-2mm} 
\end{figure}

\textbf{Hardware Design.} The hardware system in DexJoCo is designed to balance low cost and usability, as shown in Fig.~\ref{fig:data_collection_system}. Hand motion capture is performed using \textbf{Rokoko Smartgloves}, avoiding the occlusion issues of camera-based methods, while two \textbf{HTC Vive Trackers} and two \textbf{HTC Base Stations} are used to track wrist motions and control the Franka end-effector pose. This setup enables accurate teleoperated trajectory collection and remains low-cost at approximately \$2,300 USD. A simple 3D-printed connector is further designed to integrate the trackers and gloves into a unified assembly.

\textbf{Teleoperation Algorithm.} The teleoperation system consists of hand motion retargeting and wrist motion tracking. Due to the structural differences between human and robotic hands, direct linear mapping is infeasible. We adopt \textbf{GeoRT}~\cite{yin2025geometric}, a lightweight self-supervised retargeting method without requiring paired human-robot annotations. The retargeting model \(f\) maps human fingertip keypoints \(x_H\) to robot joint positions \(q_R=f(x_H)\) by minimizing:
\begin{equation}
	\mathcal{L} = \mathcal{L}_{\text{dir}} + \lambda_1 \mathcal{L}_{\text{cover}} + \lambda_2 \mathcal{L}_{\text{flat}} + \lambda_3 \mathcal{L}_{\text{pinch}} + \lambda_4 \mathcal{L}_{\text{col}}
\end{equation}
where \(\mathcal{L}_{\text{dir}}\) preserves fingertip motion directions, \(\mathcal{L}_{\text{cover}}\) enlarges workspace coverage, \(\mathcal{L}_{\text{flat}}\) maintains uniform sensitivity, \(\mathcal{L}_{\text{pinch}}\) preserves pinch behaviors, and \(\mathcal{L}_{\text{col}}\) avoids self-collisions. Only fingertip workspaces are recorded during data collection and used for training, enabling accurate real-time teleoperation. For wrist tracking, the tracker is fixed such that human wrist motions align with the Franka end-effector. The initial wrist pose is recorded as a reference, and subsequent actions are represented as relative pose changes. The robot then executes these delta actions to reproduce the desired motion.

\subsection{Task Design in the Benchmark}

\begin{figure*}[t]
	\centering
	\includegraphics[width=\linewidth]{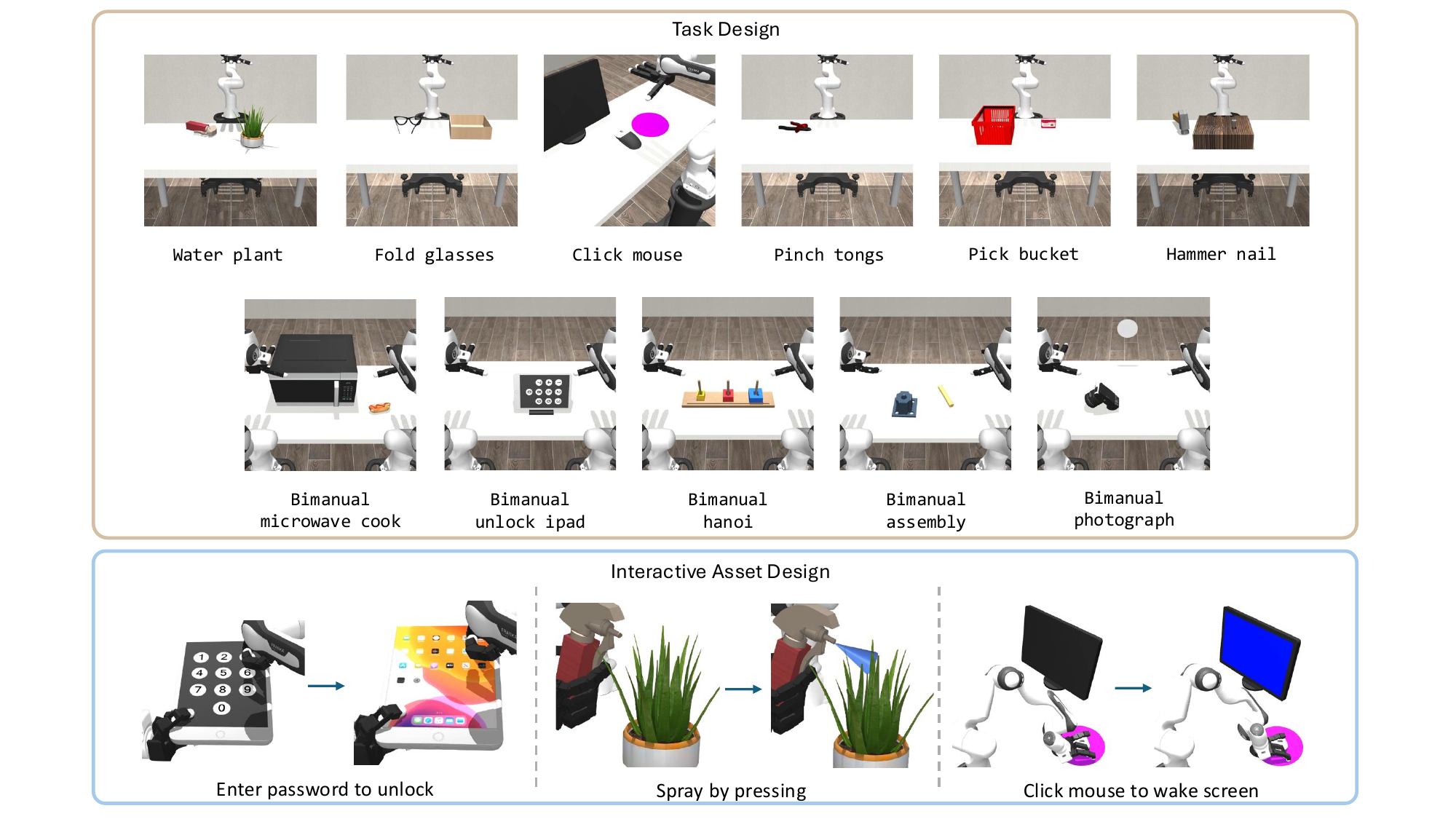}
	\caption{\textbf{Task design in DexJoCo.} The top panel illustrates the task environment design, showing the initial state of each task. The bottom panel presents the visual and interactive properties of the task assets.}
	\label{task_design}
\end{figure*}

\textbf{Formulation.} Each task in DexJoCo is defined by a set of interactive objects and task goals: \(\mathcal{T}=(\mathcal{O},\mathcal{G})\), where \(\mathcal{O}=\{o_1,o_2,\dots,o_m\}\) denotes the set of interactive objects in the scene. The task goal is formulated as a set of functional success constraints \(\mathcal{G}=\{g_{\text{seq}},g_{\text{pose}},g_{\text{joint}},g_{\text{contact}}\}\), where \(g_{\text{seq}}\) denotes temporal or sequential execution constraints, \(g_{\text{pose}}\) specifies target object pose conditions, \(g_{\text{joint}}\) represents articulated joint-state requirements, and \(g_{\text{contact}}\) defines collision. A task is considered successful only when all task-dependent goal constraints are satisfied simultaneously.

\textbf{Task Design Principles.} DexJoCo tasks are systematically constructed to cover diverse dexterous manipulation capabilities, as shown in Fig.~\ref{task_design}. We follow several core design principles. \textbf{(1) Functional interaction:} Tasks are designed with functional semantics that reflect everyday human activities rather than simple object relocation. Moreover, the involved objects provide explicit visual interaction feedback, enabling intuitive perception of task progress and completion. \textbf{(2) Dexterity dependency:} Tasks are designed such that successful execution fundamentally depends on dexterous manipulation capabilities, including fine-grained finger coordination and articulated object interaction. \textbf{(3) Long-horizon compositionality:} Tasks involve multi-stage execution with temporal dependencies between sub-goals. \textbf{(4) Bimanual coordination:} A subset of tasks requires coordinated bimanual manipulation with asymmetric functional roles between the two hands. Based on these principles, tasks are organized into capability-oriented categories, including tool-use tasks, reasoning tasks, bimanual coordination tasks, and long-horizon tasks, ensuring broad and structured benchmark coverage. The construction cost of each individual task is relatively low, enabling efficient and scalable benchmark expansion.

\textbf{Task Asset Construction.} The base scene design follows RoboSuite~\cite{robosuite2020}, and we adopt robot assets from MuJoCo Menagerie~\cite{menagerie2022github}. New tasks are constructed by instantiating task-specific objects within the base scene and defining corresponding success conditions. For each task, we curate high-quality assets from RoboCasa~\cite{robocasa2024} and PartNet-Mobility from SAPIEN~\cite{Xiang_2020_SAPIEN}, which typically provide predefined physical and dynamic parameters. For assets without such annotations, we generate them using Hunyuan3D~\cite{yang2024hunyuan3d} and manually assign physically plausible properties. To enhance functional interaction realism, we additionally incorporate explicit visual state changes into task assets. For example, in the \textit{Water Plant} task, water is displayed when the watering can handle reaches a predefined joint state threshold. In the \textit{iPad Unlock} task, buttons are highlighted upon finger contact. In the \textit{Click Mouse} task, pressing the mouse button activates the computer display, indicating successful interaction.

\subsection{Domain Randomizations}
To evaluate the policies over a broader data distribution, we introduce a domain randomization option for all task scenarios. To generate more diverse trajectories, we not only randomize the placement of objects on the table plane but also vary the table height. To increase visual diversity, we randomize the third-person camera poses, the direction and color of scene illumination, and the tabletop textures. Notably, visual randomization can be efficiently applied by replaying the same trajectories under different rendering settings, enabling scalable augmentation without additional teleoperation effort. For camera pose randomization, we first densely sample camera poses uniformly on a spherical surface, and then select 50 poses with minimal occlusion. For lighting randomization, we follow a simple procedure inspired by our implementation. Each light in the scene is randomized in terms of its position, direction, and diffuse color to introduce diverse illumination conditions. For tabletop texture randomization, we sample textures from a pre-constructed texture library.

\subsection{Imitation Learning Policy Evaluation}
\label{sec:baselines}

\begin{table*}[t]
	\centering
	\setlength{\tabcolsep}{3pt}
	\begin{tabular}{r | r @{\tiny\(\pm\)}>{\tiny}l@{\hspace{3pt}}r@{\tiny\(\pm\)}>{\tiny}l | r@{\tiny\(\pm\)}>{\tiny}l@{\hspace{3pt}}r@{\tiny\(\pm\)}>{\tiny}l | r@{\tiny\(\pm\)}>{\tiny}l@{\hspace{3pt}}r@{\tiny\(\pm\)}>{\tiny}l | r@{\tiny\(\pm\)}>{\tiny}l@{\hspace{3pt}}r@{\tiny\(\pm\)}>{\tiny}l | r@{\tiny\(\pm\)}>{\tiny}l@{\hspace{3pt}}r@{\tiny\(\pm\)}>{\tiny}l}
		\toprule
		\multicolumn{1}{c|}{\multirow{2}{*}{Task}} & \multicolumn{4}{c|}{DP-T}    & \multicolumn{4}{c|}{DP-C}      & \multicolumn{4}{c|}{ACT}     & \multicolumn{4}{c|}{\(\pi_{0.5}\)} & \multicolumn{4}{c}{GR00T N1.5}                                                                                                                                                                                                                                                                        \\
		                                           & \multicolumn{2}{c}{rand-obj} & \multicolumn{2}{c|}{rand-full} & \multicolumn{2}{c}{rand-obj} & \multicolumn{2}{c|}{rand-full}     & \multicolumn{2}{c}{rand-obj}   & \multicolumn{2}{c|}{rand-full} & \multicolumn{2}{c}{rand-obj} & \multicolumn{2}{c|}{rand-full} & \multicolumn{2}{c}{rand-obj} & \multicolumn{2}{c}{rand-full}                                                                                                        \\
		\midrule
		Hammer Nail                                & 81.3                         & 3.1                            & 18.7                         & 1.2                                & 58.7                           & 4.2                            & 19.3                         & 3.1                            & 50.0                         & 7.2                           & 22.7 & 6.1 & \textbf{84.7} & 5.0  & 17.3          & 5.0 & 67.3          & 4.2  & \textbf{38.7} & 8.3 \\
		Click Mouse                                & 62.0                         & 2.0                            & 25.3                         & 8.1                                & 74.0                           & 5.3                            & 34.7                         & 4.2                            & 61.3                         & 3.1                           & 48.7 & 5.0 & 64.7          & 8.1  & 54.7          & 7.0 & \textbf{85.3} & 3.1  & \textbf{74.0} & 2.0 \\
		Pick Bucket                                & 83.3                         & 3.1                            & 58.7                         & 15.0                               & 70.0                           & 2.0                            & 68.0                         & 3.5                            & 64.0                         & 4.0                           & 36.0 & 5.3 & \textbf{84.0} & 7.2  & \textbf{78.7} & 6.1 & 72.0          & 6.0  & 69.3          & 6.1 \\
		Pinch Tongs                                & 22.7                         & 5.8                            & 18.7                         & 3.1                                & \textbf{57.3}                  & 6.4                            & \textbf{28.7}                & 11.7                           & 31.3                         & 3.1                           & 23.3 & 7.0 & 24.0          & 6.9  & 18.7          & 1.2 & 12.7          & 2.3  & 5.3           & 2.3 \\
		Fold Glasses                               & 53.3                         & 3.1                            & 11.3                         & 1.2                                & 54.0                           & 15.9                           & 15.3                         & 7.6                            & 47.3                         & 11.0                          & 7.3  & 3.1 & \textbf{72.0} & 3.5  & \textbf{39.3} & 3.1 & 27.3          & 2.3  & 20.7          & 3.1 \\
		Water Plant                                & 84.0                         & 3.5                            & 56.0                         & 8.7                                & 63.3                           & 3.1                            & 54.0                         & 5.3                            & 47.3                         & 4.6                           & 52.7 & 8.1 & \textbf{88.7} & 3.1  & \textbf{75.3} & 6.4 & 72.7          & 1.2  & 66.0          & 5.3 \\
		Unlock iPad /B                             & 8.0                          & 2.0                            & 2.0                          & 2.0                                & \textbf{52.0}                  & 2.0                            & 12.0                         & 3.5                            & 9.3                          & 3.1                           & 0.7  & 1.2 & 36.0          & 14.4 & \textbf{32.7} & 5.0 & 12.7          & 11.0 & 0.0           & 0.0 \\
		Hanoi /B                                   & \textbf{24.7}                & 4.6                            & 0.7                          & 1.2                                & 12.7                           & 3.1                            & 9.3                          & 6.1                            & 6.0                          & 2.0                           & 4.7  & 2.3 & 23.3          & 7.0  & \textbf{33.3} & 2.3 & 0.7           & 1.2  & 0.0           & 0.0 \\
		Assembly /B                                & \textbf{4.7}                 & 3.1                            & 0.0                          & 0.0                                & 3.3                            & 1.2                            & 0.0                          & 0.0                            & 0.0                          & 0.0                           & 0.0  & 0.0 & 2.7           & 2.3  & 0.7           & 1.2 & 0.7           & 1.2  & \textbf{1.3}  & 1.2 \\
		Microwave /B                               & \textbf{73.3}                & 11.6                           & 21.3                         & 4.6                                & 54.0                           & 12.5                           & 62.7                         & 6.4                            & 66.0                         & 2.0                           & 50.0 & 6.9 & 60.0          & 5.3  & \textbf{69.3} & 8.1 & 50.7          & 4.6  & 42.0          & 7.2 \\
		Photograph /B                              & \textbf{56.7}                & 4.6                            & 7.3                          & 1.2                                & 24.0                           & 8.7                            & 8.7                          & 4.2                            & 7.3                          & 1.2                           & 3.3  & 1.2 & 54.7          & 8.1  & \textbf{23.3} & 4.6 & 40.7          & 7.0  & 18.7          & 7.0 \\
		Avg.                                       & 50.4                         & 1.4                            & 20.0                         & 1.4                                & 47.6                           & 2.0                            & 28.4                         & 1.5                            & 35.5                         & 2.0                           & 22.7 & 1.3 & \textbf{54.1} & 1.8  & \textbf{40.3} & 2.9 & 40.2          & 0.3  & 30.5          & 1.1 \\
		\bottomrule
	\end{tabular}
	\vspace{6pt}
	\caption{\textbf{Performance comparison on benchmark tasks.} Mean success rate (\%) \(\pm\) std over 11 tasks for five models. ``/B'': bimanual tasks; ``rand-obj'': only object placement and table height randomized; ``rand-full'': additionally randomizes camera poses, illumination direction/color, and tabletop textures. Each task is trained under both ``rand-obj'' and ``rand-full'' data regimes.}
	\label{tab:benchmark_results}
\end{table*}

\textbf{Baseline Models.} We benchmark four policies on DexJoCo: ACT~\cite{zhao2023aloha}, Diffusion Policy~\cite{chi2023diffusionpolicy} (DP-T and DP-C), \(\pi_{0.5}\)~\cite{black2025pi_}, and GR00T N1.5~\cite{gr00tn1_2025}.
ACT (via C-VAE) and DP (via diffusion) are trained from scratch using vision and proprioception. In contrast, \(\pi_{0.5}\) and GR00T N1.5 (fine-tuned via LoRA~\cite{hu2022lora}) use flow-matching and additionally condition on language.
Because their default 32-dimensional action heads are insufficient for bimanual tasks, we retain these pretrained weights but randomly initialize the extra dimensions (partial pretrain-AH).
All baselines formulate action chunking as:
\begin{equation}
	\mathcal{P}(a_{t:t+k-1}) = \pi_{\theta}(a_{t:t+k-1} \mid s_{t-h+1:t}, l)
\end{equation}
In the formula, given \(h\) frames of historical observations \(s\) and an optional language instruction \(l\), it models the conditional probability of a future \(k\)-step action chunk.

\textbf{Model Deployment.} For evaluation, we use an asynchronous inference mechanism inspired by SmolVLA~\cite{shukor2025smolvla}: the next action chunk is generated while the current one executes, eliminating idle waiting. Overlapping chunks are temporally ensembled for smoothness. This mirrors real-world deployment and highlights the impact of inference frequency: Lighter policies run faster, utilizing more recent observations to reduce idle frames and improve reactivity.

\section{Experiments}

\begin{figure*}[t]
	\centering
	\includegraphics[width=0.9\linewidth]{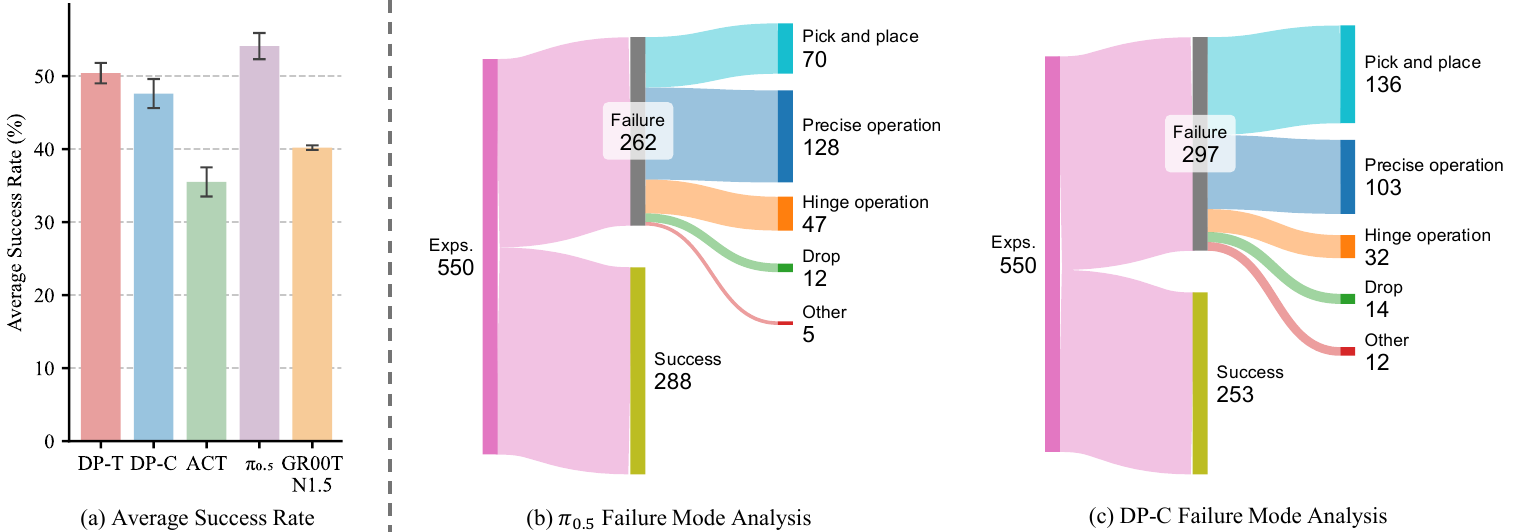}
	\caption{\textbf{Performance evaluation and failure mode analysis.} DP denotes Diffusion Policy, with -T and -C representing Transformer and CNN-based architectures, respectively. \textbf{(a)} Comparison of average success rates across different baselines under the ``rand-obj'' (Table~\ref{tab:benchmark_results}) condition. \textbf{(b)} and \textbf{(c)} provide a detailed breakdown of failure modes for \(\pi_{0.5}\) and DP-C. These statistics are aggregated from 550 evaluation trials (50 runs across 11 tasks) to identify main bottlenecks in dexterous manipulation.}
	\label{fig:sr+sankey}
\end{figure*}

\begin{figure*}[t]
	\centering
	\includegraphics[width=\linewidth]{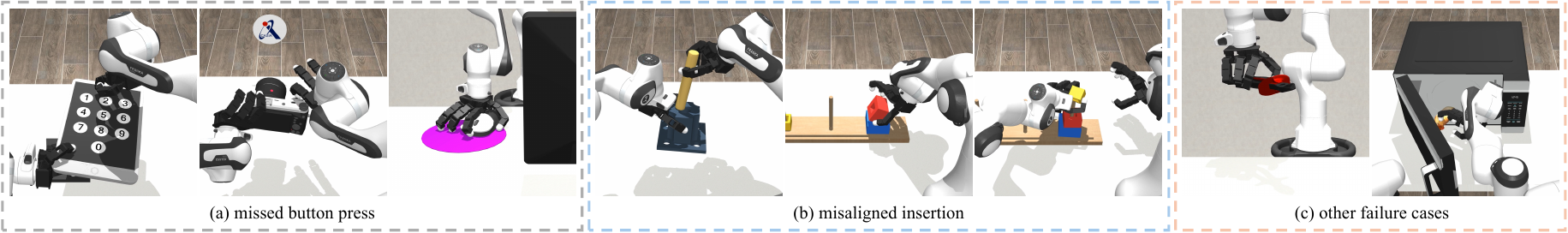}
	\caption{Visualization of failure cases in typical tasks.}
	\label{fig:failure_case}
\end{figure*}

\begin{table}[tb]
	\centering
	\begin{tabular}{lccccc}
		\toprule
		      & ACT & DP-C & DP-T & \(\pi_{0.5}\) & GR00T \\
		\midrule
		1-arm & 7.0 & 37.7 & 33.6 & 85.1          & 80.0  \\
		2-arm & 7.4 & 42.3 & 36.1 & 83.9          & 90.0  \\
		\bottomrule
	\end{tabular}
	\caption{Mean latency on NVIDIA RTX 4090 (ms).}
	\label{tab:latency}
\end{table}

\textbf{Experimental Setup.}
We evaluate the five baselines introduced in Section~\ref{sec:baselines} (DP-T, DP-C, ACT, \(\pi_{0.5}\), and GR00T N1.5). DexJoCo contains 11 tasks with 100 human demonstrations per task (1.1K in total). For each task and randomization setting, each policy is trained once and evaluated with three random seeds. Each seed contains 50 test episodes per task, yielding 1,650 test episodes per setting (\(3 \times 50 \times 11\)). Inputs, action space, and language conditioning follow Section~\ref{sec:robot_setup} and Section~\ref{sec:baselines}. Specifically, the input state consists of end-effector translation (3-D), quaternion rotation (4-D), and dexterous hand joint angles (16-D). The output action comprises translation (3-D), rotation vectors (3-D), and dexterous hand joints (16-D). All values are absolute. Therefore, actions are 22-D for single-arm (\(3 + 3 + 16\)) and 44-D for dual-arm settings, and the control frequency is 30 Hz. All policies are trained on NVIDIA A6000 and L40 GPUs. For deployment, all models are deployed on an NVIDIA RTX 4090 GPU. The mean latency of each model is reported in Table~\ref{tab:latency}. The model inference latency is significantly smaller than the execution time of a single action chunk. The action chunk sizes are DP-T: 45, DP-C: 43, ACT: 100, \(\pi_{0.5}\): 30, and GR00T N1.5: 50. We use the default training hyperparameters of each policy whenever possible; other hyperparameters are provided in the released codebase.

\textbf{DexJoCo Reveals Trade-offs in Pre-training, Scale, and Architecture.} As shown in Table~\ref{tab:benchmark_results} and Fig.~\ref{fig:sr+sankey}, the benchmark proves highly challenging: some policies never succeed on difficult bimanual tasks. For each task, policies are trained on in-domain data under both ``rand-obj'' and ``rand-full'' regimes. Under visual randomization (``rand-full'' in Table~\ref{tab:benchmark_results}), success rates drop sharply across nearly all policies, indicating limited robustness.
\(\pi_{0.5}\) achieves the highest overall success rates, benefiting from large-scale pre-training, yet the much smaller DP-T (\({\sim}100\)M, trained from scratch) performs comparably.
Surprisingly, DP-C substantially outperforms all other policies on \textit{Unlock iPad} and \textit{Pinch Tongs}. The right panel of Fig.~\ref{fig:sr+sankey} reveals that DP-C excels at precise operations (e.g., button pressing) and hinge interactions (e.g., squeezing tongs).
We hypothesize that this advantage stems from being the only policy to use FiLM~\cite{perez2018film} for observation injection, rather than self or cross attention, which may provide stronger fine-grained visual perception and benefit precise manipulation.

\textbf{Failures in Fine-grained Actions, Insertion, and Memory.} As Fig.~\ref{fig:failure_case} shows, in button-based tasks (\textit{Unlock iPad}, \textit{Click Mouse}, \textit{Photograph}), the policies are able to pick up the tablet or camera, push the mouse onto the mousepad, yet often fail to click the intended buttons, suggesting they can perceive the object but overlook its interactive elements.
Insertion steps pose a high probability of failure, as observed in \textit{Assembly} and \textit{Hanoi}.
In \textit{Pinch Tongs}, the policies often grasp but fail to squeeze and release the tongs, possibly due to insufficient temporal memory.
In \textit{Microwave}, the policies typically place the hot dog into the microwave but then withdraw it alongside the hand.

\begin{table}[t]
	\centering
	\scriptsize
	\setlength{\tabcolsep}{3pt}
	\begin{tabular}{r | r @{\tiny\(\pm\)}>{\tiny}l r@{\tiny\(\pm\)}>{\tiny}l | r@{\tiny\(\pm\)}>{\tiny}l r @{\tiny\(\pm\)}>{\tiny}l | r@{\tiny\(\pm\)}>{\tiny}l}
		\toprule
		                          & \multicolumn{4}{c|}{multi-task} & \multicolumn{4}{c|}{rand-dynamics} & \multicolumn{2}{c}{rand-AH}                                                                                                            \\
		\multicolumn{1}{c|}{Task} & \multicolumn{2}{c}{DP-T}        & \multicolumn{2}{c|}{\(\pi_{0.5}\)} & \multicolumn{2}{c}{DP-T}    & \multicolumn{2}{c|}{\(\pi_{0.5}\)} & \multicolumn{2}{c}{\(\pi_{0.5}\)}                                   \\
		\midrule
		Hammer Nail               & 58.7                            & 5.0                                & 74.7                        & 2.3                                & 77.3                              & 6.4 & 82.0 & 12.5 & 76.7 & 1.2  \\
		Click Mouse               & 38.7                            & 3.1                                & 48.0                        & 6.9                                & 0.0                               & 0.0 & 65.3 & 4.2  & 54.0 & 2.0  \\
		Pick Bucket               & 55.3                            & 7.6                                & 60.7                        & 2.3                                & 80.7                              & 3.1 & 90.7 & 1.2  & 86.0 & 5.3  \\
		Pinch Tongs               & 6.0                             & 5.3                                & 44.0                        & 15.9                               & 15.3                              & 4.2 & 17.3 & 7.6  & 26.0 & 2.0  \\
		Fold Glasses              & 11.3                            & 5.0                                & 28.0                        & 4.0                                & 40.7                              & 4.6 & 60.7 & 7.0  & 64.0 & 3.5  \\
		Water Plant               & 60.0                            & 6.9                                & 79.3                        & 9.5                                & 76.0                              & 6.0 & 88.0 & 4.0  & 90.7 & 5.0  \\
		Unlock iPad /B            & 0.0                             & 0.0                                & 0.0                         & 0.0                                & 0.7                               & 1.2 & 34.0 & 5.3  & 36.0 & 2.0  \\
		Hanoi /B                  & 8.0                             & 2.0                                & 4.7                         & 3.1                                & 29.3                              & 2.3 & 22.7 & 12.2 & 20.7 & 4.6  \\
		Assembly /B               & 1.3                             & 2.3                                & 4.0                         & 5.3                                & 8.0                               & 5.3 & 5.3  & 4.2  & 8.0  & 5.3  \\
		Microwave /B              & 42.7                            & 6.4                                & 34.7                        & 13.6                               & 70.0                              & 9.2 & 58.0 & 0.0  & 68.0 & 5.3  \\
		Photograph /B             & 28.0                            & 6.0                                & 7.3                         & 5.0                                & 59.3                              & 6.4 & 53.3 & 4.2  & 59.3 & 11.0 \\
		Avg.                      & 28.2                            & 2.2                                & 35.0                        & 2.8                                & 41.6                              & 0.3 & 52.5 & 2.7  & 53.6 & 1.9  \\
		\bottomrule
	\end{tabular}
	\vspace{2pt}
	\caption{\textbf{Multi-task, dynamics, and action-head evaluations.} ``multi-task'': models trained jointly on all tasks; ``rand-dynamics'': evaluation with randomized dynamics parameters; ``rand-AH'': \(\pi_{0.5}\) with randomly reinitialized action head.}
	\label{tab:ablation}
	\vspace{-5mm}
\end{table}


\textbf{Multi-task Training Degradation.} When jointly training on all tasks (Table~\ref{tab:ablation}, multi-task) with the same number of steps as single-task training, DP-T degrades on every task, while \(\pi_{0.5}\) achieves a success rate increase on \textit{Pinch Tongs}, though its average success rate drops.


\textbf{\(\pi_{0.5}\) Shows Stronger Robustness.} Under randomized joint friction, stiffness, and object mass (Table~\ref{tab:ablation}, rand-dynamics), \(\pi_{0.5}\) averages higher success than DP-T. This suggests that our benchmark can capture differences in robustness under varying dynamics.



\textbf{Pretrained Action-Head Yields No Performance Gains.} We compare partial pretrain-AH (Table~\ref{tab:benchmark_results}) against fully random reinitialization (Table~\ref{tab:ablation}, rand-AH). We observe that their error margins overlap on most tasks, indicating no statistically meaningful difference in performance. We also note that, for the bimanual tasks, continuing training from 60k to 120k steps improves the rand-obj average success rate to 59.0\%. Individual task results: Photograph 74.0, Hanoi 39.3, Microwave 77.3, Assembly 6.0, and iPad 34.0.

\textbf{VLA Model Fails to Exhibit Language Generalization.} We train \(\pi_{0.5}\) on \textit{Unlock iPad} using single-digit passwords (1-5) and evaluate on seen digits (1,2,4), arithmetic expressions (1+1, 2+2), and English words (\textit{two}, \textit{one plus one}). The results show that the model defaults to a fixed action bias rather than true language conditioning.

\section{Conclusion}

In this paper, we present DexJoCo, a benchmark and toolkit for task-oriented dexterous manipulation, featuring 11 functionally grounded tasks and 1.1K human demonstration trajectories. We provide a low-cost teleoperation system and an easy-to-use evaluation pipeline for training and evaluating modern behavior cloning and VLA policies. Our experiments reveal that current policies remain sensitive to visual variations and struggle with fine-grained interaction, bimanual coordination, and long-horizon tasks. We further observe that large-scale pretraining and multi-task training do not consistently improve dexterous manipulation performance, while current VLA models show limited language generalization and action-space transfer. We hope DexJoCo serves as a standardized platform for advancing dexterous robot learning.




\bibliographystyle{IEEEtran}
\bibliography{references}

\clearpage

\section*{APPENDIX}

\section{Statistical Analysis for Language Generalization Results}

\begin{figure}[ht]
	\centering
	\includegraphics[width=\linewidth]{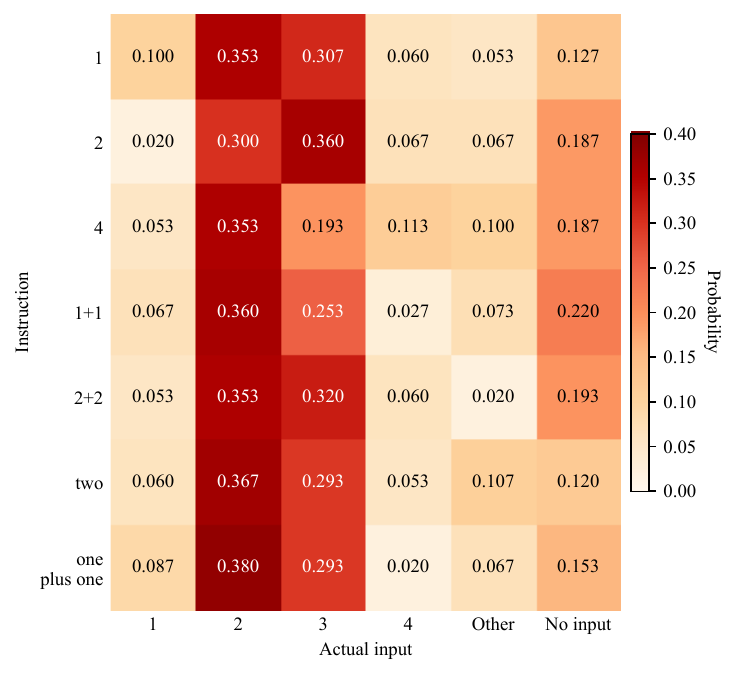}
	\caption{
		Output distribution of \(\pi_{0.5}\) (trained on single digits 1--5) across instructions on the \textit{Unlock iPad}.
	}
	\label{fig:heatmap}
\end{figure}

As shown in Fig.~\ref{fig:heatmap}, the policy exhibits severe mode collapse.
While certain unseen prompts like ``two'' (36.7\%\(\pm\)1.2) and
``1+1'' (36.0\%\(\pm\)2.0) appear to yield moderate precision, the heatmap
reveals this to be a statistical illusion caused by the model's prior bias.
Specifically, the probability of outputting ``2'' remains nearly constant
(\(\sim\)35\%) even when the correct answer is ``1'' or ``4''. This lack of
language conditioning is further evidenced by the model's failure on the
seen digit ``4'', where precision drops to only 11.3\%\(\pm\)5.0 because the
model stubbornly outputs ``2'' (35.3\%) or ``3'' (19.3\%) instead of the
requested digit.

Quantitatively, although a chi-square test rejects the hypothesis of strict
independence (\(p = 6.37 \times 10^{-3}\)), confirming that the VLA does react
to varying language instructions, the Normalized Mutual Information between
instruction and output is only \(0.015\), indicating a negligible relationship.
The average JS divergence across all pairs of instructions is \(0.023\), with
a maximum of \(0.040\) (between ``4'' and ``one plus one''), further
demonstrating that the policy's action distribution remains nearly identical
regardless of the prompt. We therefore conclude that the model fails to
achieve true language generalization. The average precision (\%) \(\pm\) std
across 3 seeds is ``1'': 10.0\%\(\pm\)2.0; ``2'': 30.0\%\(\pm\)5.3;
``4'': 11.3\%\(\pm\)5.0; ``1+1'': 36.0\%\(\pm\)2.0;
``2+2'': 6.0\%\(\pm\)2.0; ``two'': 36.7\%\(\pm\)1.2;
``one plus one'': 38.0\%\(\pm\)8.7.

\begin{table}[t]
	\centering
	\footnotesize
	\setlength{\tabcolsep}{2pt}
	\renewcommand{\arraystretch}{1.08}

	\begin{tabular}{
		>{\centering\arraybackslash}p{0.13\columnwidth}
		>{\raggedright\arraybackslash}p{0.85\columnwidth}
		}
		\toprule
		\textbf{Password} & \textbf{Language Instruction}                                                                 \\
		\midrule

		``1''             & ``Grasp the iPad and enter the password 1 to unlock the device.''                             \\
		``2''             & ``Grasp the iPad and enter the password 2 to unlock the device.''                             \\
		``3''             & ``Grasp the iPad and enter the password 3 to unlock the device.''                             \\
		``4''             & ``Grasp the iPad and enter the password 4 to unlock the device.''                             \\
		``5''             & ``Grasp the iPad and enter the password 5 to unlock the device.''                             \\
		``1+1''           & ``Grasp the iPad and enter the result of 1+1 as the password to unlock the device.''          \\
		``2+2''           & ``Grasp the iPad and enter the result of 2+2 as the password to unlock the device.''          \\
		``two''           & ``Grasp the iPad and enter the password two to unlock the device.''                           \\
		``one plus one''  & ``Grasp the iPad and enter the result of one plus one as the password to unlock the device.'' \\
		\bottomrule
	\end{tabular}
	\caption{Detailed language instructions used in the language generalization experiment.}\label{tab:ipad_reasoning_prompt}
\end{table}

\section{Randomization Settings of DexJoCo Tasks}

\begin{figure}[ht]
	\centering
	\includegraphics[width=0.49\textwidth]{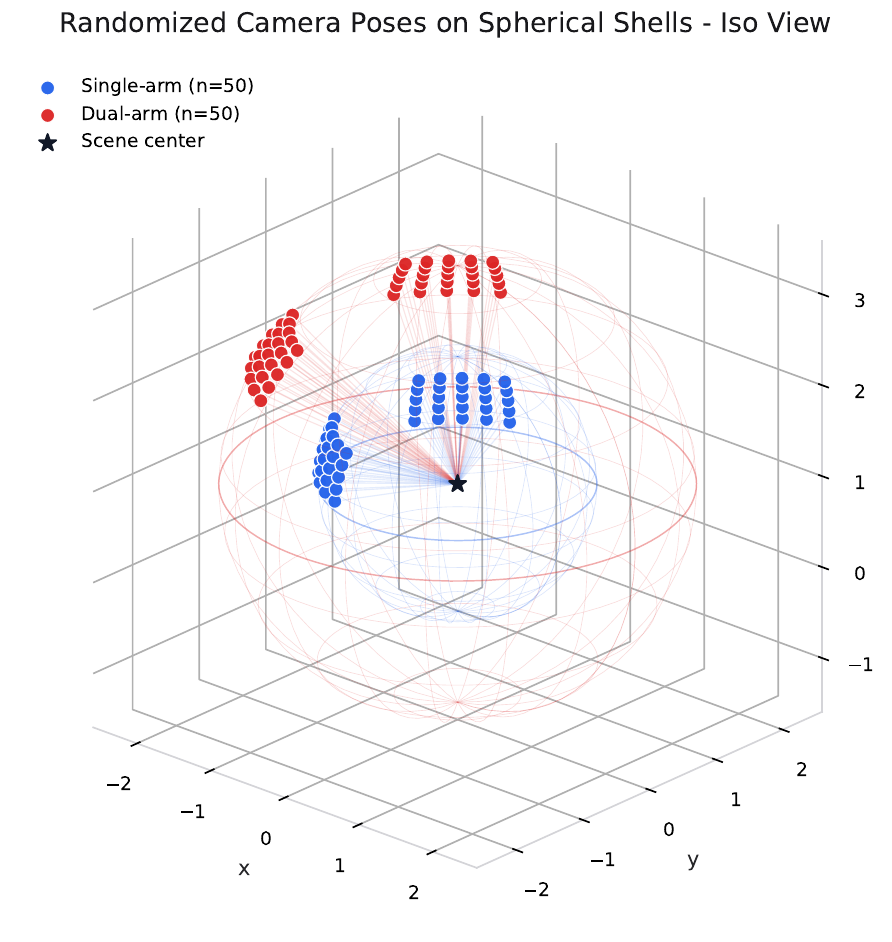}
	\caption{Preset third-person camera poses used for visual randomization.}
	\label{fig:app_camera_pose_viz}
\end{figure}

The visual randomization protocol is shared by all 11 task environments.
At reset time, each environment samples one preset third-person camera pose
from the replay-camera pool, randomly selects a tabletop texture from the
texture library, and perturbs scene lighting. Specifically, each light
position is perturbed in the \(x\) and \(y\) axes by \(U(-0.3,0.3)\), each light
direction is perturbed in the \(x\) and \(y\) axes by \(U(-0.4,0.4)\), light
diffuse RGB values are sampled from \(U(0.3,0.8)\), headlight ambient RGB
values are sampled from \(U(0.3,0.7)\), and headlight diffuse RGB values are
sampled from \(U(0.2,0.6)\). Fig.~\ref{fig:app_camera_pose_viz} visualizes the
preset camera-pose pool used for third-person view randomization.

Table-height randomization is also shared across all task environments.
At reset time, the table height offset is sampled as
\(\Delta h \sim U(0,0.05)\) m, and task-relevant object heights are shifted
consistently with this offset.

\begin{figure*}[htbp]
	\centering
	\includegraphics[width=0.95\textwidth]{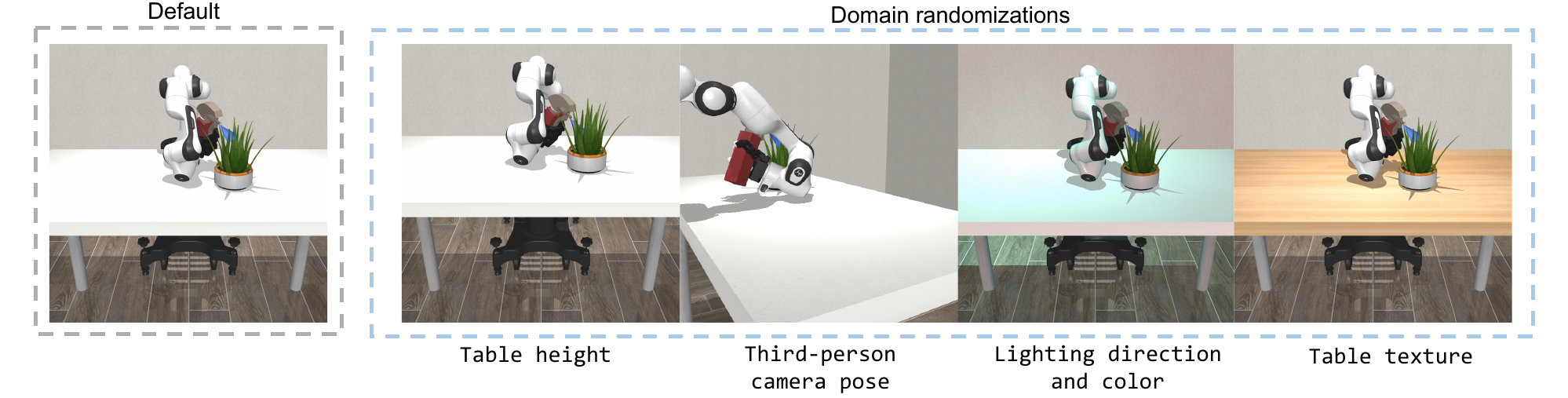}
	\caption{
		\textbf{Domain randomization settings.} The left panel shows the default scene configuration, while the right panel illustrates the effects of domain randomization, including variations in table height, third-person camera viewpoints, lighting conditions, and tabletop textures.
	}
	\label{fig:app_domain_randomization}
\end{figure*}

\begin{table*}
	\scriptsize
	\setlength{\tabcolsep}{4pt}
	\renewcommand{\arraystretch}{1.22}
	\begin{tabular}{
		>{\raggedright\arraybackslash}m{0.10\linewidth}
		>{\raggedright\arraybackslash}m{0.52\linewidth}
		>{\raggedright\arraybackslash}m{0.32\linewidth}
		}
		\toprule
		\textbf{Task} & \textbf{Object randomization} & \textbf{Dynamics randomization} \\
		\midrule

		Hammer Nail &
		\begin{minipage}[t]{\linewidth}\raggedright
			Hammer \((x,y)\): low \([-0.40,-0.50]\), high \([-0.25,-0.35]\); yaw \(\sim U(-10^\circ,10^\circ)\).\\
			Nail \((x,y)\): low \([-0.10,0.00]\), high \([0.00,0.10]\).
		\end{minipage}    &
		Hammer mass multiplier: \(U(0.75,1.25)\).\\
		\midrule

		Click Mouse &
		\begin{minipage}[t]{\linewidth}\raggedright
			Mouse \((x,y)\): low \([-0.25,0.00]\), high \([-0.20,0.05]\);
			yaw \(\sim U(-10^\circ,10^\circ)\).\\
			Monitor/mouse-pad target \((x,y)\): fixed at \([0.12,0.30]\).
		\end{minipage} &
		Mouse mass multiplier: \(U(0.75,1.25)\).\\
		\midrule

		Pick Bucket &
		\begin{minipage}[t]{\linewidth}\raggedright
			Bucket \((x,y)\): low \([-0.20,-0.25]\), high \([-0.15,-0.20]\); yaw \(\sim U(-10^\circ,10^\circ)\).\\
			Boxed food \((x,y)\): low \([-0.35,0.15]\), high \([-0.30,0.20]\); yaw \(\sim U(-10^\circ,10^\circ)\).
		\end{minipage} &
		\begin{minipage}[t]{\linewidth}\raggedright
			Bucket joint friction multiplier: \(U(0.75,1.25)\).\\
			Bucket and boxed-food mass multipliers: \(U(0.75,1.25)\).
		\end{minipage}\\
		\midrule

		Pinch Tongs &
		Tongs \((x,y)\): low \([-0.35,-0.25]\), high \([-0.30,-0.20]\). &
		\begin{minipage}[t]{\linewidth}\raggedright
			Tongs joint friction loss: \(U(0,0.05)\).\\
			Joint stiffness multiplier: \(U(0.75,1.25)\).\\
			Tongs mass multiplier: \(U(0.75,1.25)\).
		\end{minipage}\\
		\midrule

		Fold Glasses &
		\begin{minipage}[t]{\linewidth}\raggedright
			Glasses \((x,y)\): low \([-0.40,-0.225]\), high \([-0.35,-0.175]\); yaw \(\sim U(-10^\circ,10^\circ)\).\\
			Storage box \((x,y)\): low \([-0.275,0.25]\), high \([-0.225,0.30]\); yaw \(\sim U(-10^\circ,10^\circ)\).
		\end{minipage} &
		\begin{minipage}[t]{\linewidth}\raggedright
			Glasses joint friction loss: \(U(0,0.05)\).\\
			Joint stiffness multiplier: \(U(1.0,1.5)\).\\
			Glasses mass multiplier: \(U(0.75,1.25)\).
		\end{minipage}\\
		\midrule

		Water Plant &
		\begin{minipage}[t]{\linewidth}\raggedright
			Spray bottle \((x,y)\): from \([-0.35,-0.25]\) to \([-0.30,-0.20]\).\\
			Plant \((x,y)\): from \([-0.10,0.15]\) to \([-0.05,0.20]\).
		\end{minipage} &
		\begin{minipage}[t]{\linewidth}\raggedright
			Spray joint friction loss: \(U(0,0.05)\).\\
			Joint stiffness multiplier: \(U(0.75,1.25)\).\\
			Spray body mass multiplier: \(U(0.75,1.25)\).
		\end{minipage}\\
		\midrule

		Unlock iPad /B &
		\begin{minipage}[t]{\linewidth}\raggedright
			iPad stand \((x,y)\): low \([-0.35,0.05]\), high \([-0.30,0.10]\).\\
			iPad and stand heights are shifted by the shared height offset.
		\end{minipage} &
		iPad mass multiplier: \(U(0.75,1.25)\).\\
		\midrule

		Hanoi /B &
		\begin{minipage}[t]{\linewidth}\raggedright
			Hanoi base \((x,y)\): low \([-0.25,0.00]\), high \([-0.20,0.00]\).\\
			All disks are translated consistently with the base and shared height offset.
		\end{minipage} &
		Each disk mass multiplier: \(U(0.75,1.25)\).\\
		\midrule

		Assembly /B &
		\begin{minipage}[t]{\linewidth}\raggedright
			Peg \((x,y)\): from \([-0.30,-0.25]\) to \([-0.25,-0.20]\); yaw \(\sim U(-10^\circ,10^\circ)\).\\
			Socket/tray \((x,y)\): from \([-0.30,0.15]\) to \([-0.20,0.25]\); yaw \(\sim U(-20^\circ,20^\circ)\).
		\end{minipage} &
		Peg and socket/tray mass multipliers: \(U(0.75,1.25)\).\\
		\midrule

		Microwave /B &
		\begin{minipage}[t]{\linewidth}\raggedright
			Hot dog \((x,y)\): low \([-0.35,-0.40]\), high \([-0.25,-0.30]\).\\
			Hot dog yaw: \(U(-20^\circ,20^\circ)\).
		\end{minipage} &
		\begin{minipage}[t]{\linewidth}\raggedright
			Microwave joint friction multiplier: \(U(0.75,1.25)\).\\
			Hot dog and plate mass multipliers: \(U(0.75,1.25)\).
		\end{minipage}\\
		\midrule

		Photograph /B&
		\begin{minipage}[t]{\linewidth}\raggedright
			Logo \((y,z)\): from \([-0.10,1.22]\) to \([0.10,1.38]\).\\
			Camera \((x,y)\): from \([-0.30,0.10]\) to \([-0.20,0.20]\).
		\end{minipage} &
		Camera mass multiplier: \(U(0.75,1.25)\).\\
    \bottomrule
	\end{tabular}
    \caption{Task-specific randomization settings for the 11 DexJoCo task environments. Object placement bounds are reported as planar \((x,y)\) sampling ranges following the corresponding environment implementation; shared visual and table-height randomization are described in the text.}
	\label{tab:app_randomization_settings}
\end{table*}

\clearpage
\onecolumn
{
	\scriptsize

	\newlength{\imgcolwidth}
	\setlength{\imgcolwidth}{%
		\dimexpr
		\linewidth
		- 1.6cm
		- 2.9cm
		- 6\tabcolsep
		- 3pt
		\relax
	}
	\setlength{\imgcolwidth}{0.25\imgcolwidth}  

	\begin{longtable}{
		>{\raggedleft\arraybackslash}m{1.6cm}
		>{\centering\arraybackslash}m{\imgcolwidth}
		@{\hspace{1pt}}
		>{\centering\arraybackslash}m{\imgcolwidth}
		@{\hspace{1pt}}
		>{\centering\arraybackslash}m{\imgcolwidth}
		@{\hspace{1pt}}
		>{\centering\arraybackslash}m{\imgcolwidth}
		m{2.9cm}
		}

		\toprule
		\multicolumn{1}{c}{\textbf{Task}} & \multicolumn{4}{c}{\textbf{Visualization}} & \multicolumn{1}{c}{\textbf{Language Instruction}}\\
		\midrule
		\endfirsthead

		\toprule
		\multicolumn{1}{c}{\textbf{Task}} & \multicolumn{4}{c}{\textbf{Visualization}} & \multicolumn{1}{c}{\textbf{Language Instruction}}\\
		\midrule
		\endhead

		\midrule
		\multicolumn{6}{r}{\footnotesize Continued on next page}\\
		\endfoot

		\bottomrule
		\caption{Visualization and language instructions of DexJoCo tasks.}
		\label{tab:app_vis}\\
		\endlastfoot

		Hammer Nail &
		\includegraphics[width=\linewidth]{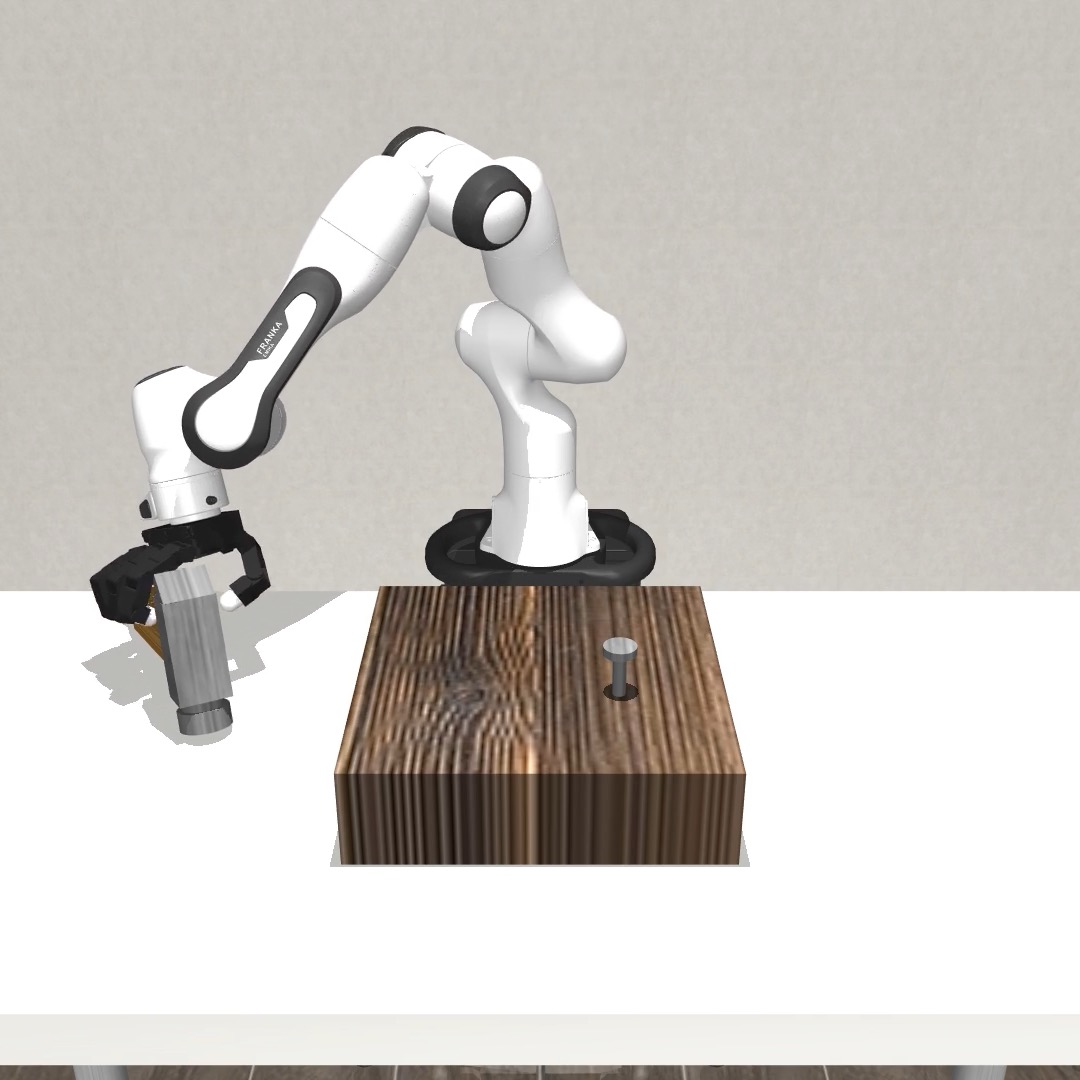} &
		\includegraphics[width=\linewidth]{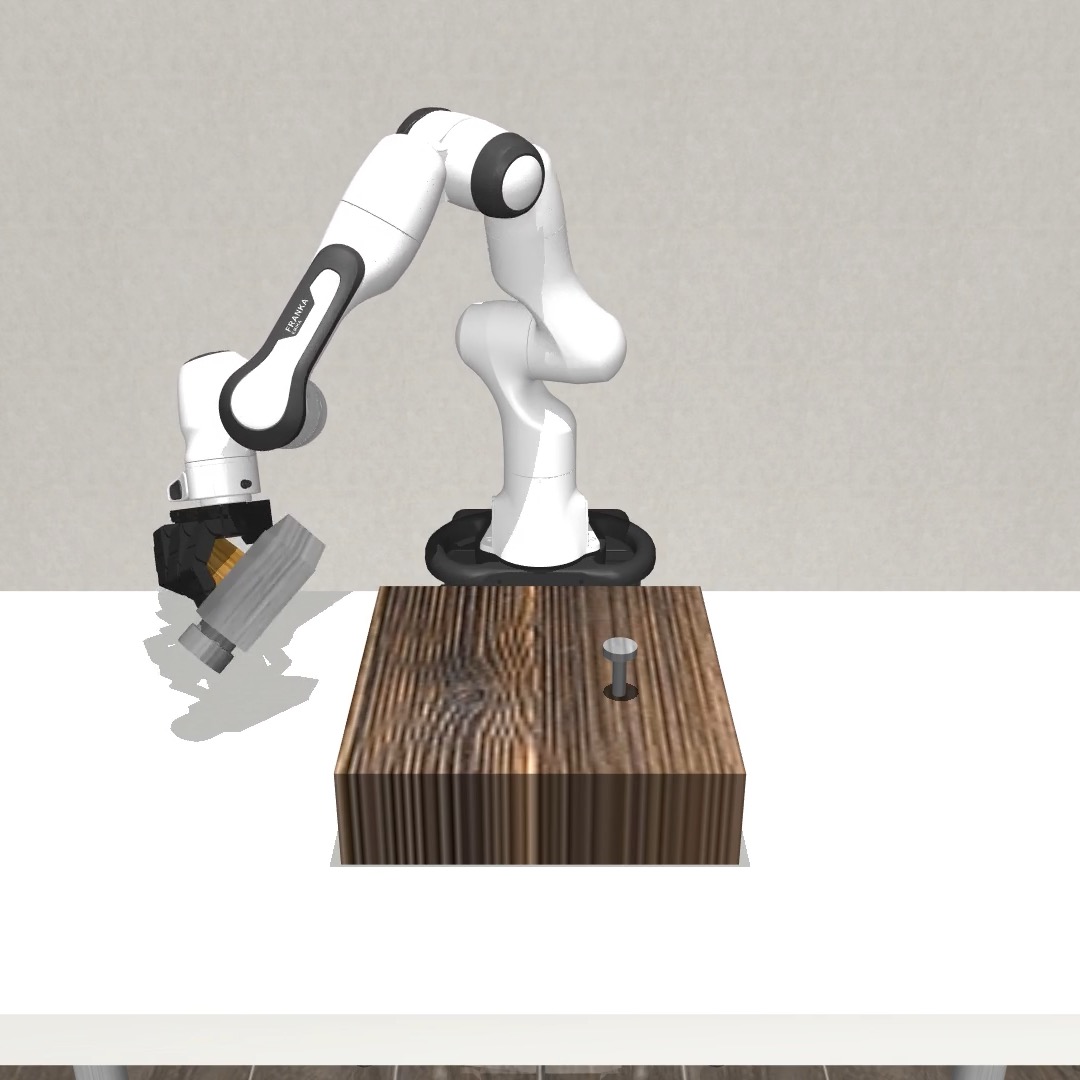} &
		\includegraphics[width=\linewidth]{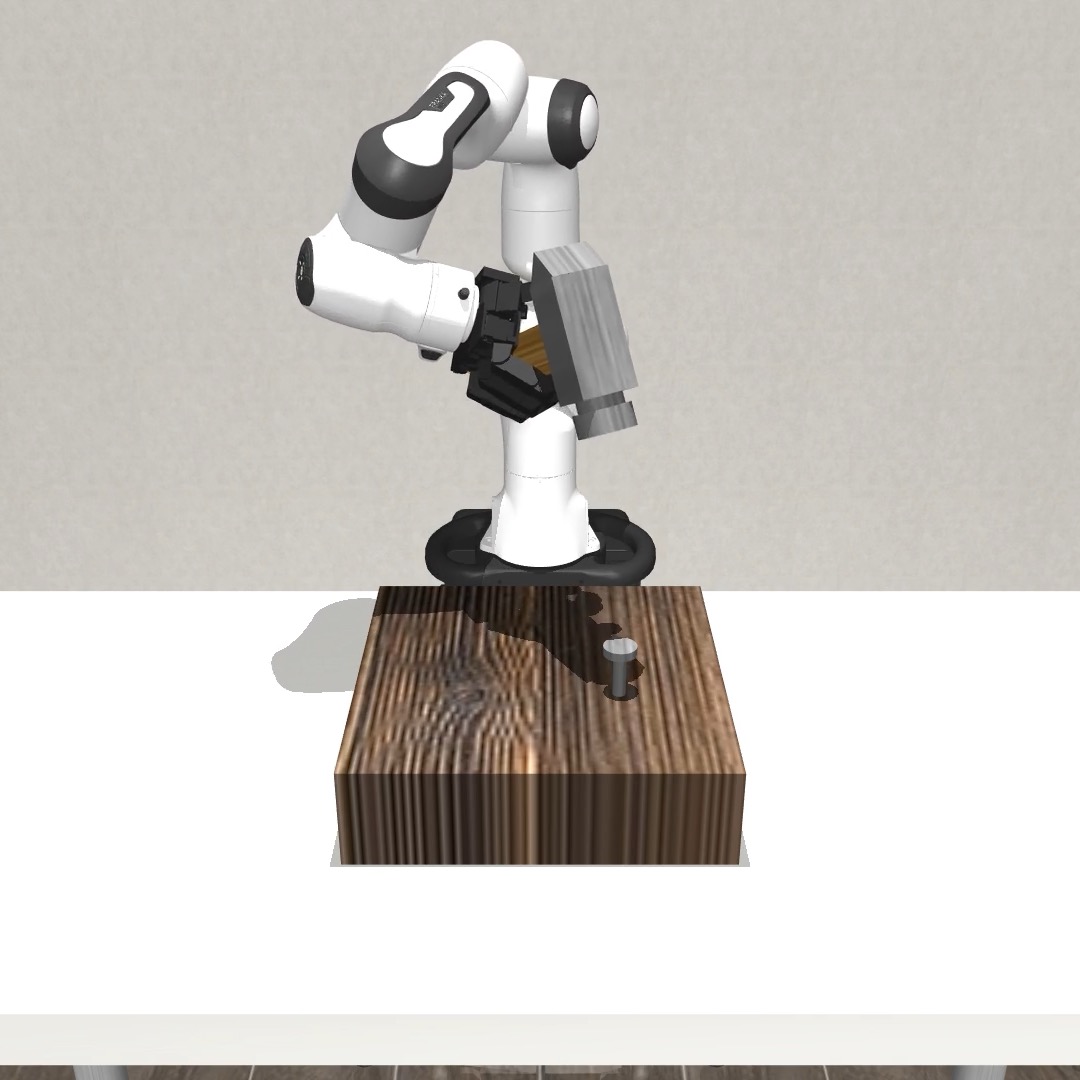} &
		\includegraphics[width=\linewidth]{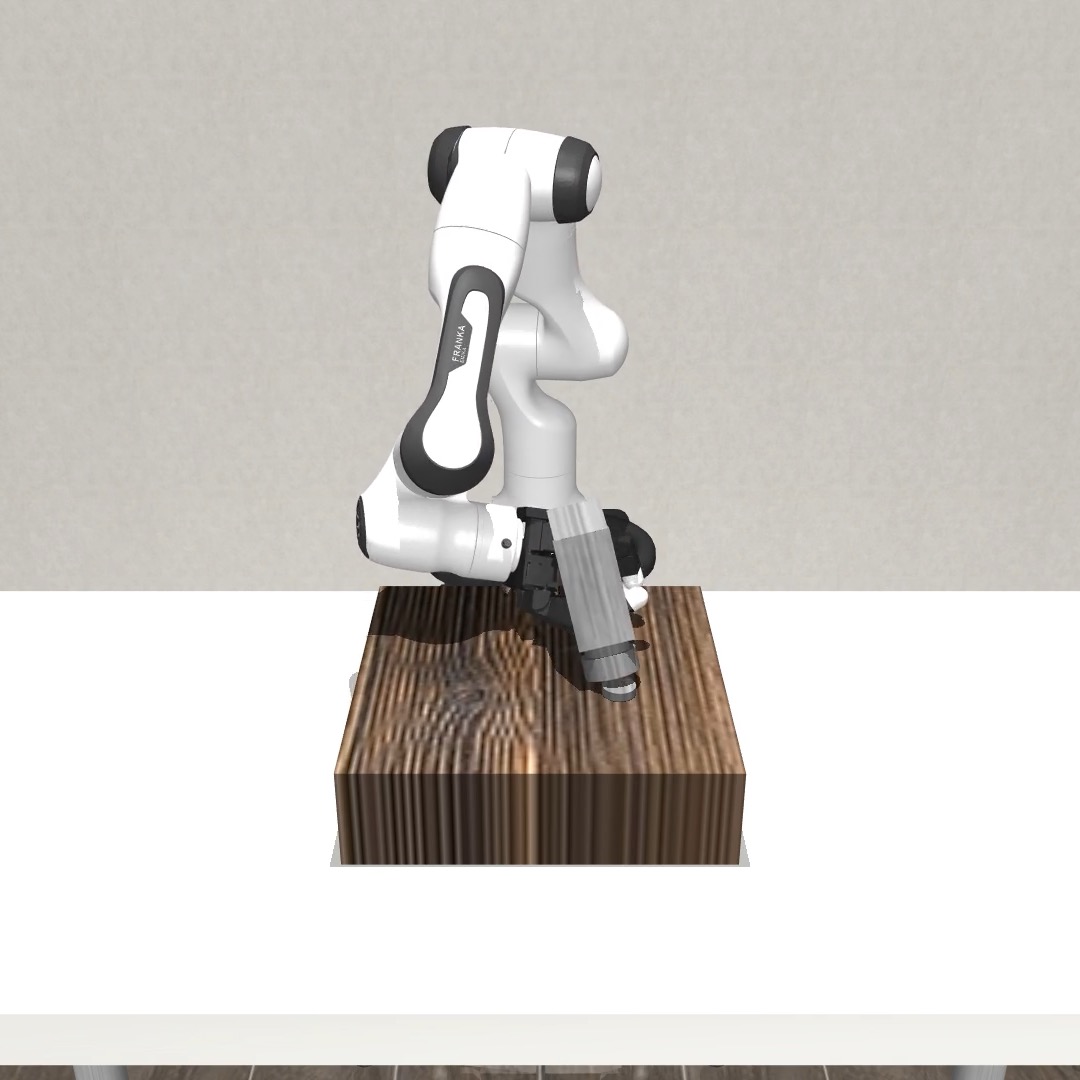} &
		Use the hammer to drive the nail into the wooden board.\\

		Click Mouse &
		\includegraphics[width=\linewidth]{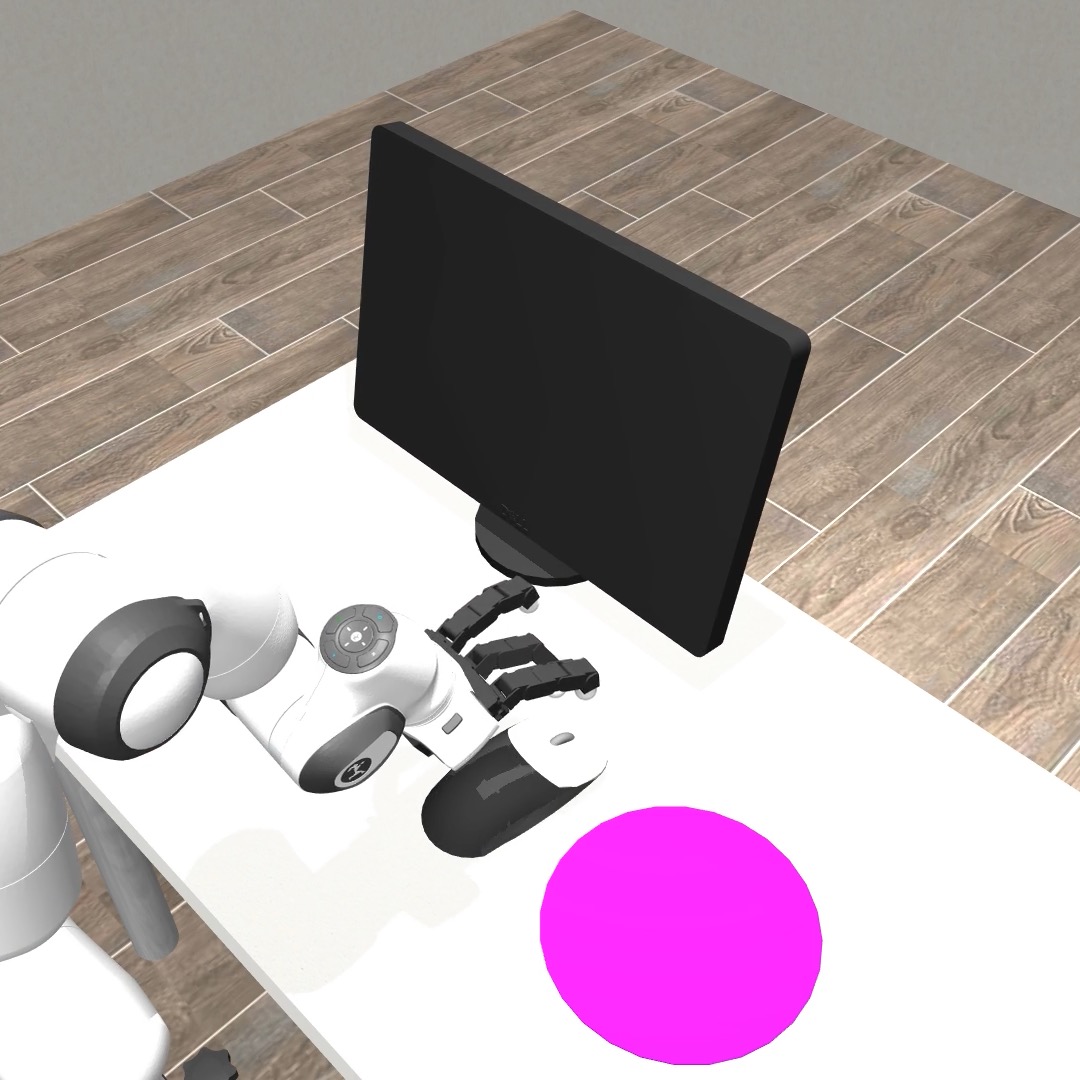} &
		\includegraphics[width=\linewidth]{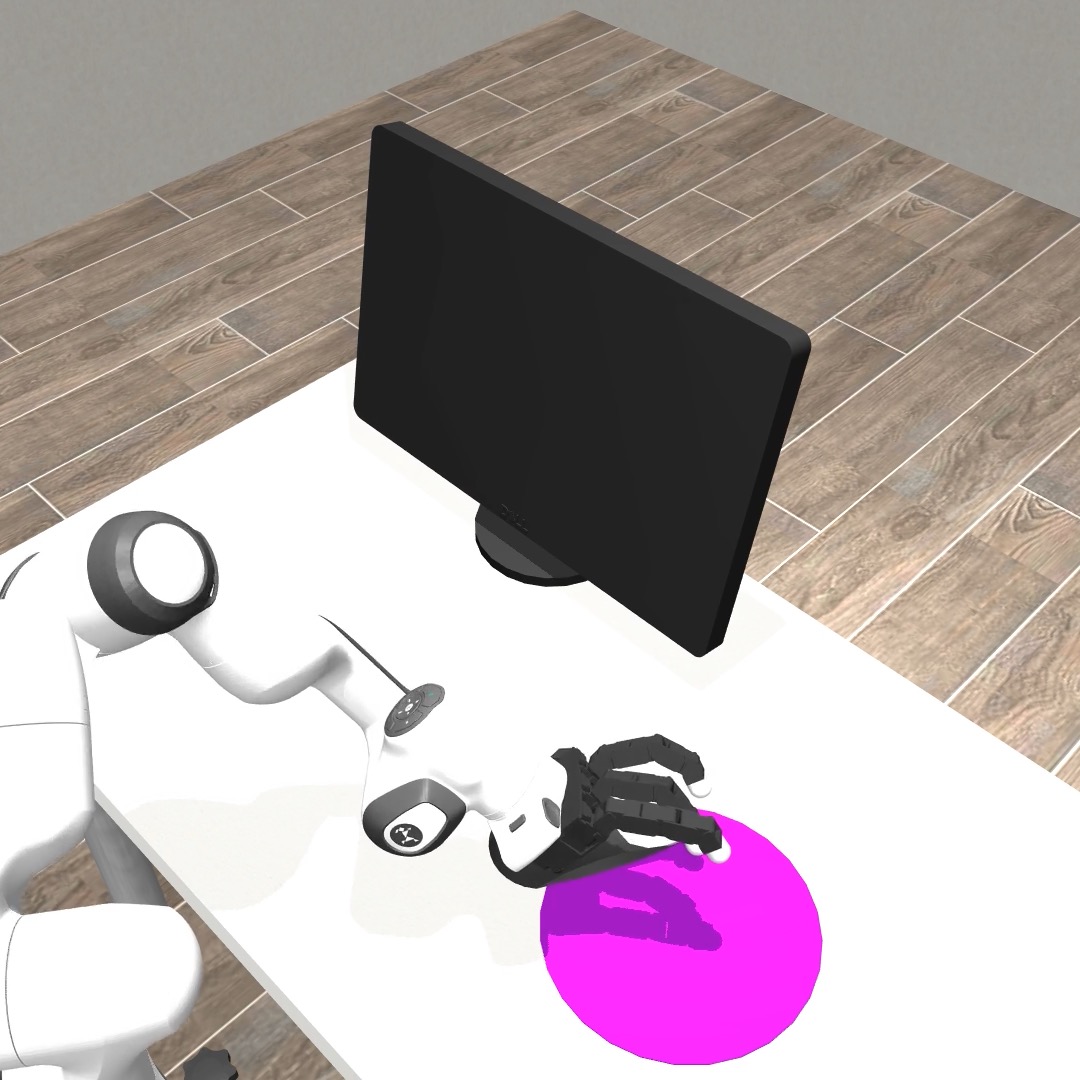} &
		\includegraphics[width=\linewidth]{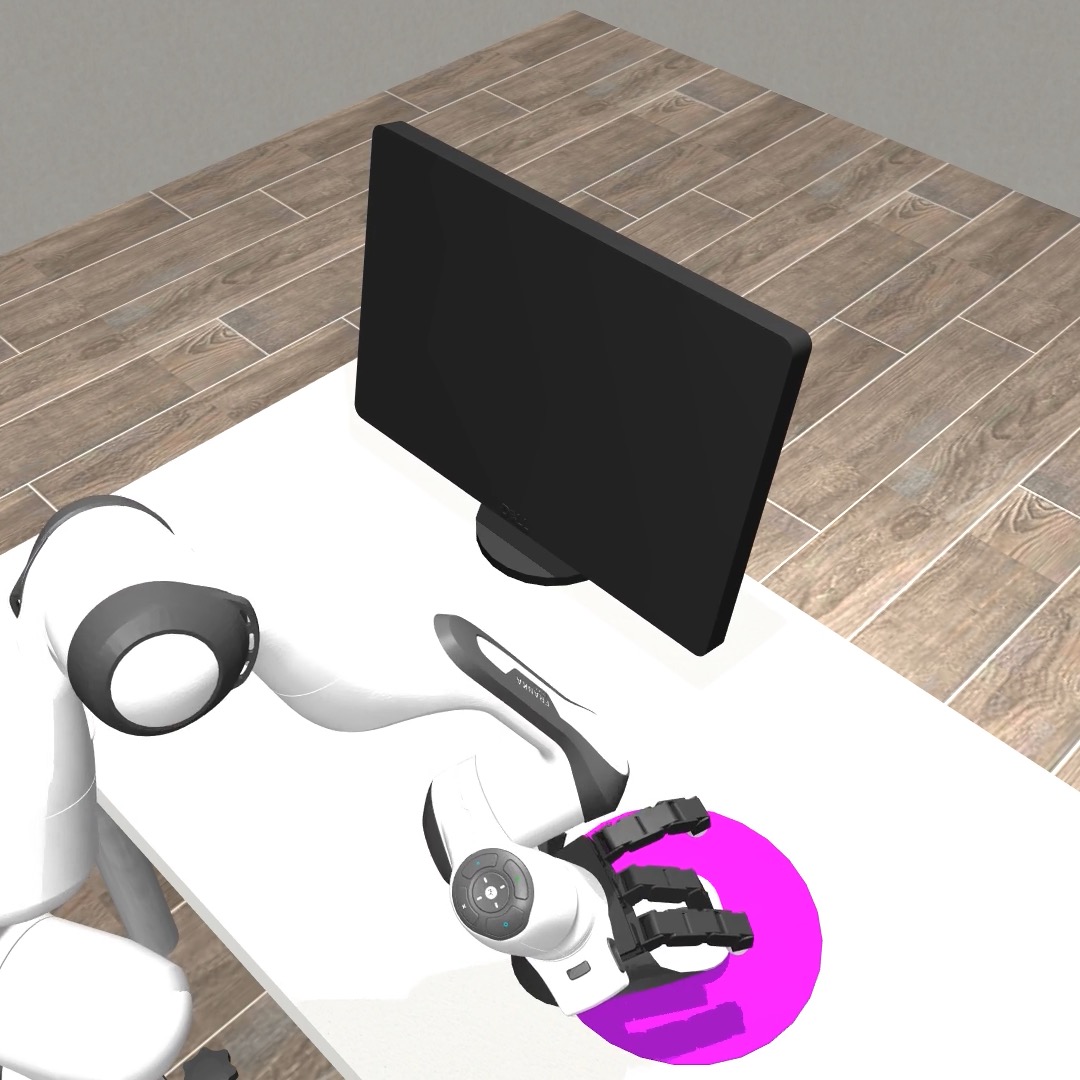} &
		\includegraphics[width=\linewidth]{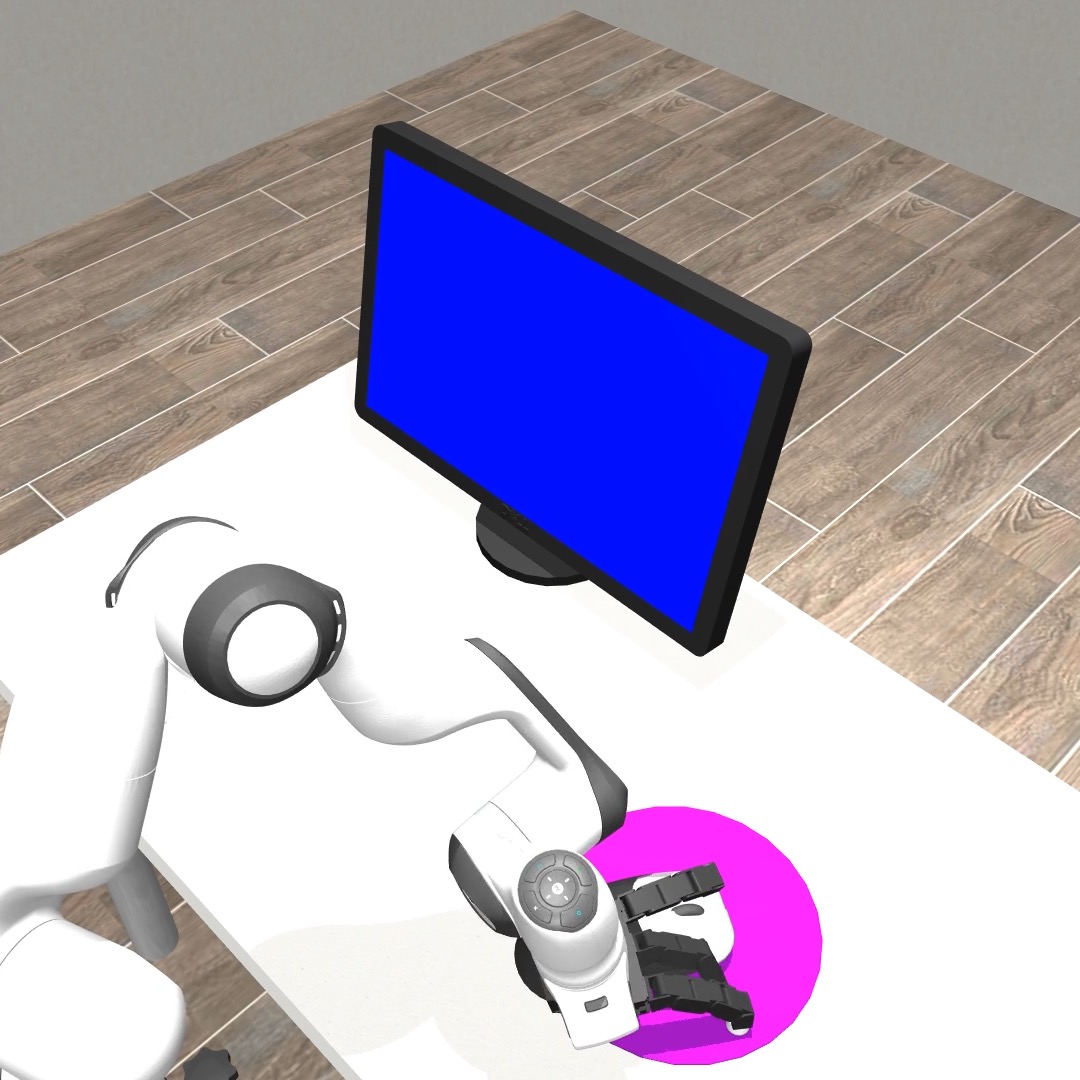} &
		Move the mouse to the purple mouse pad and click the left mouse button.\\

		Pick Bucket &
		\includegraphics[width=\linewidth]{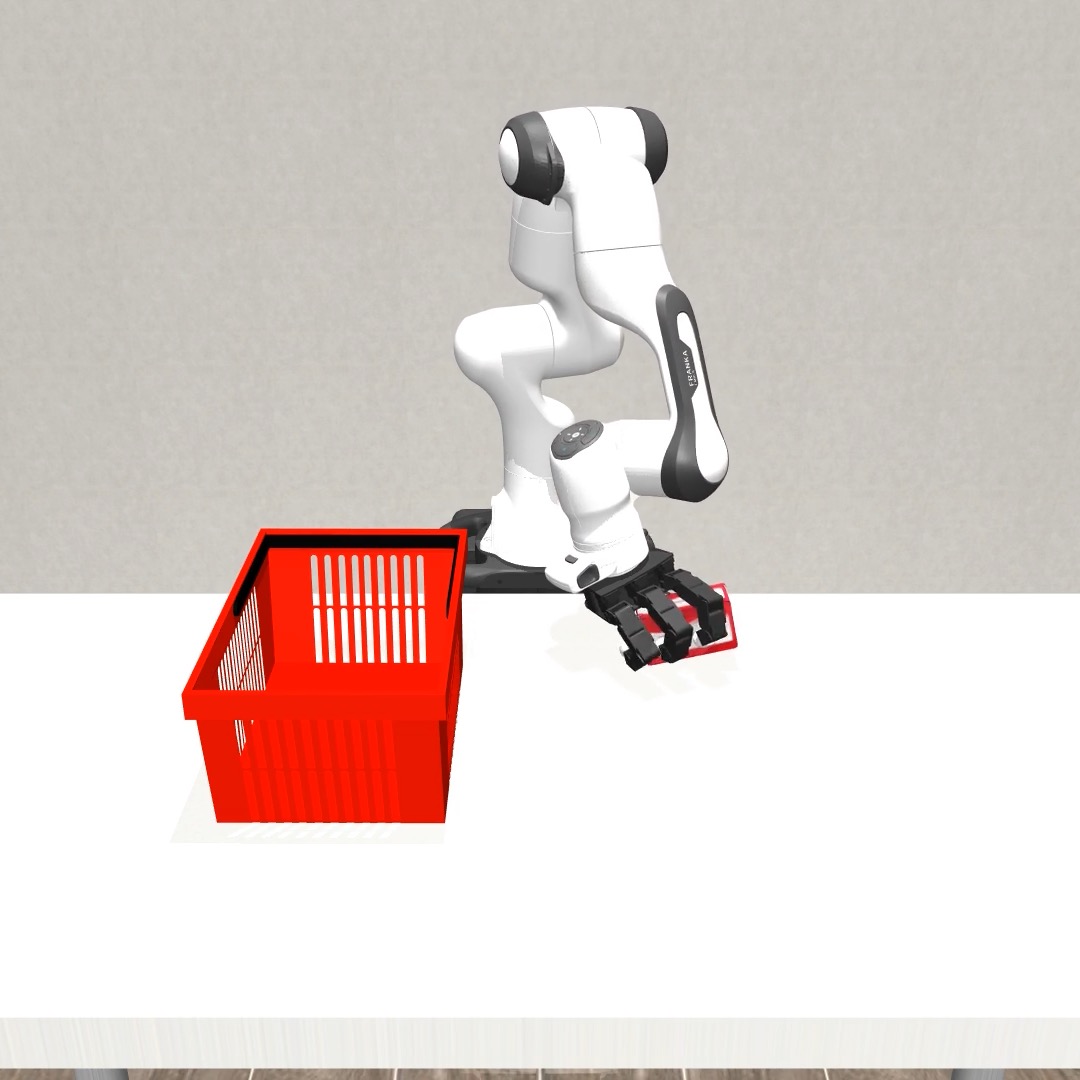} &
		\includegraphics[width=\linewidth]{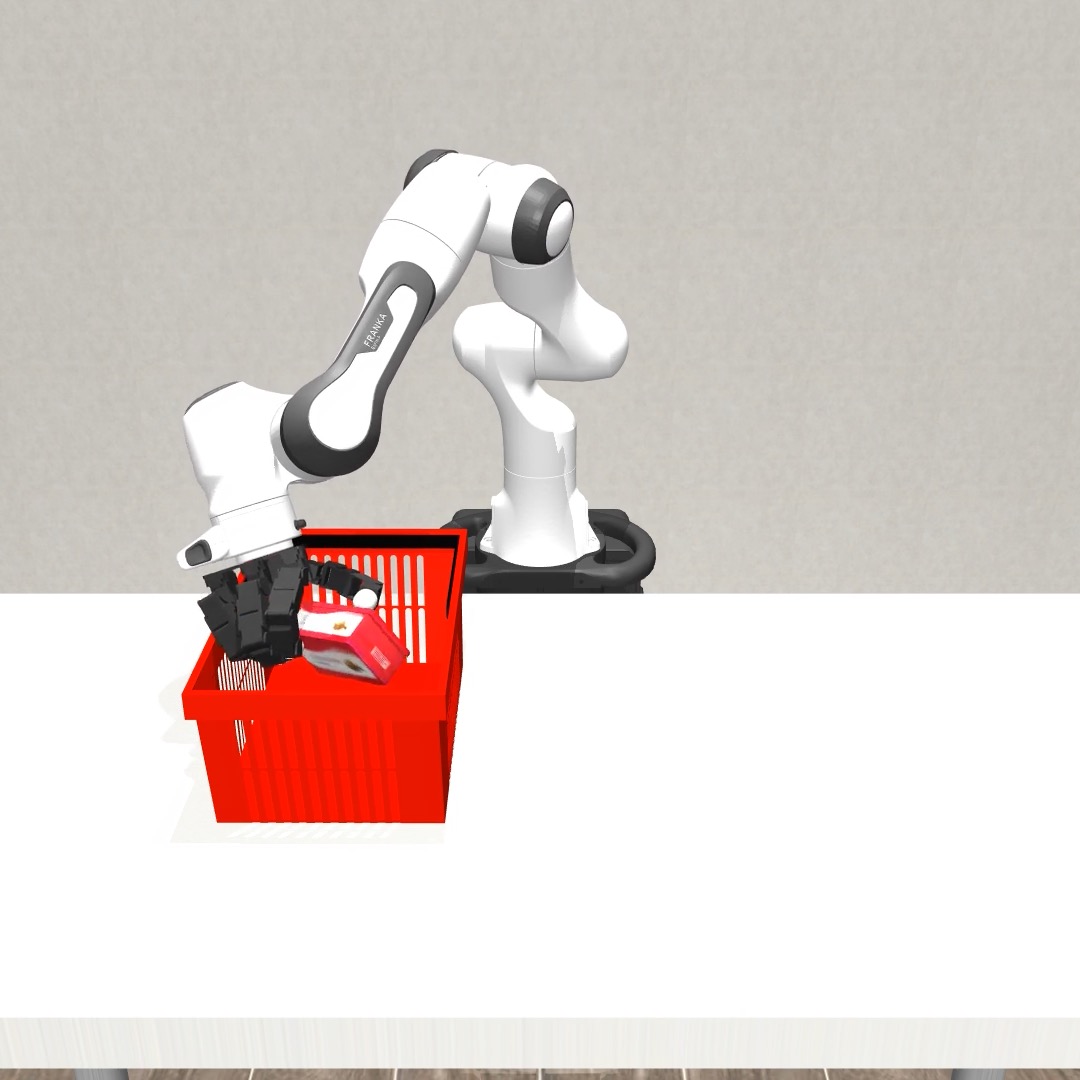} &
		\includegraphics[width=\linewidth]{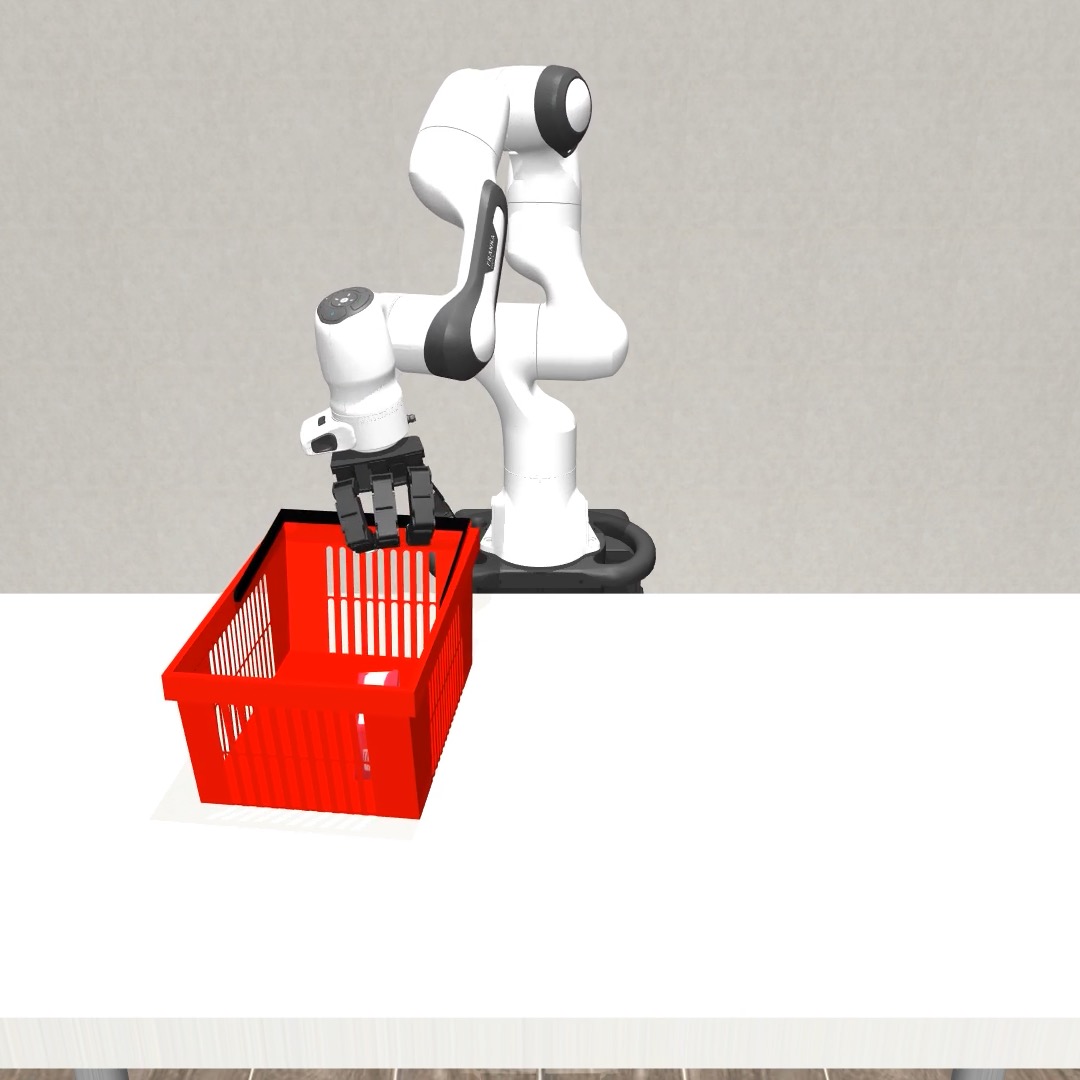} &
		\includegraphics[width=\linewidth]{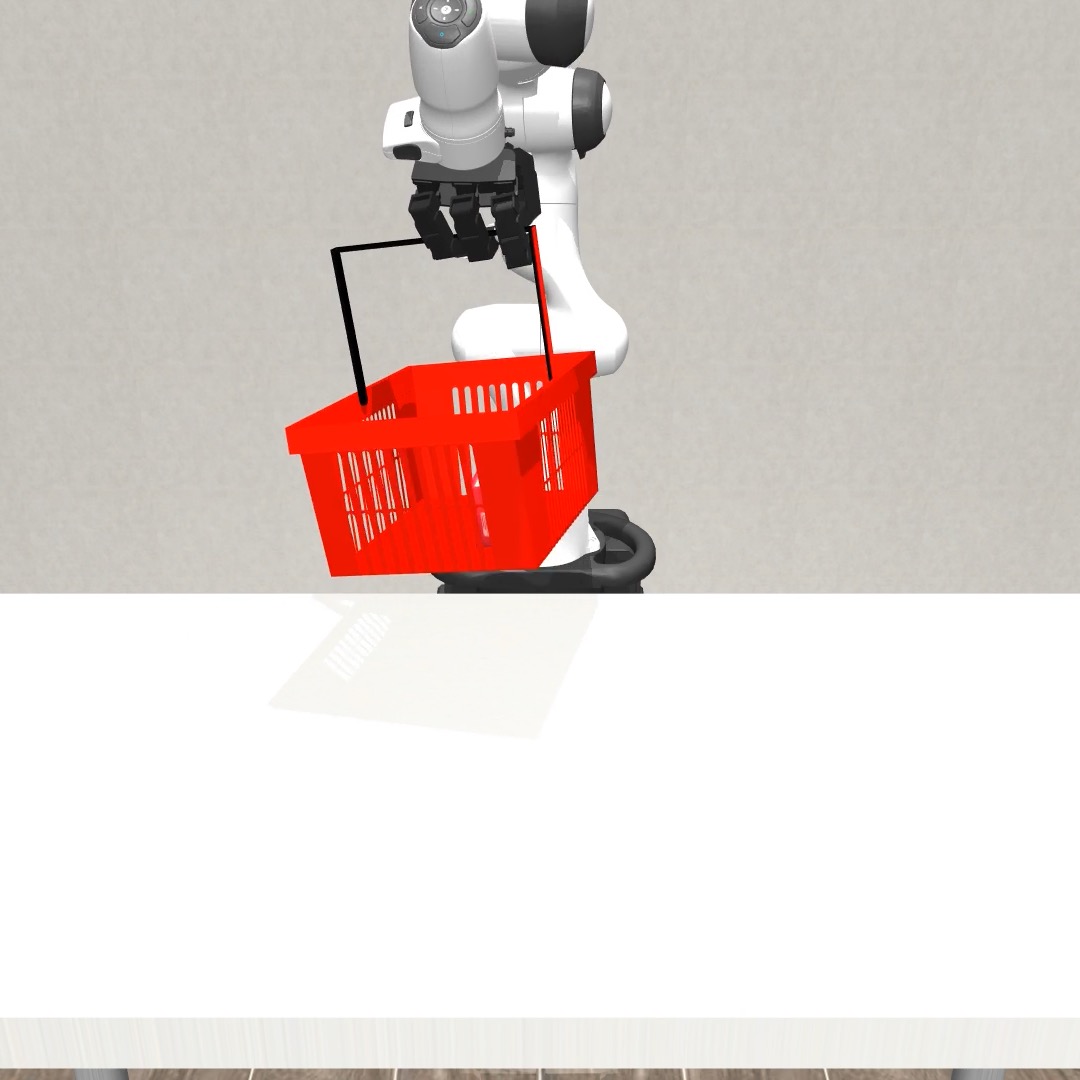} &
		Place the boxed food into the bucket and then lift the bucket.\\

		Pinch Tongs &
		\includegraphics[width=\linewidth]{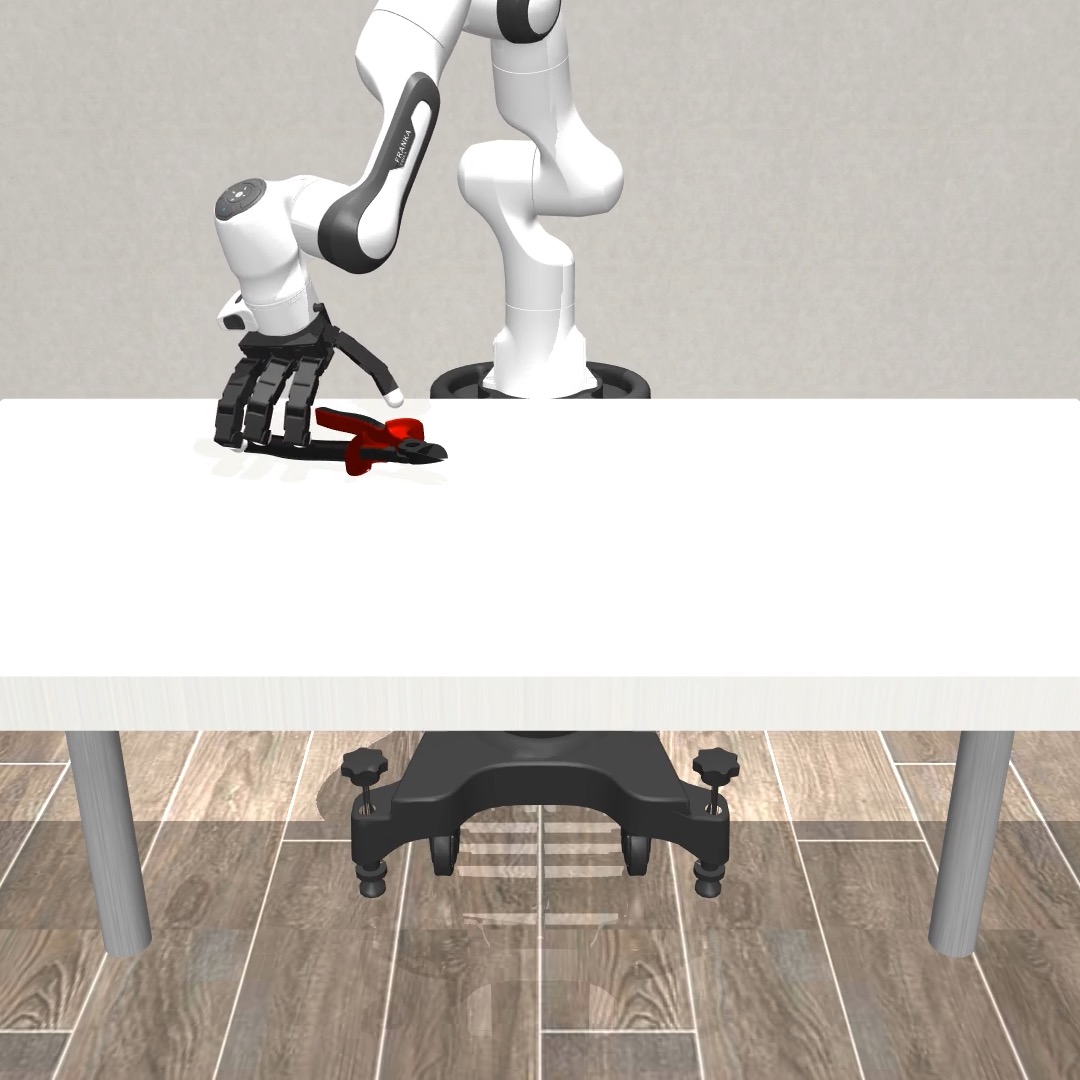} &
		\includegraphics[width=\linewidth]{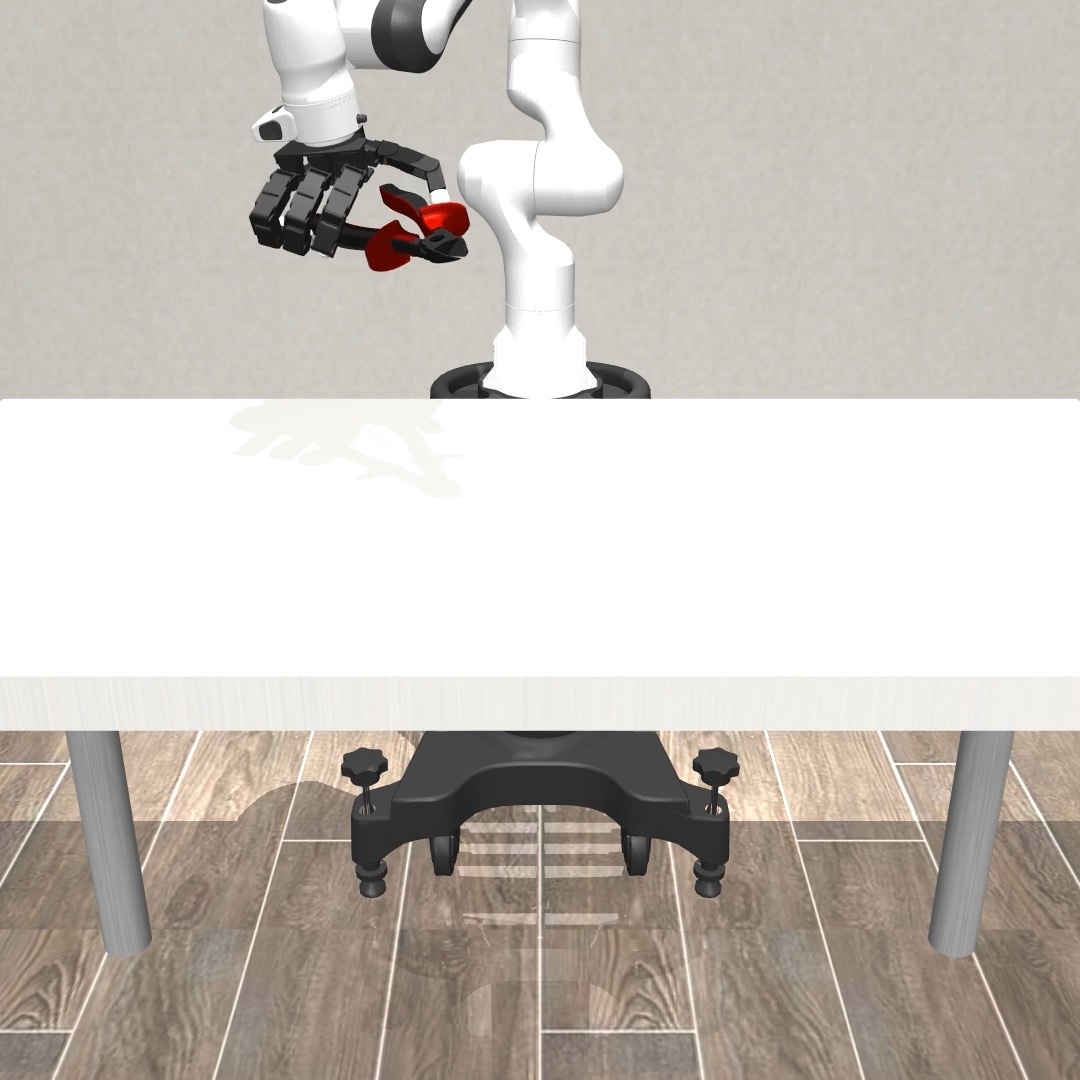} &
		\includegraphics[width=\linewidth]{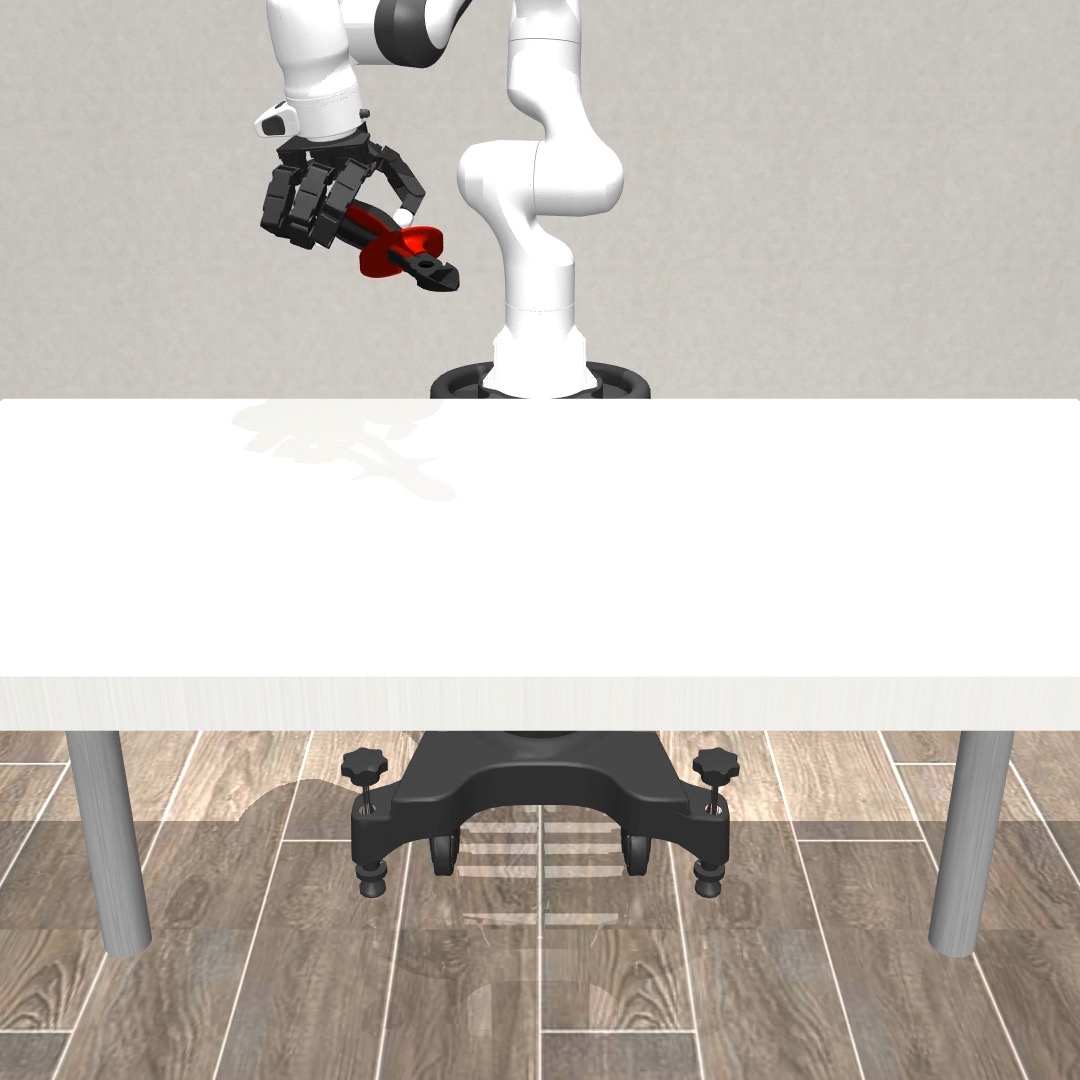} &
		\includegraphics[width=\linewidth]{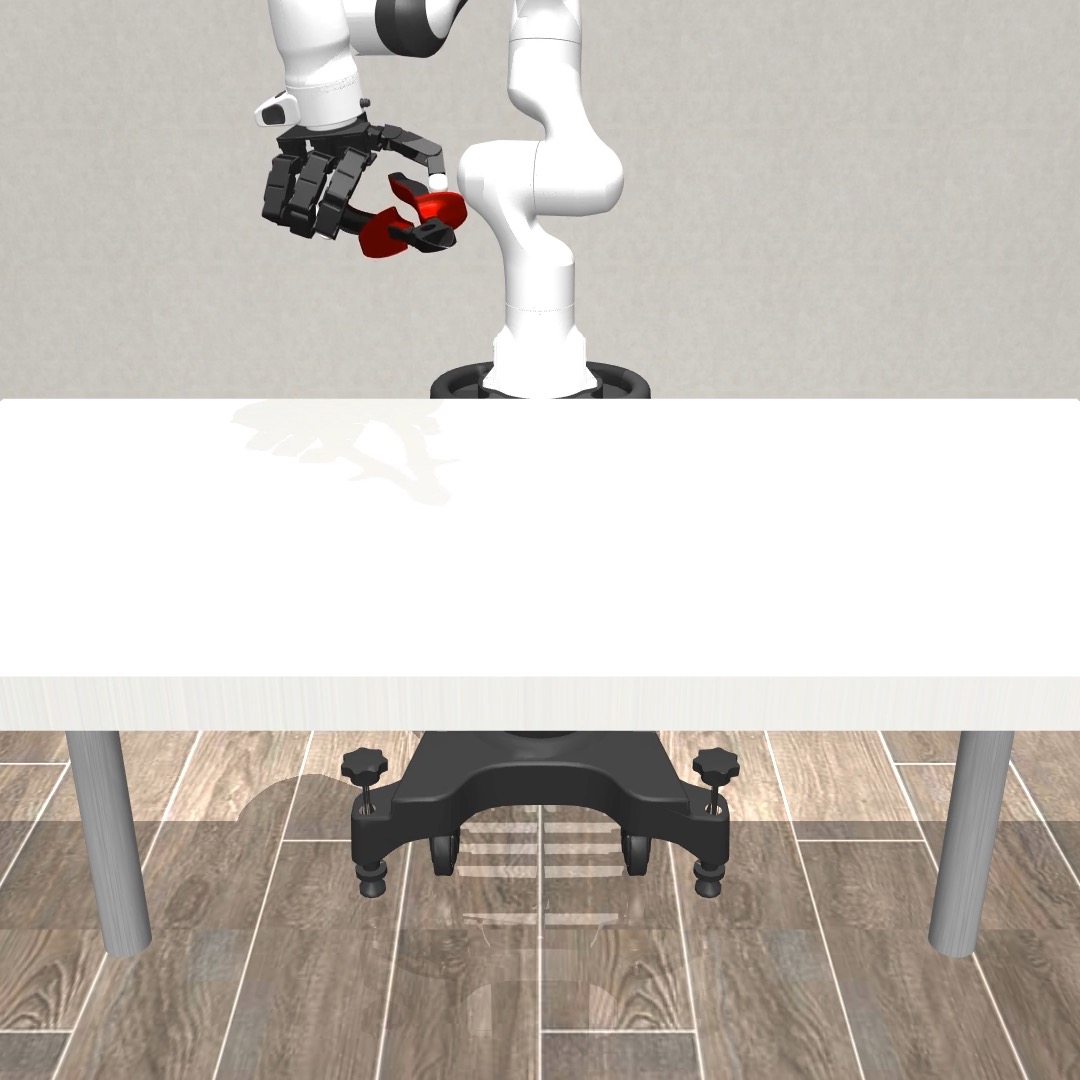} &
		Grasp the tongs and perform three consecutive open-close motions.\\

		Fold Glasses &
		\includegraphics[width=\linewidth]{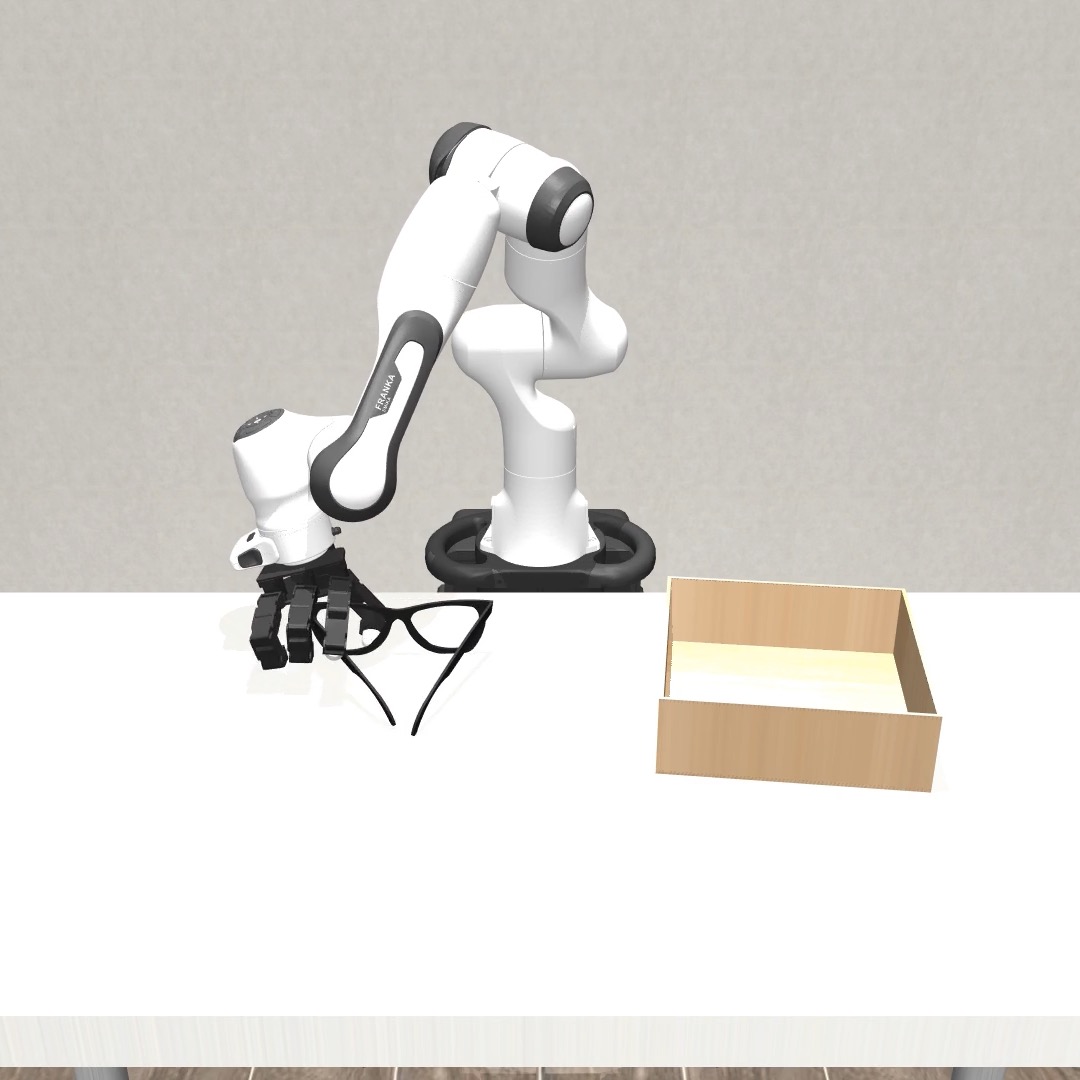} &
		\includegraphics[width=\linewidth]{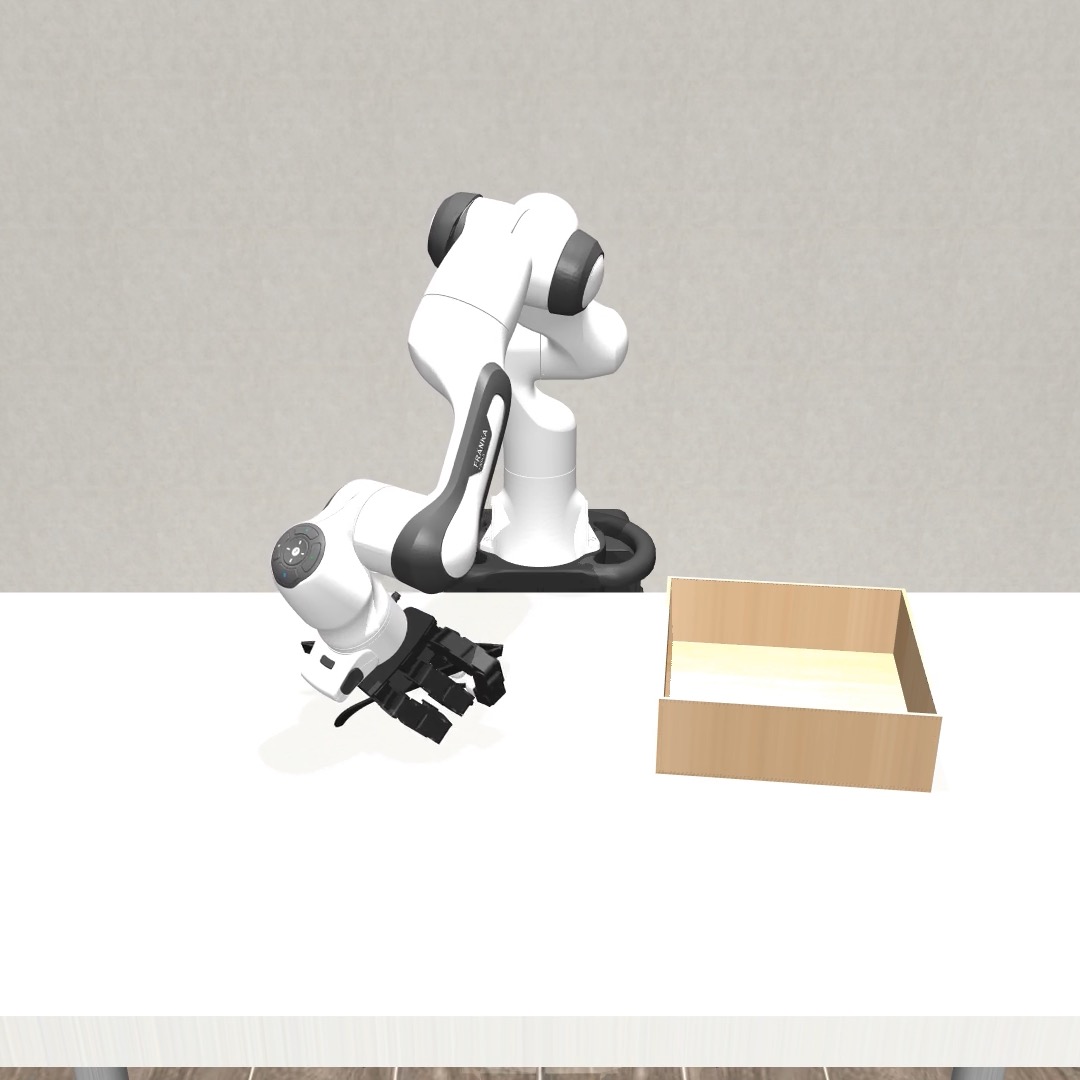} &
		\includegraphics[width=\linewidth]{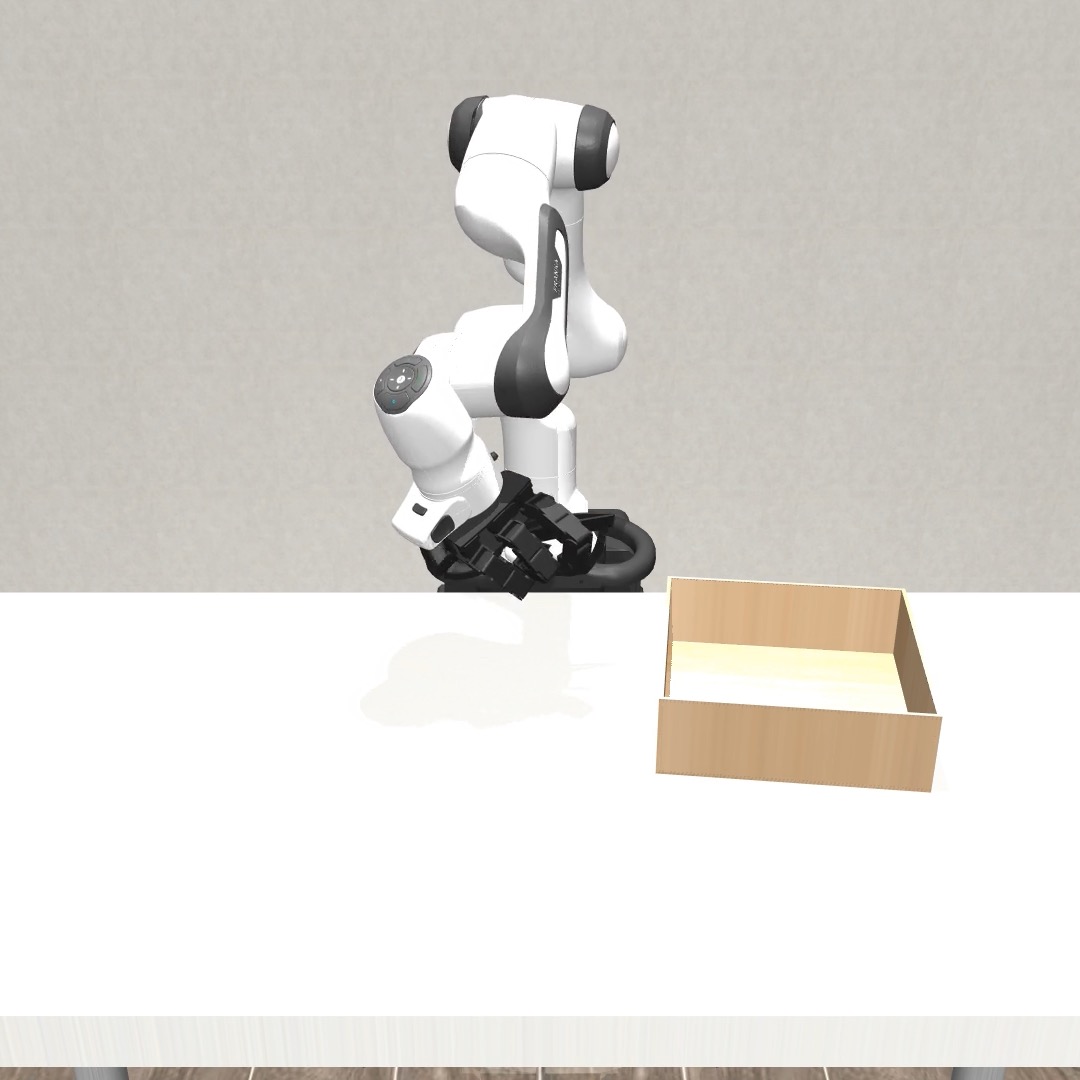} &
		\includegraphics[width=\linewidth]{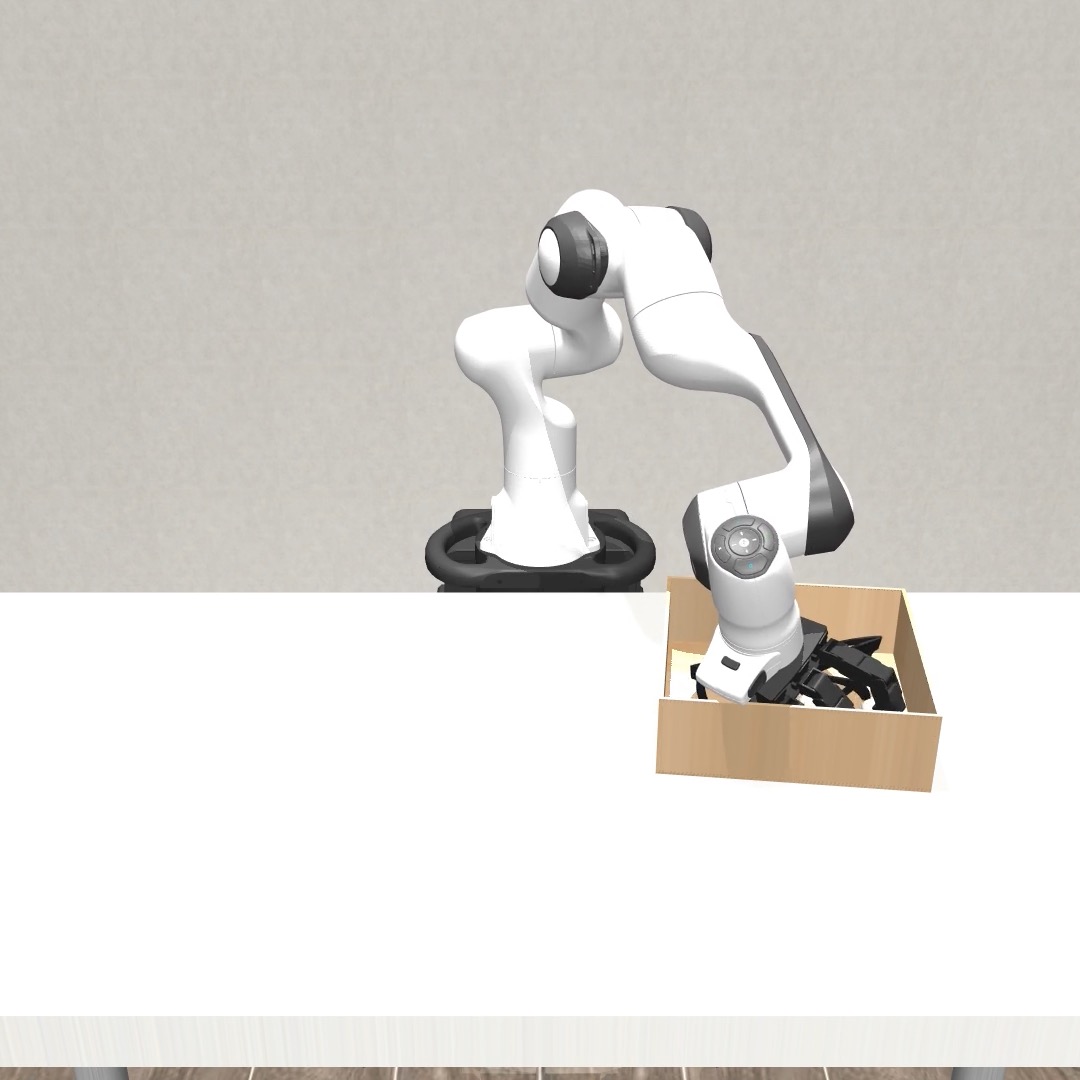} &
		Fold the glasses and place them into the case.\\

		Water Plant &
		\includegraphics[width=\linewidth]{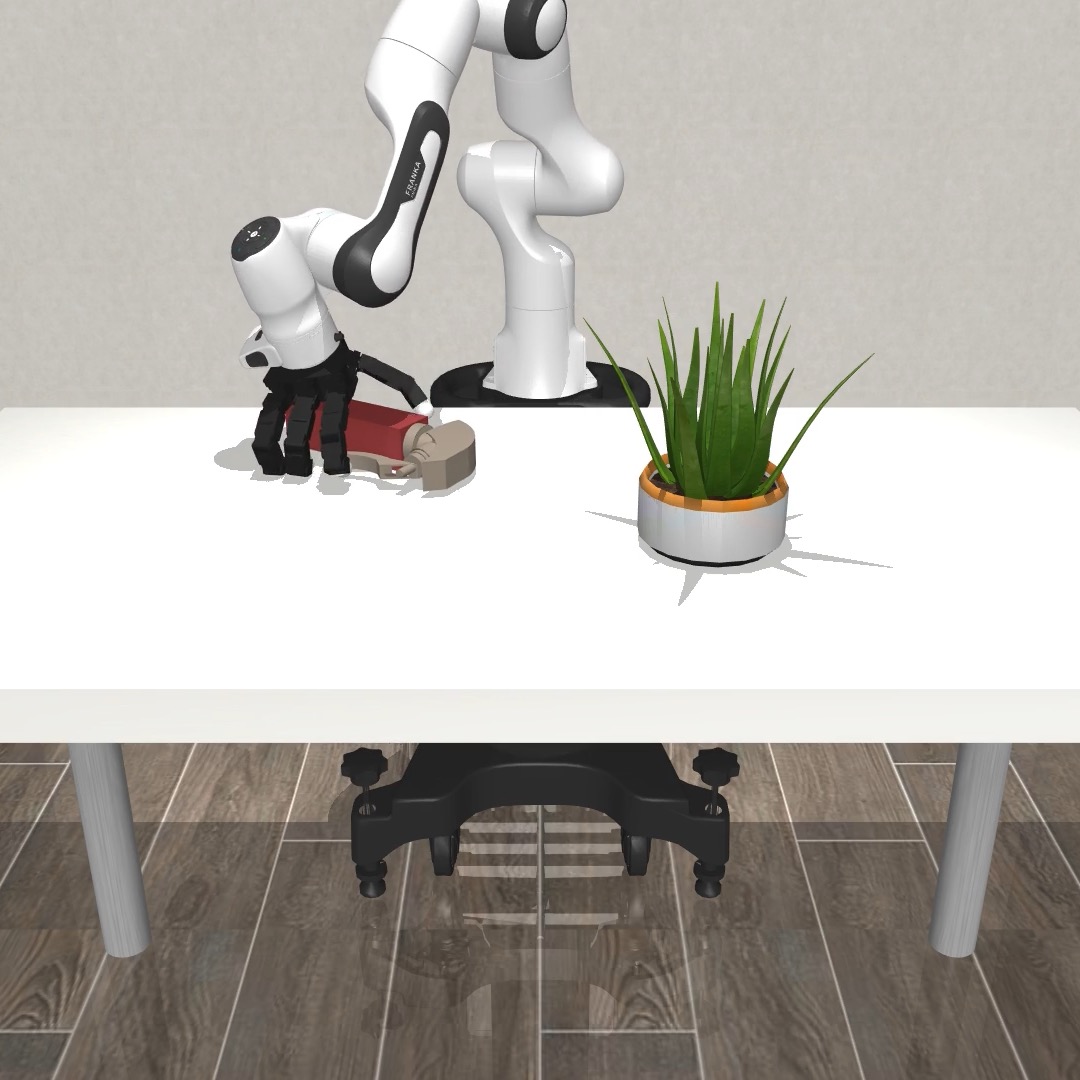} &
		\includegraphics[width=\linewidth]{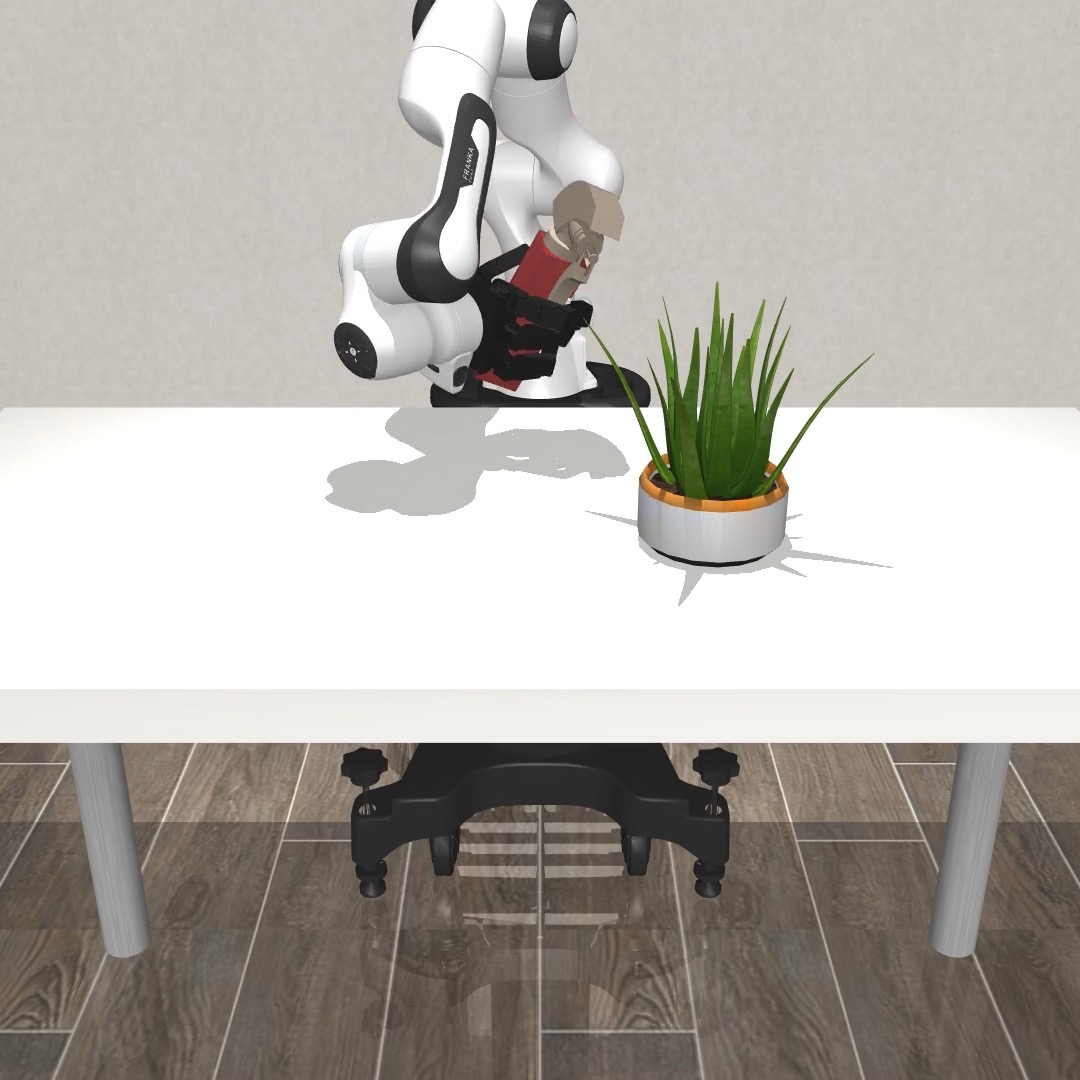} &
		\includegraphics[width=\linewidth]{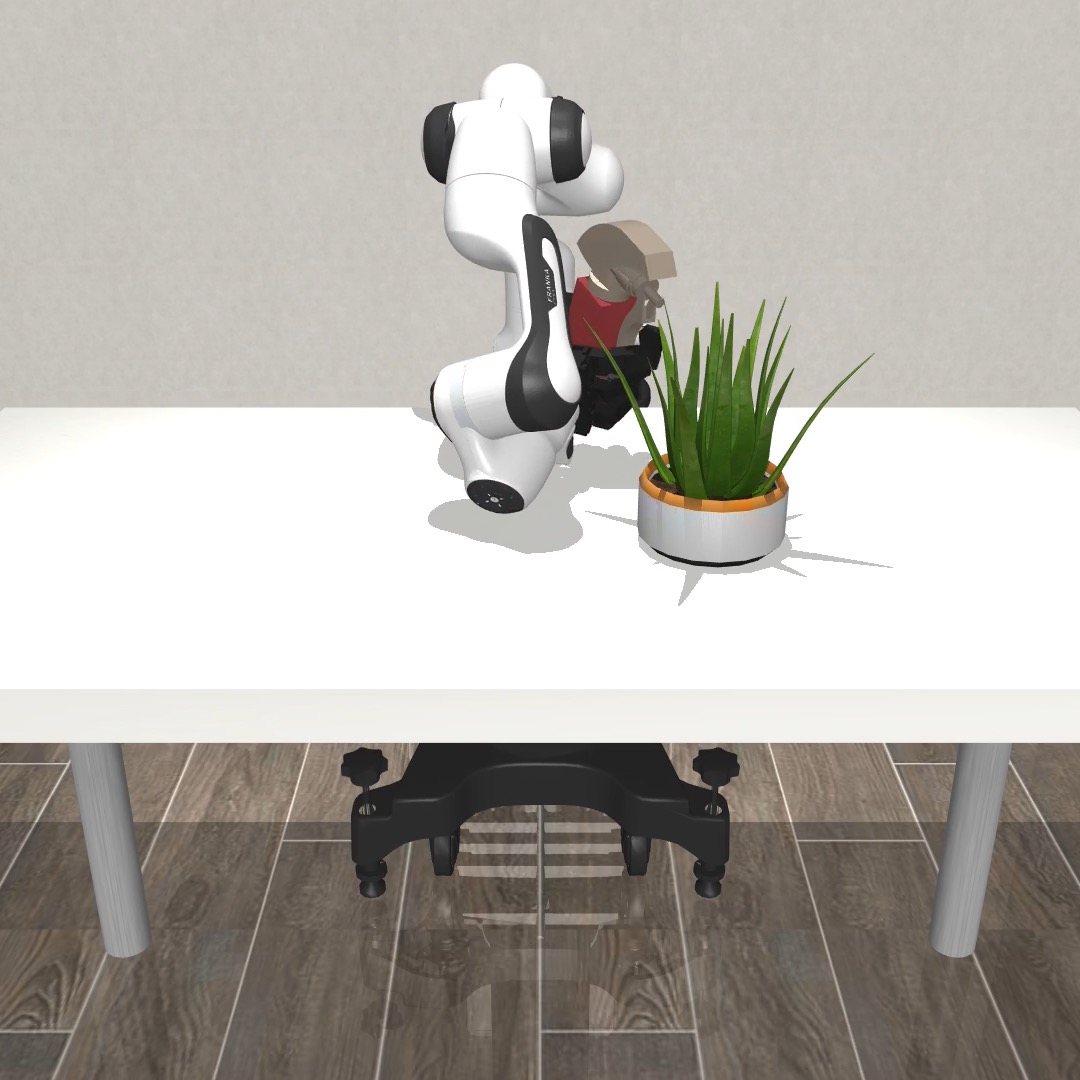} &
		\includegraphics[width=\linewidth]{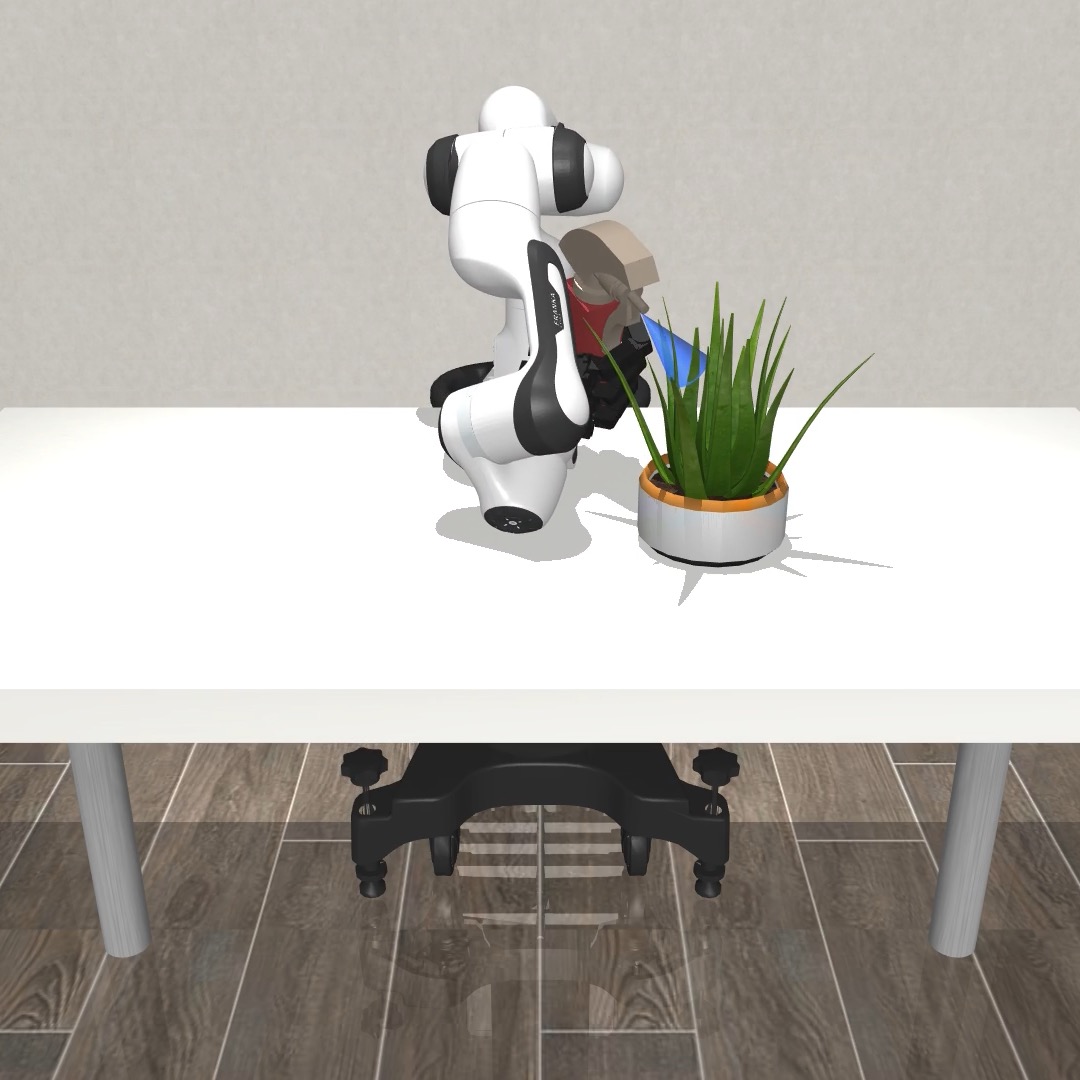} &
		Grasp the watering can and apply water to the plant.\\

		Unlock iPad /B &
		\includegraphics[width=\linewidth]{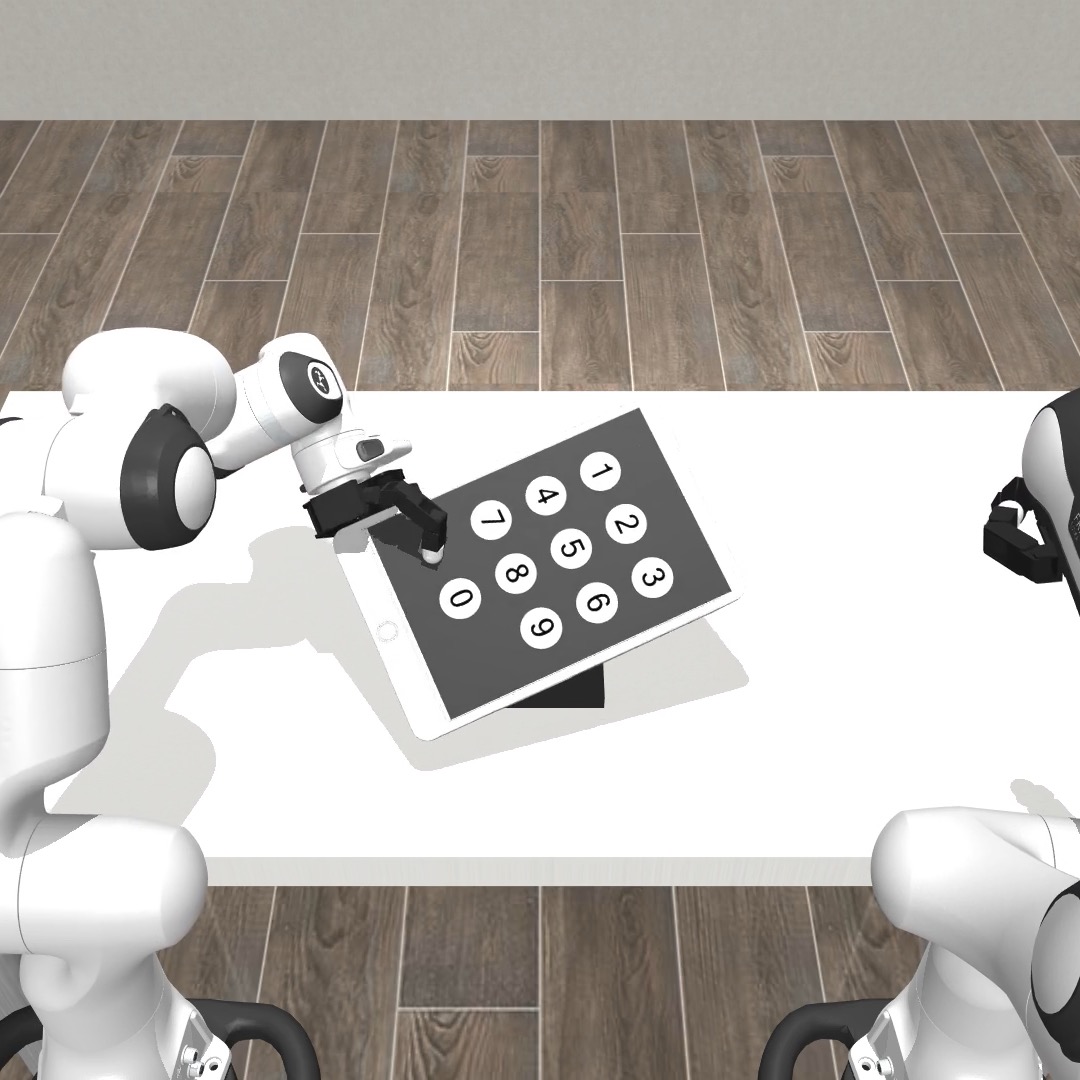} &
		\includegraphics[width=\linewidth]{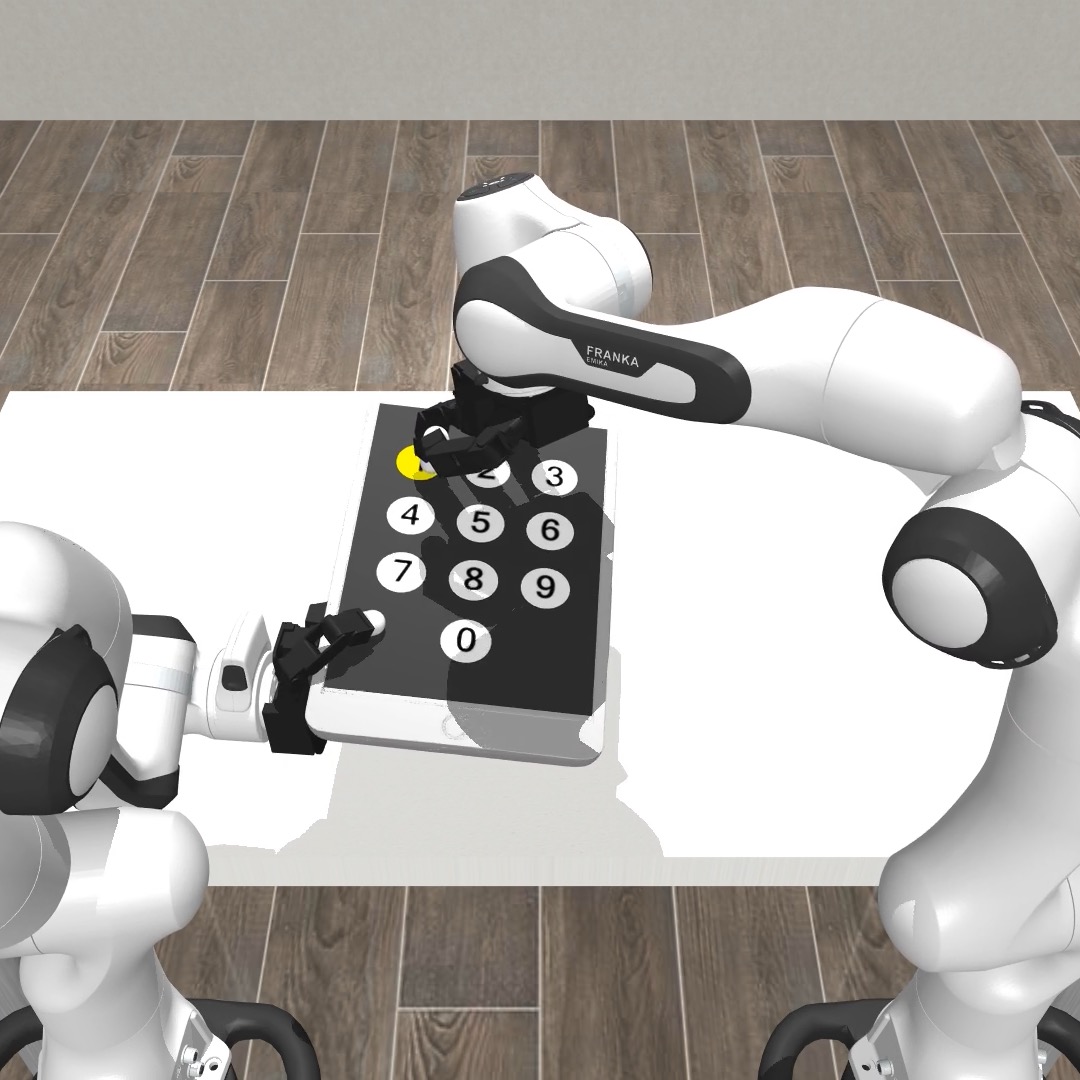} &
		\includegraphics[width=\linewidth]{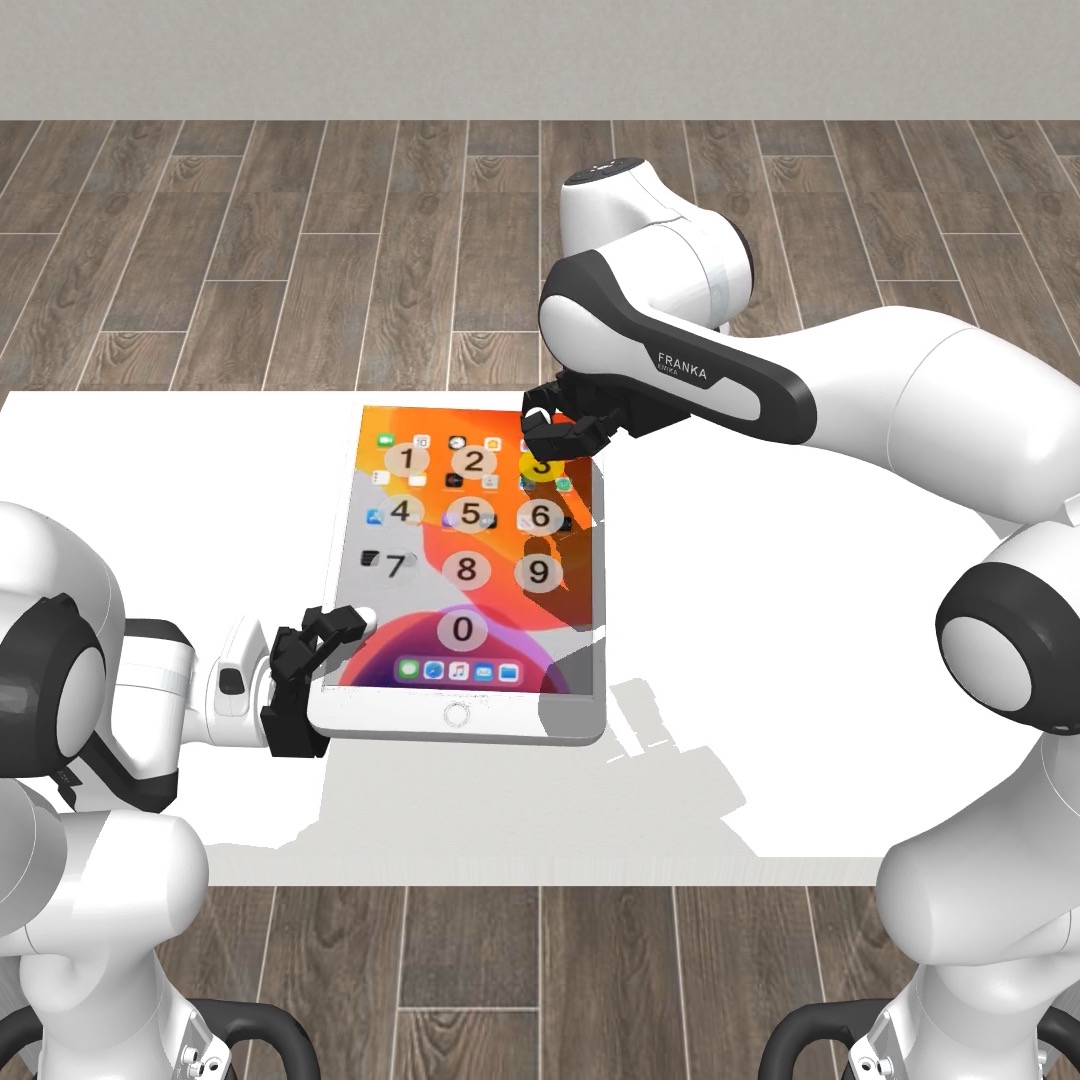} &
		\includegraphics[width=\linewidth]{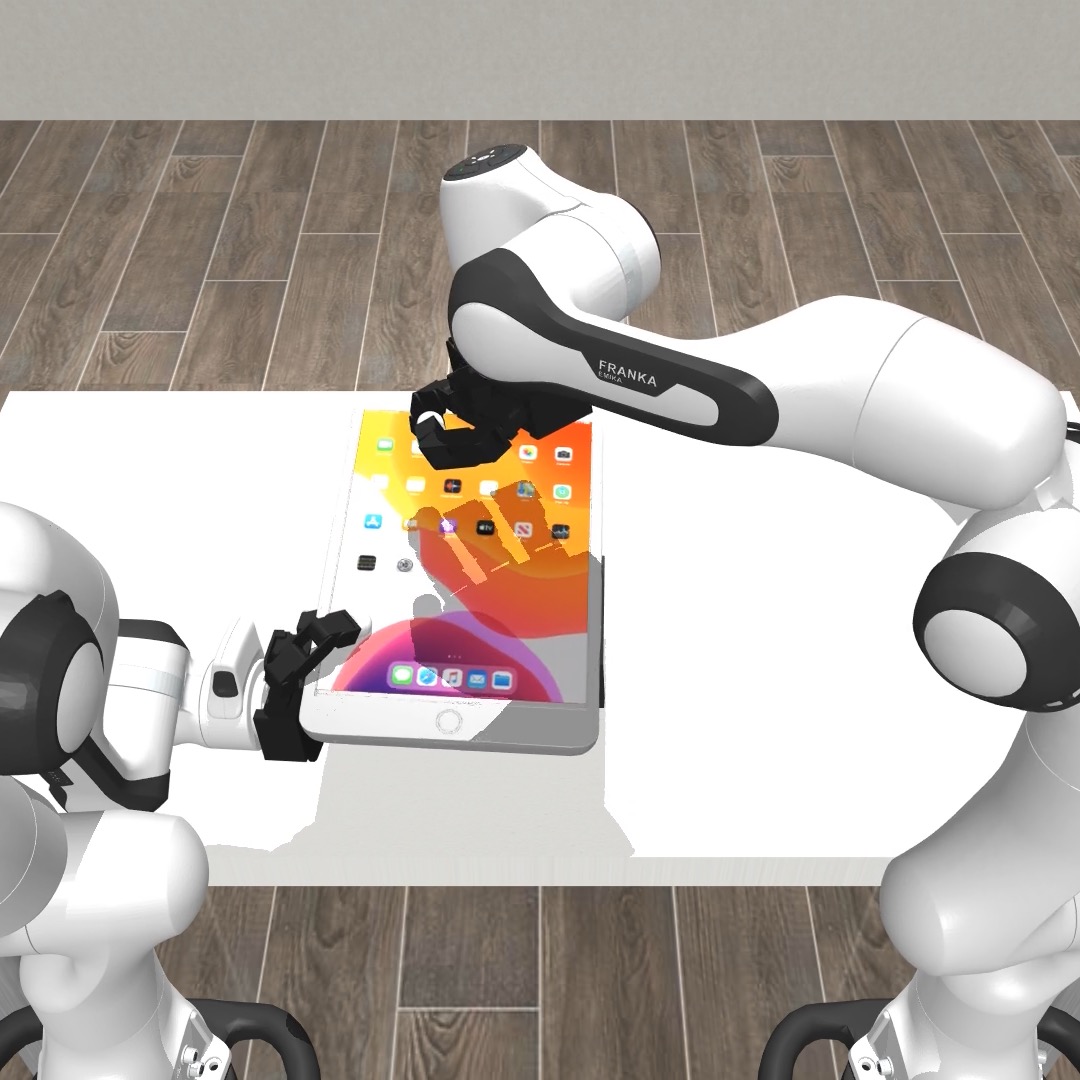} &
		Grasp the iPad and enter the password 123 to unlock the device.\\

		Hanoi /B &
		\includegraphics[width=\linewidth]{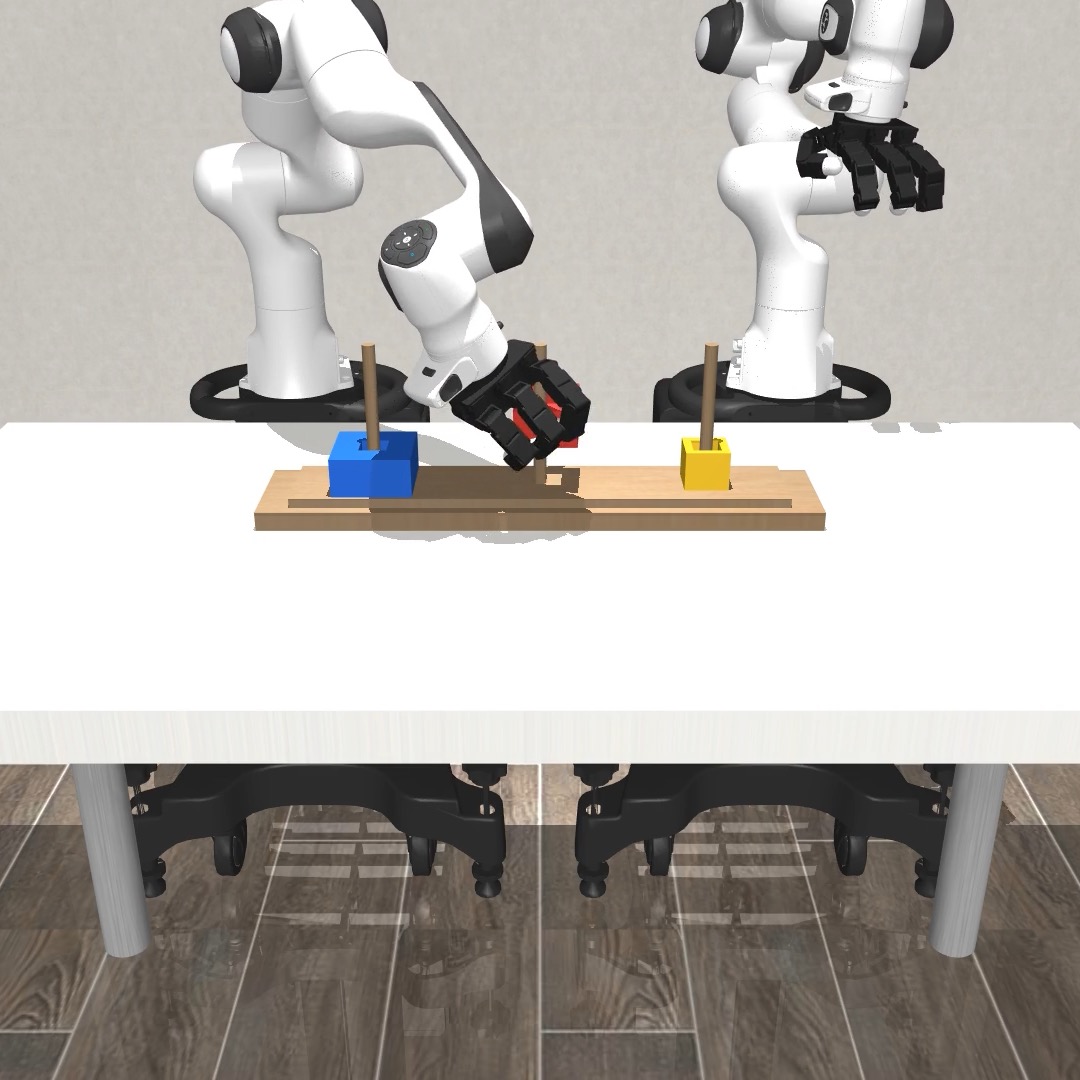} &
		\includegraphics[width=\linewidth]{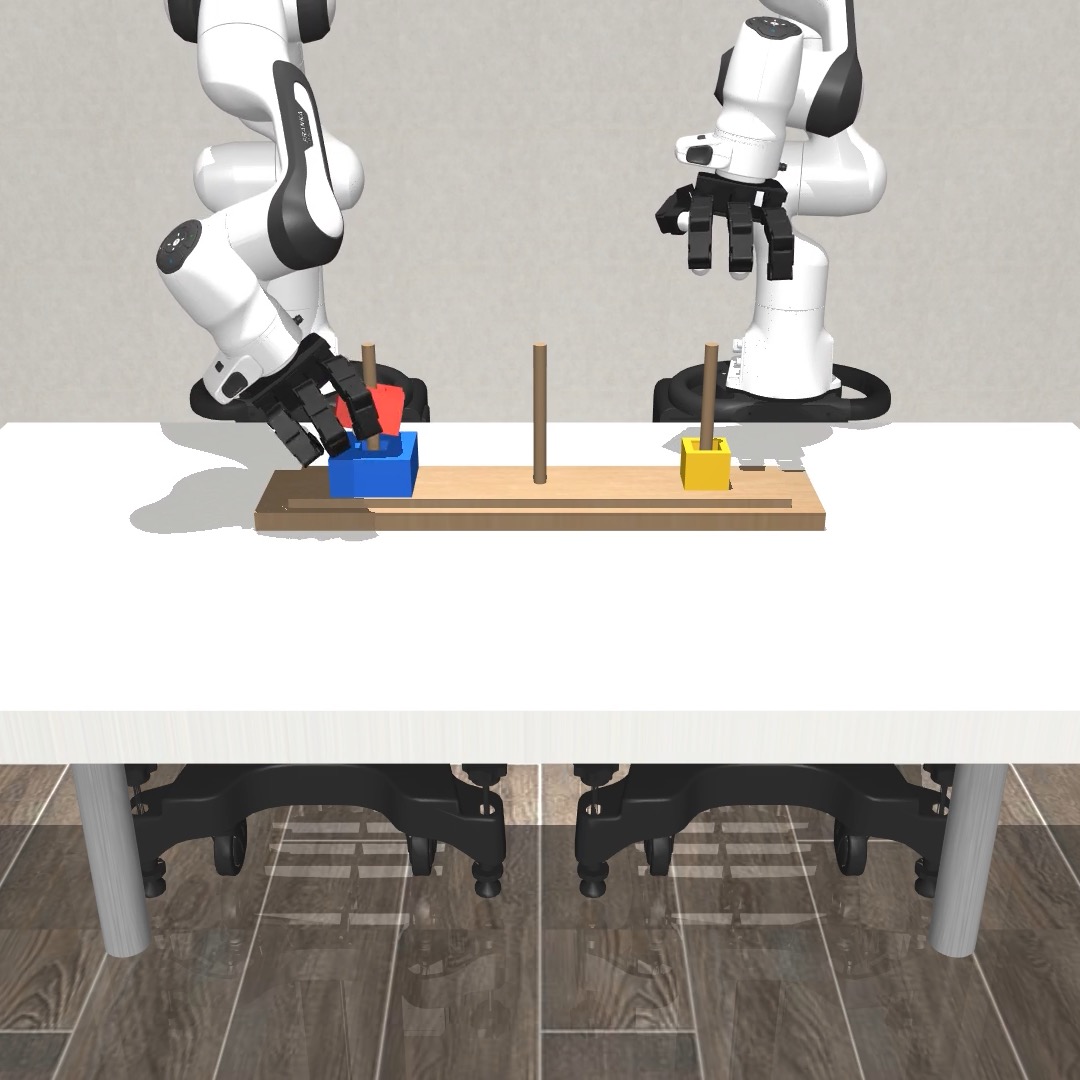} &
		\includegraphics[width=\linewidth]{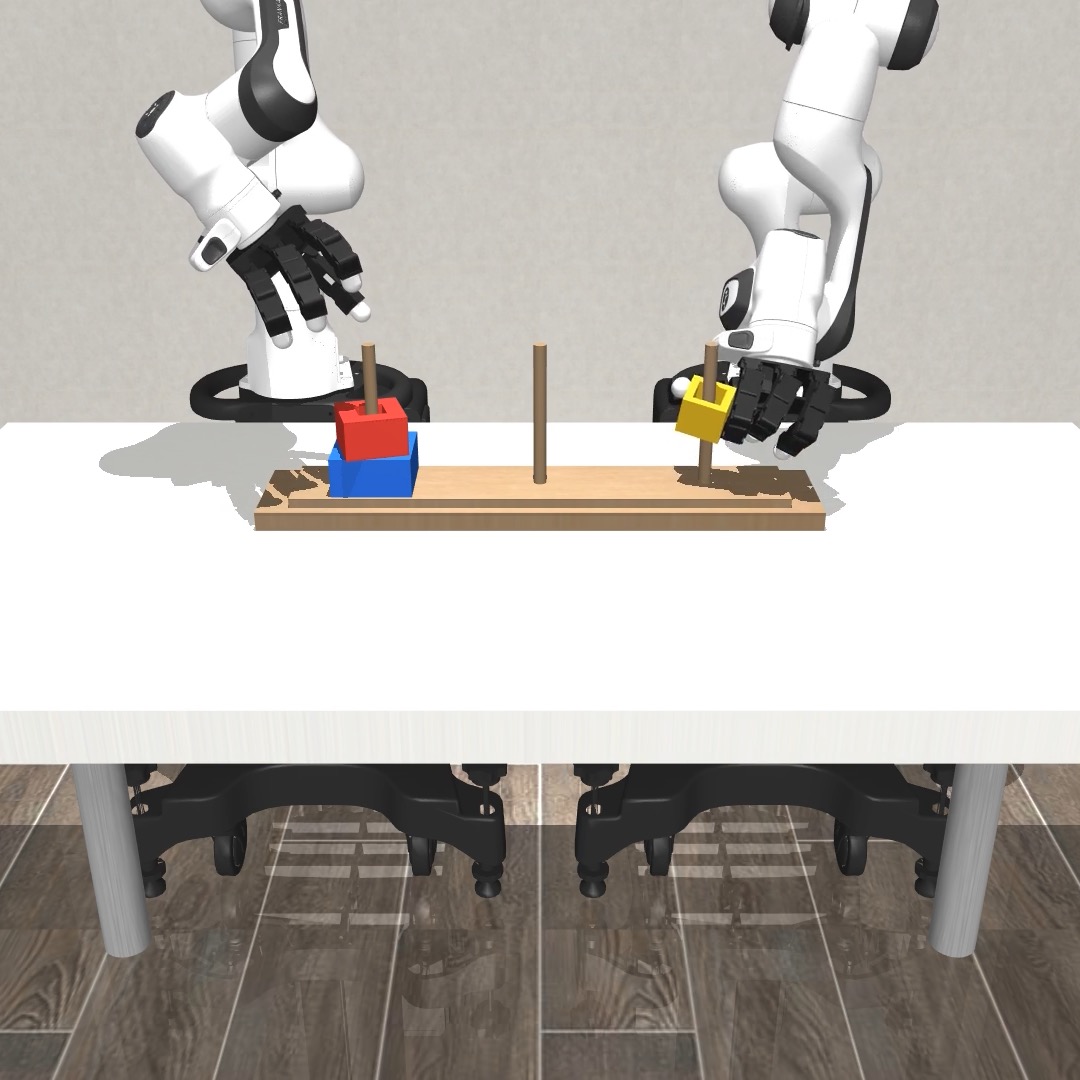} &
		\includegraphics[width=\linewidth]{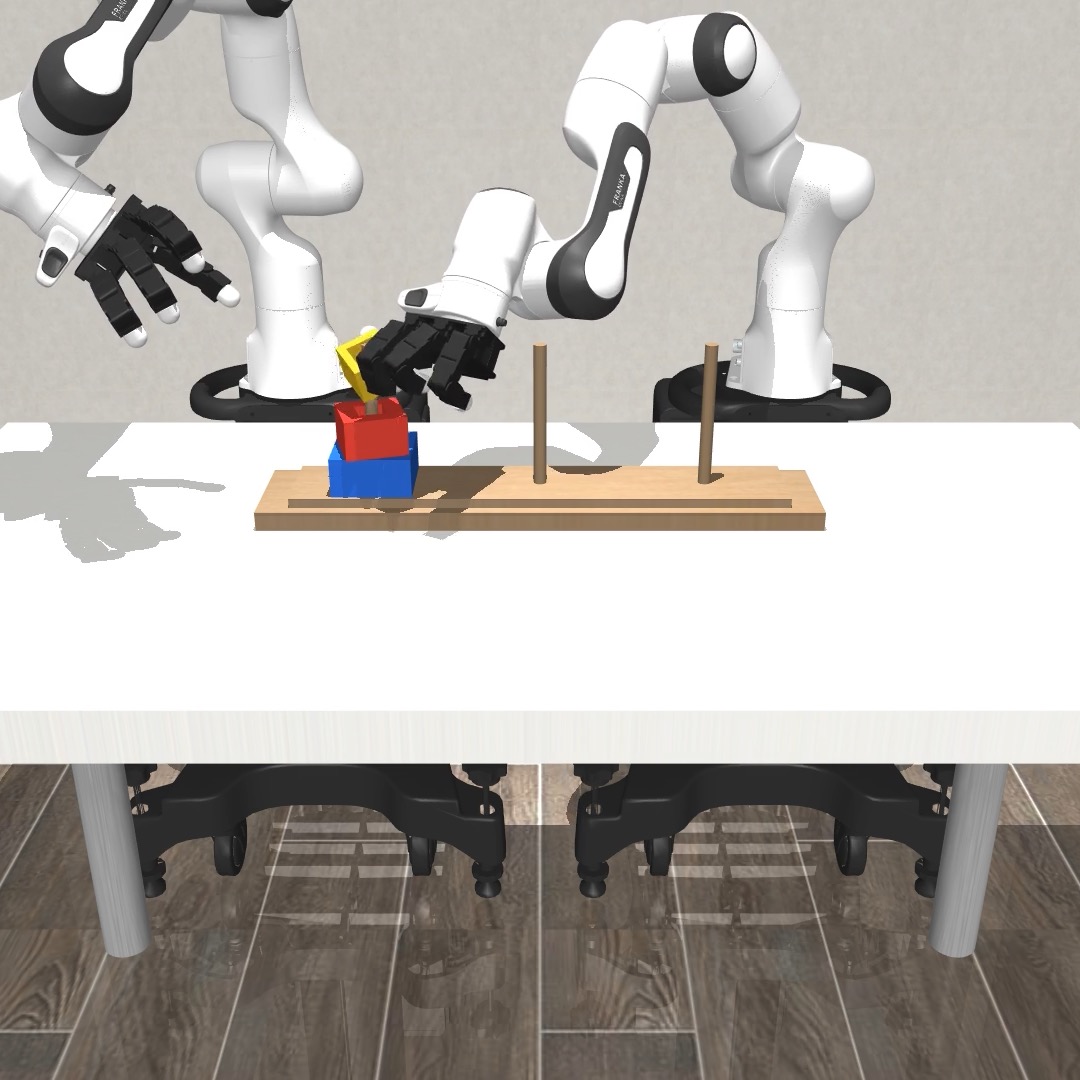} &
		Execute the final two moves of the three-level Tower of Hanoi: move the medium disk from the middle peg to the right peg with the right hand, then move the small disk from the left peg to the right peg with the left hand.\\

		Assembly /B &
		\includegraphics[width=\linewidth]{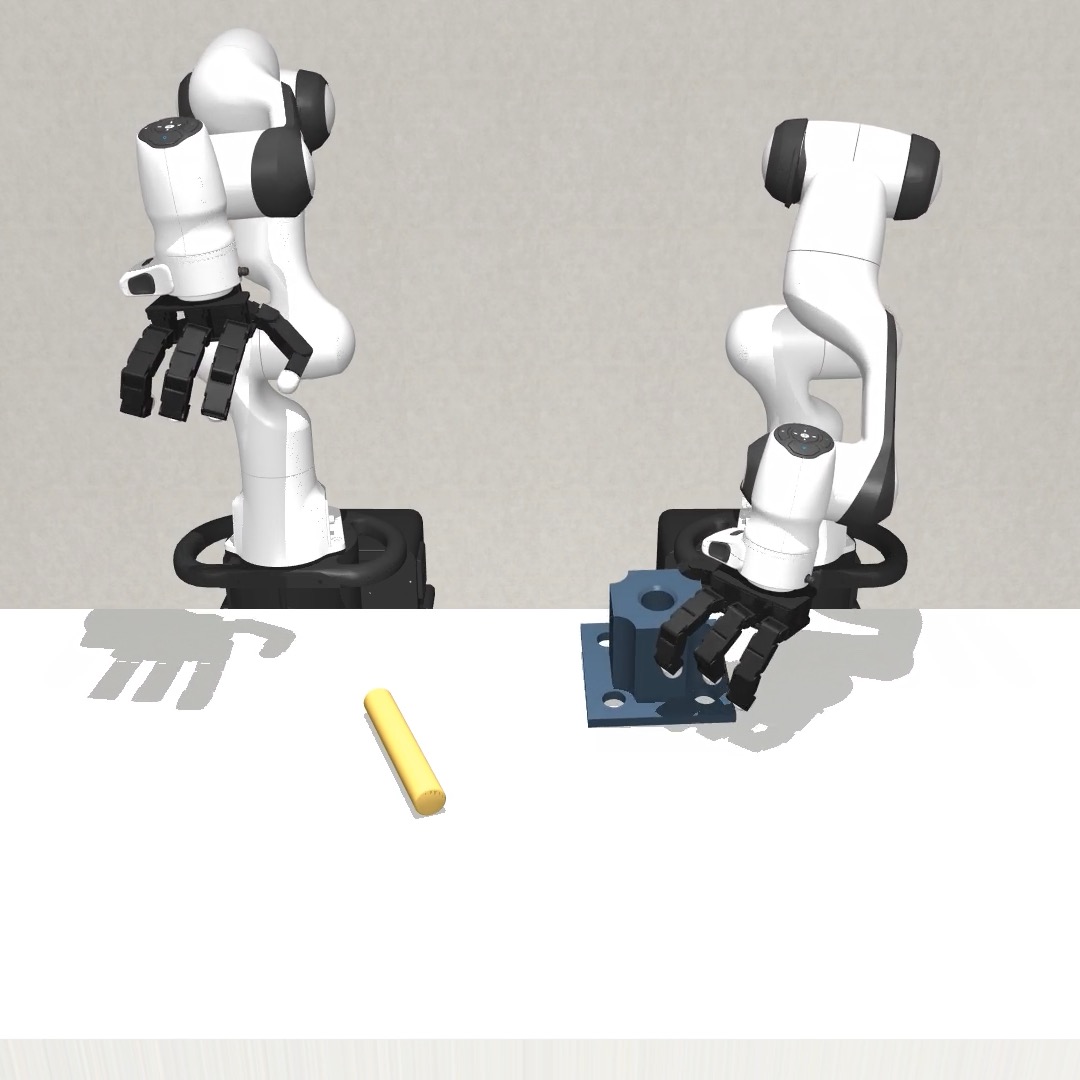} &
		\includegraphics[width=\linewidth]{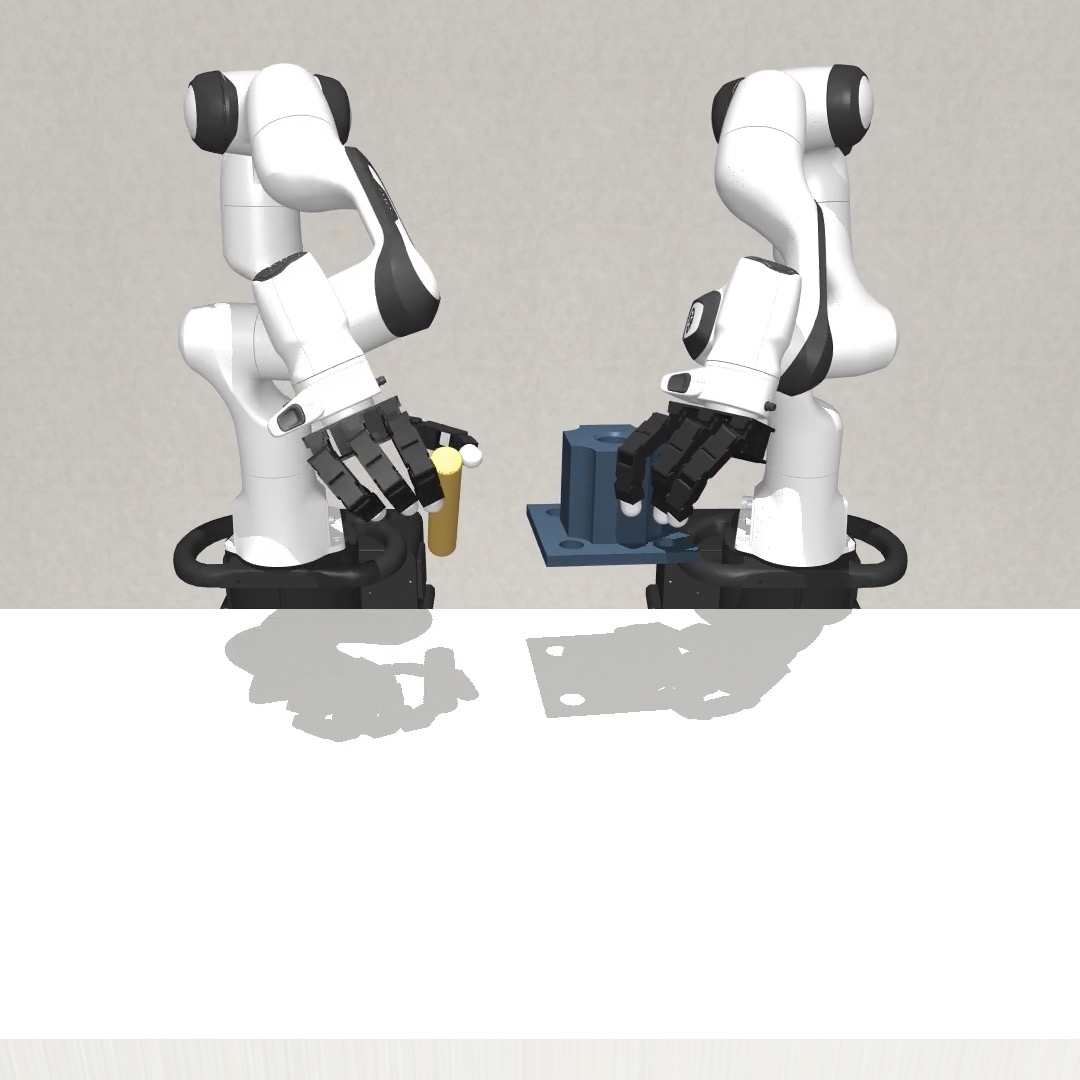} &
		\includegraphics[width=\linewidth]{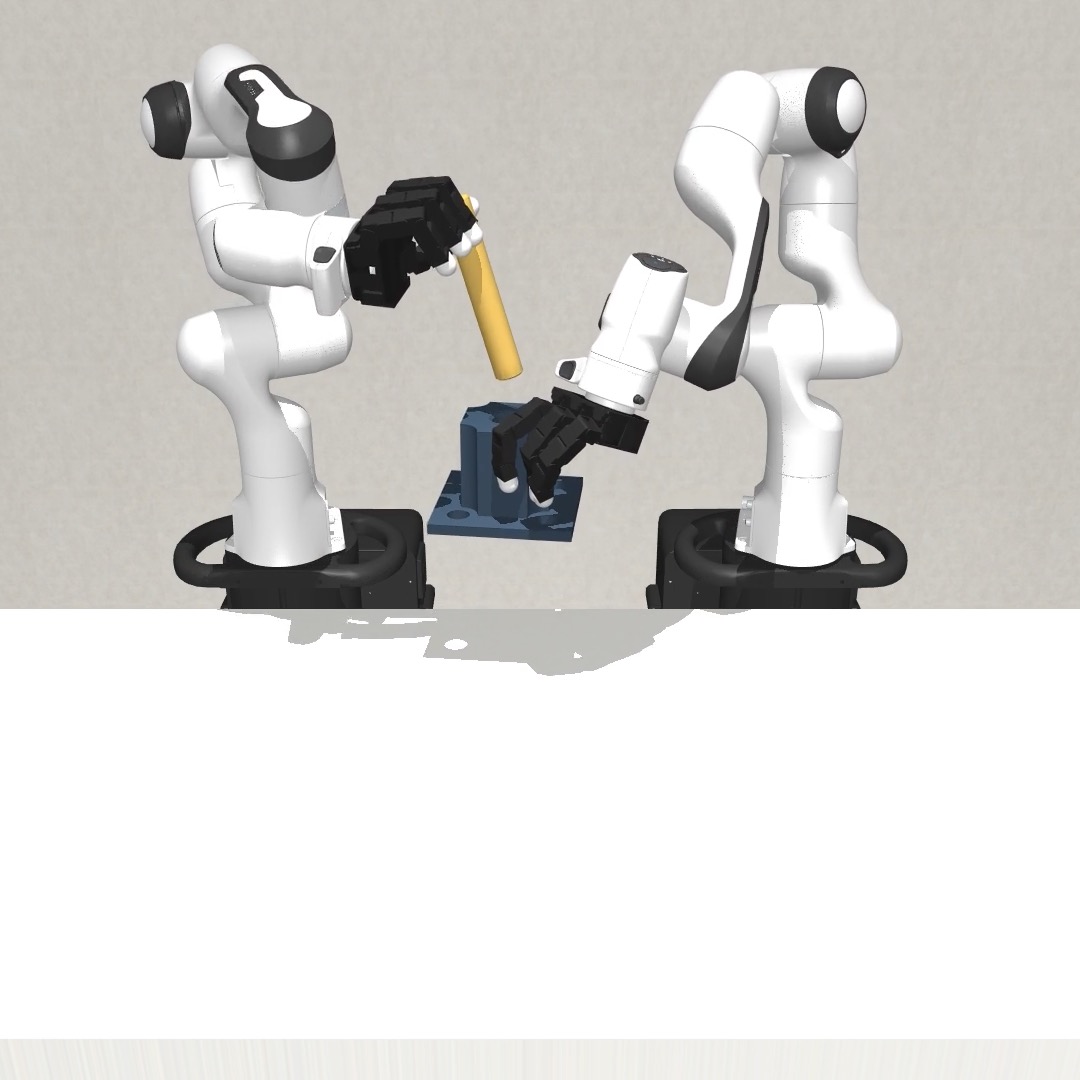} &
		\includegraphics[width=\linewidth]{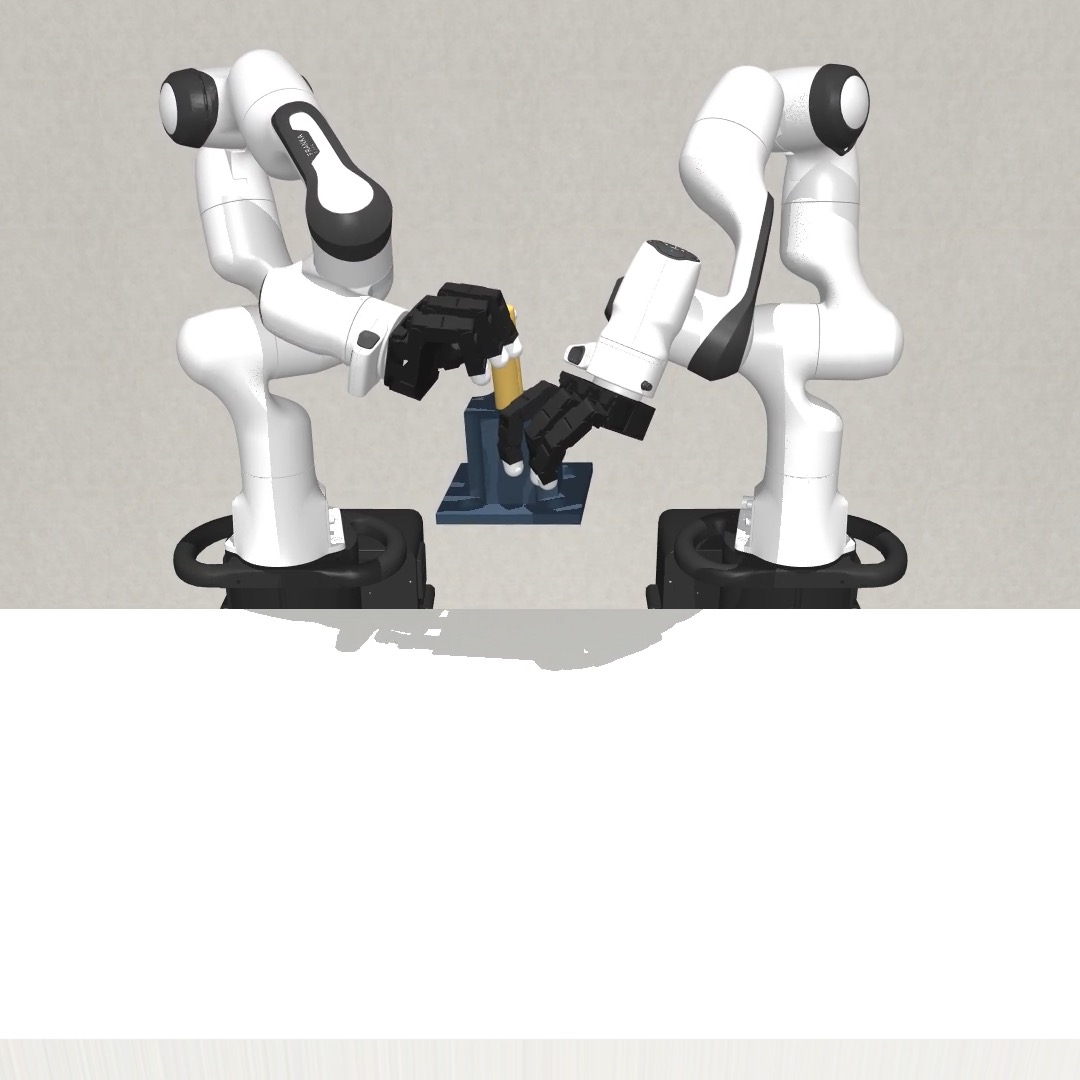} &
		Grasp the tray with the left hand and the peg with the right hand, then insert the peg into the hole. \\

		Microwave /B &
		\includegraphics[width=\linewidth]{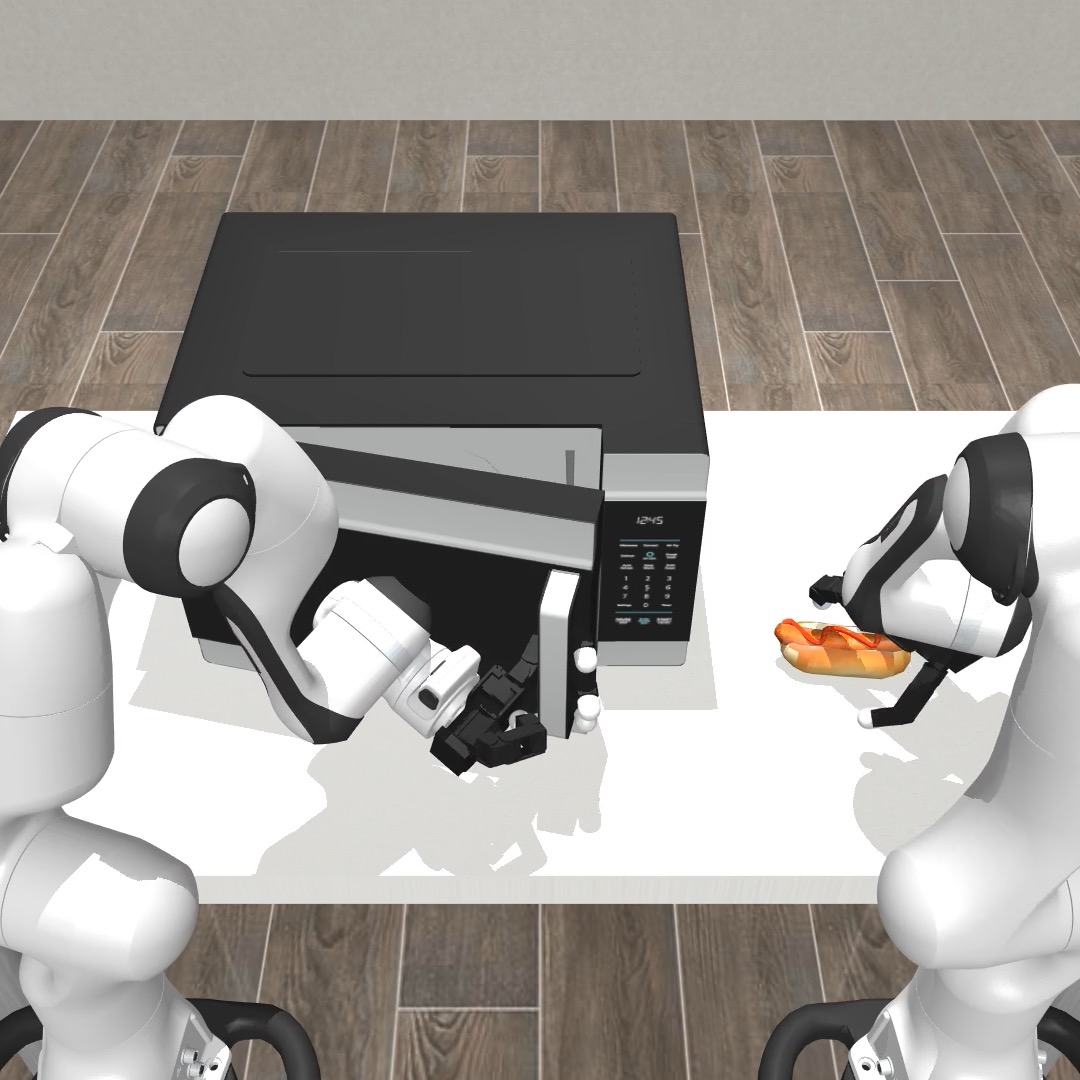} &
		\includegraphics[width=\linewidth]{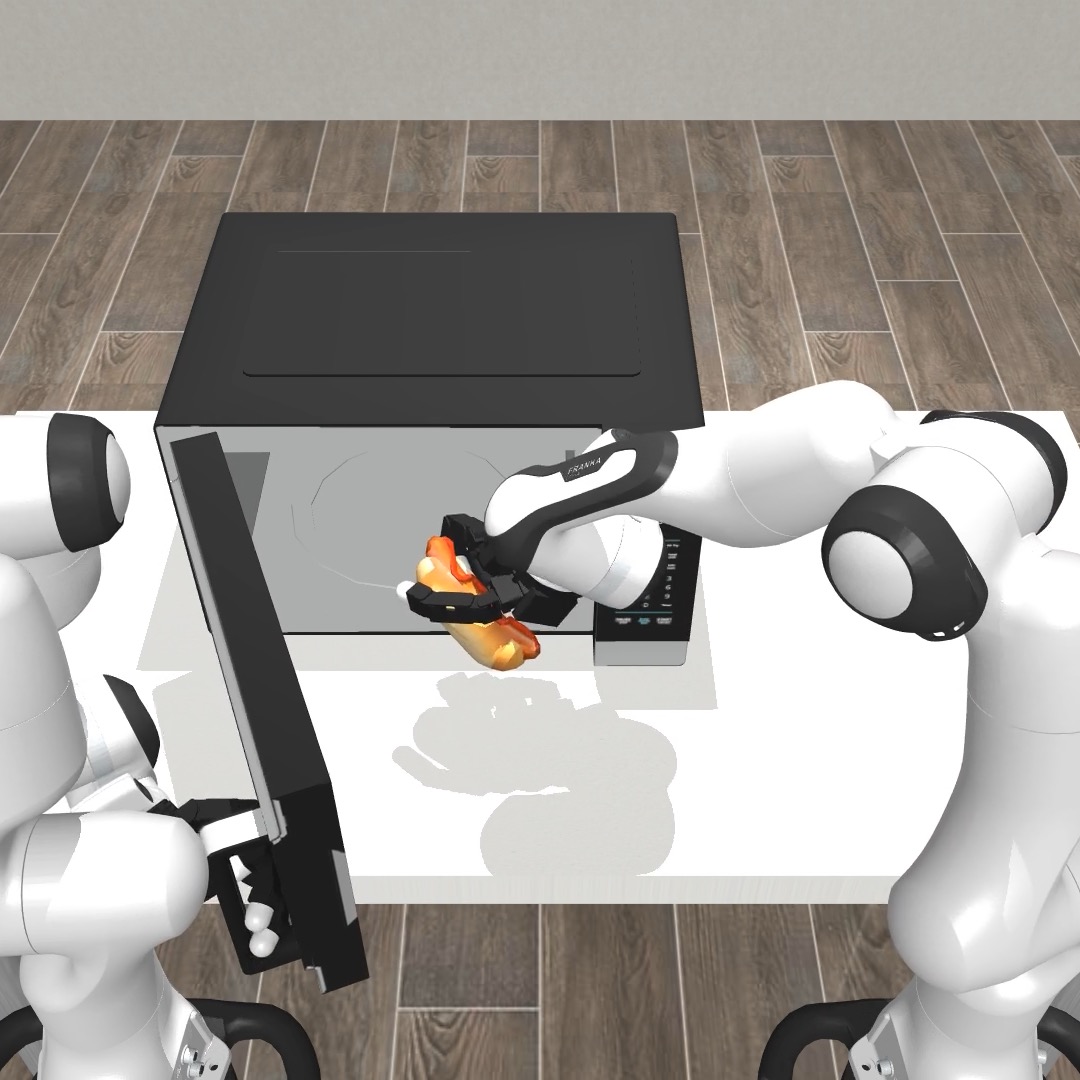} &
		\includegraphics[width=\linewidth]{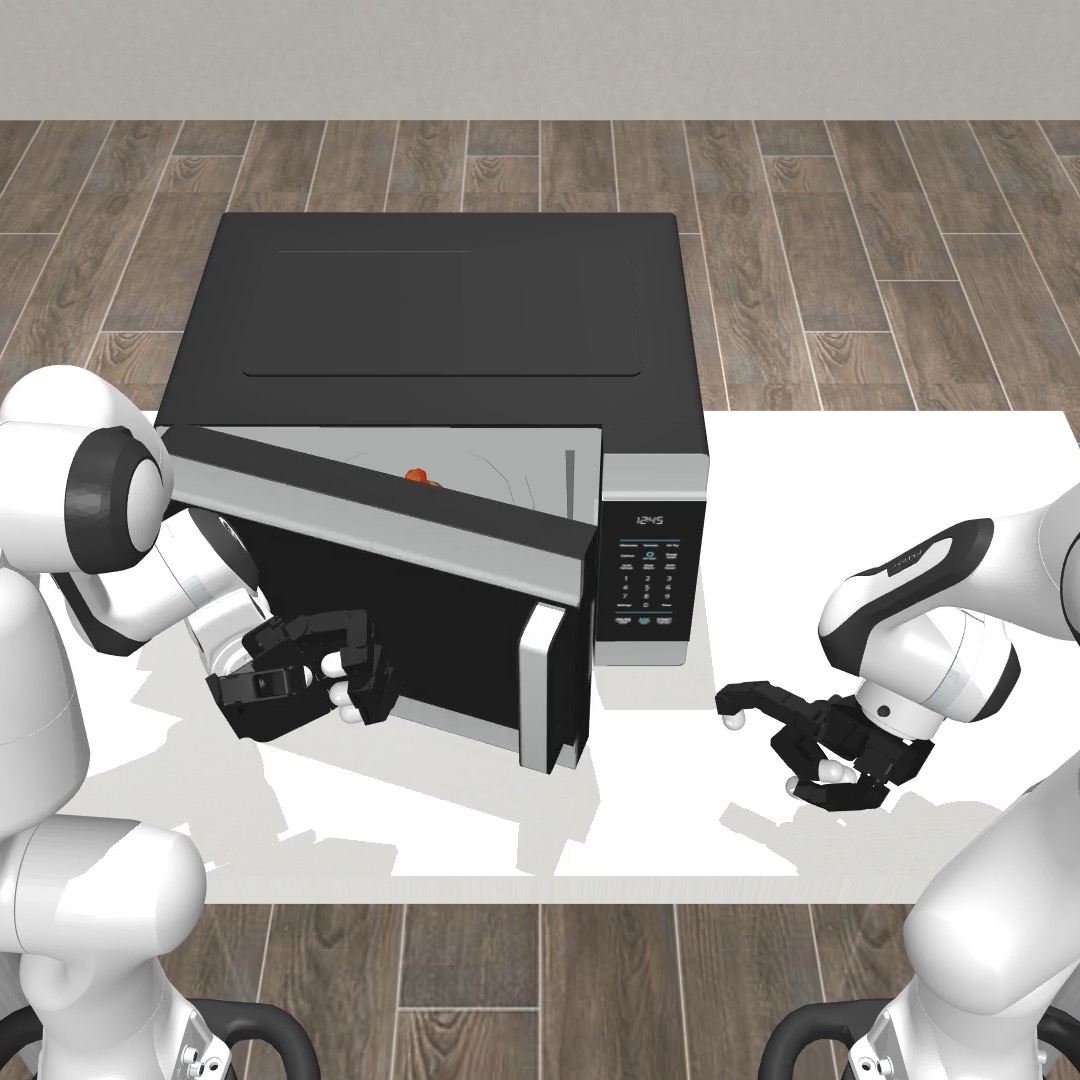} &
		\includegraphics[width=\linewidth]{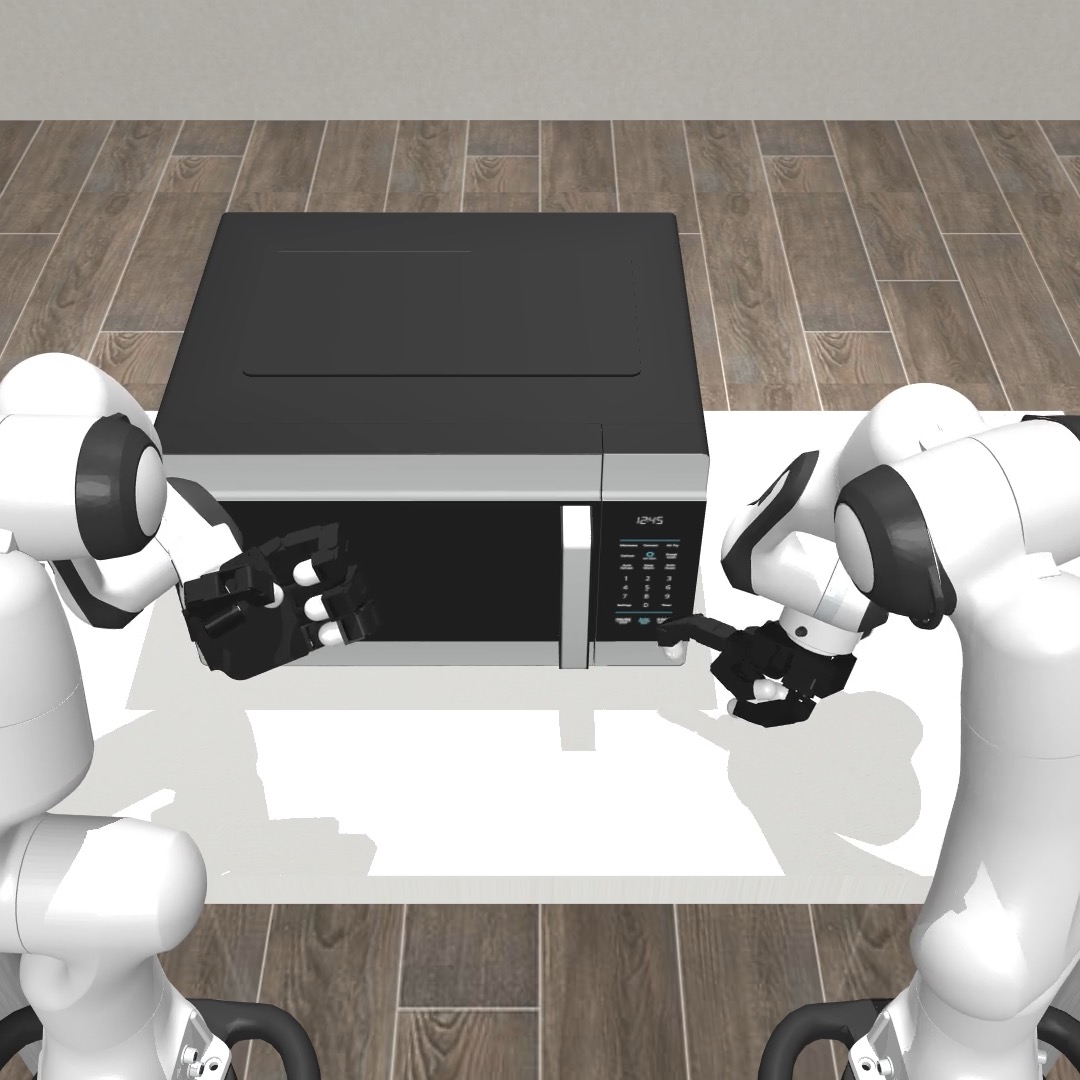} &
		Open the microwave door, place the food inside the microwave, close the door, and press the start button.\\

		Photograph /B &
		\includegraphics[width=\linewidth]{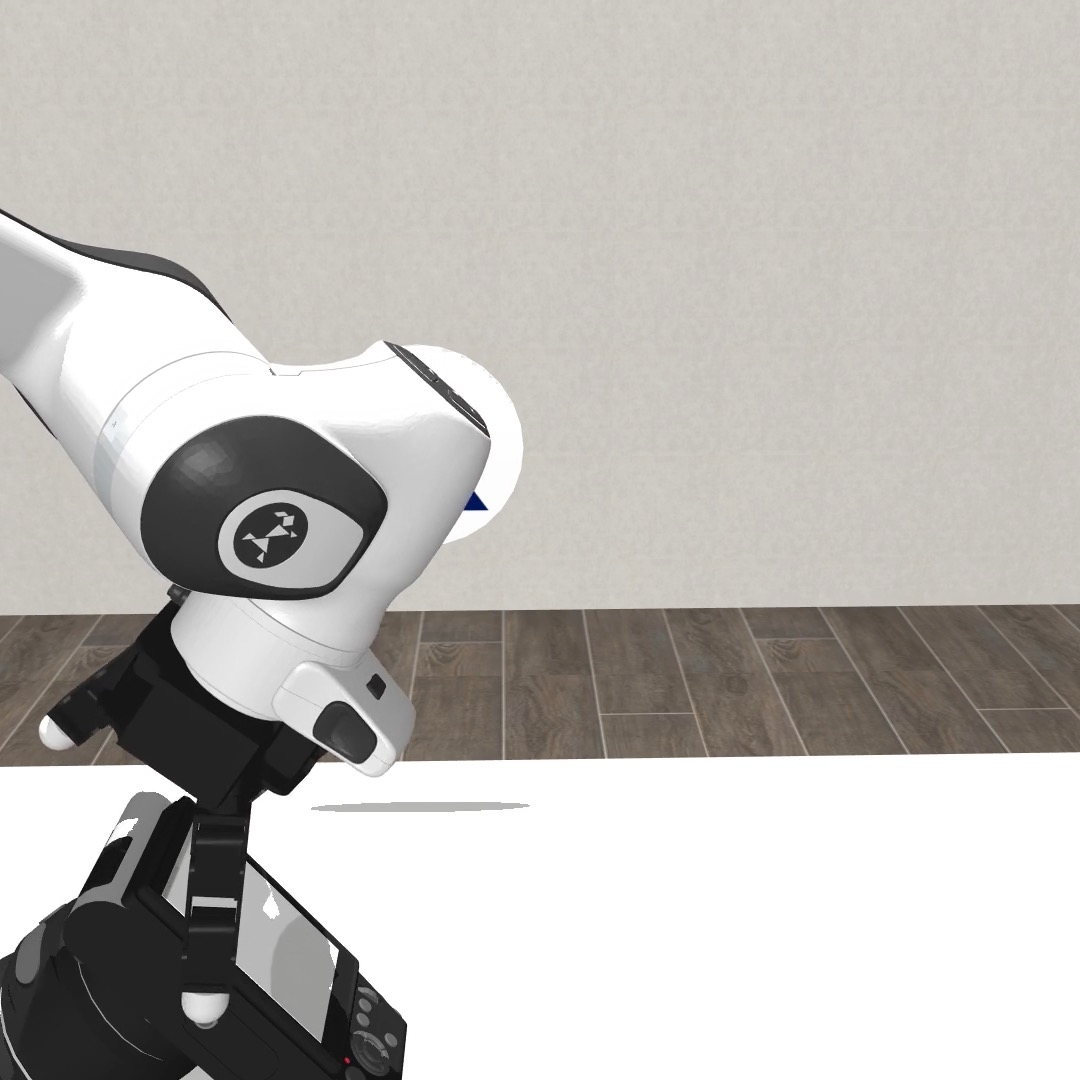} &
		\includegraphics[width=\linewidth]{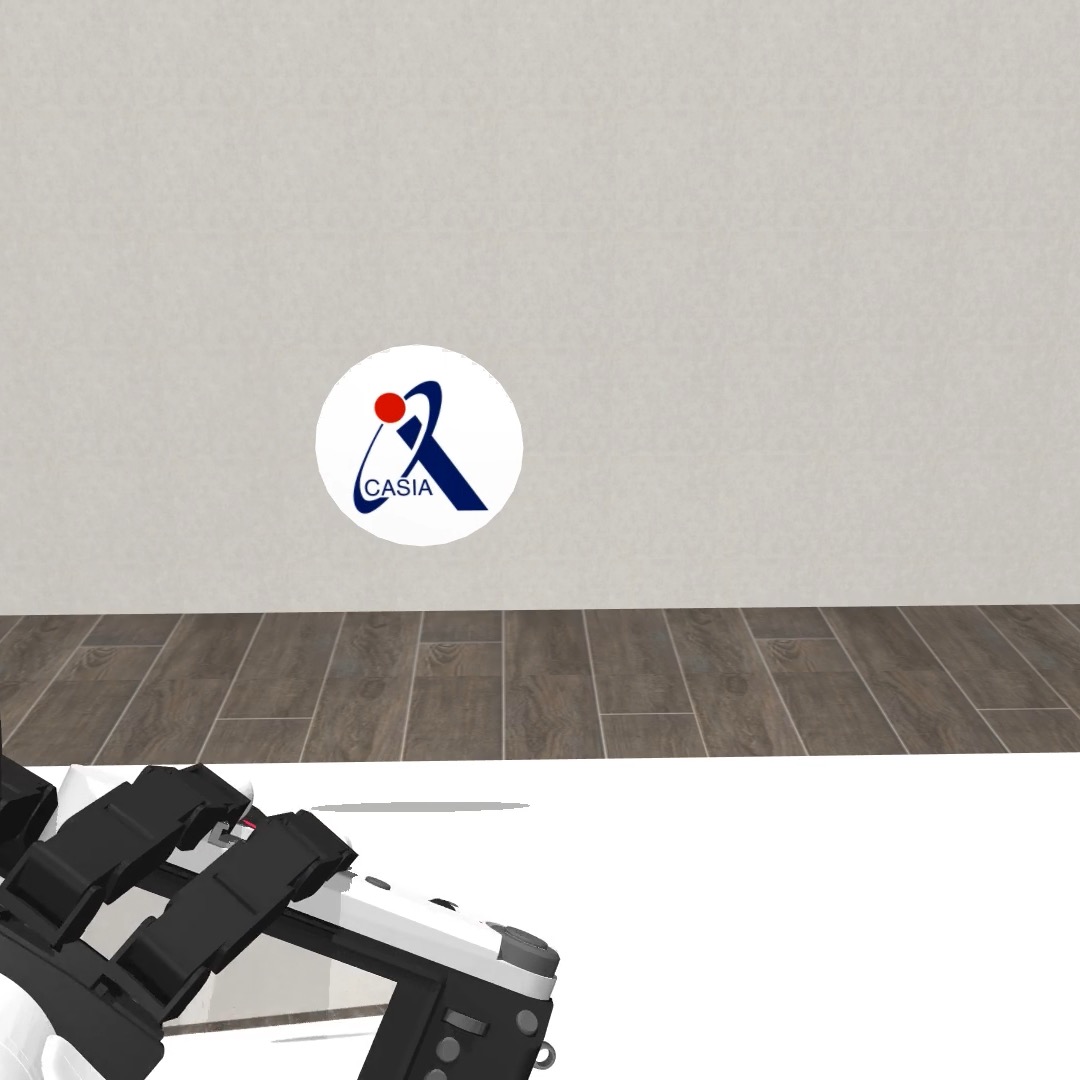} &
		\includegraphics[width=\linewidth]{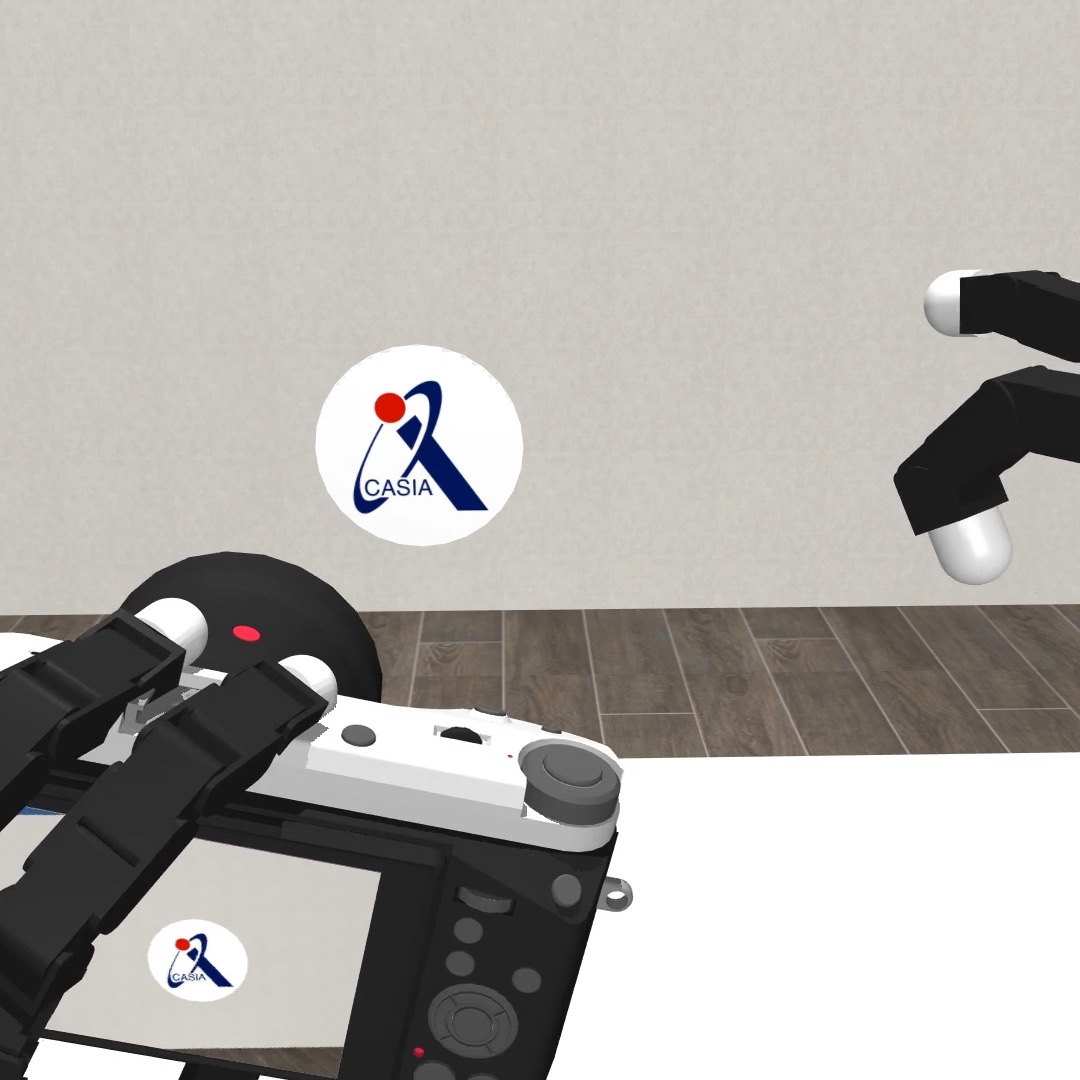} &
		\includegraphics[width=\linewidth]{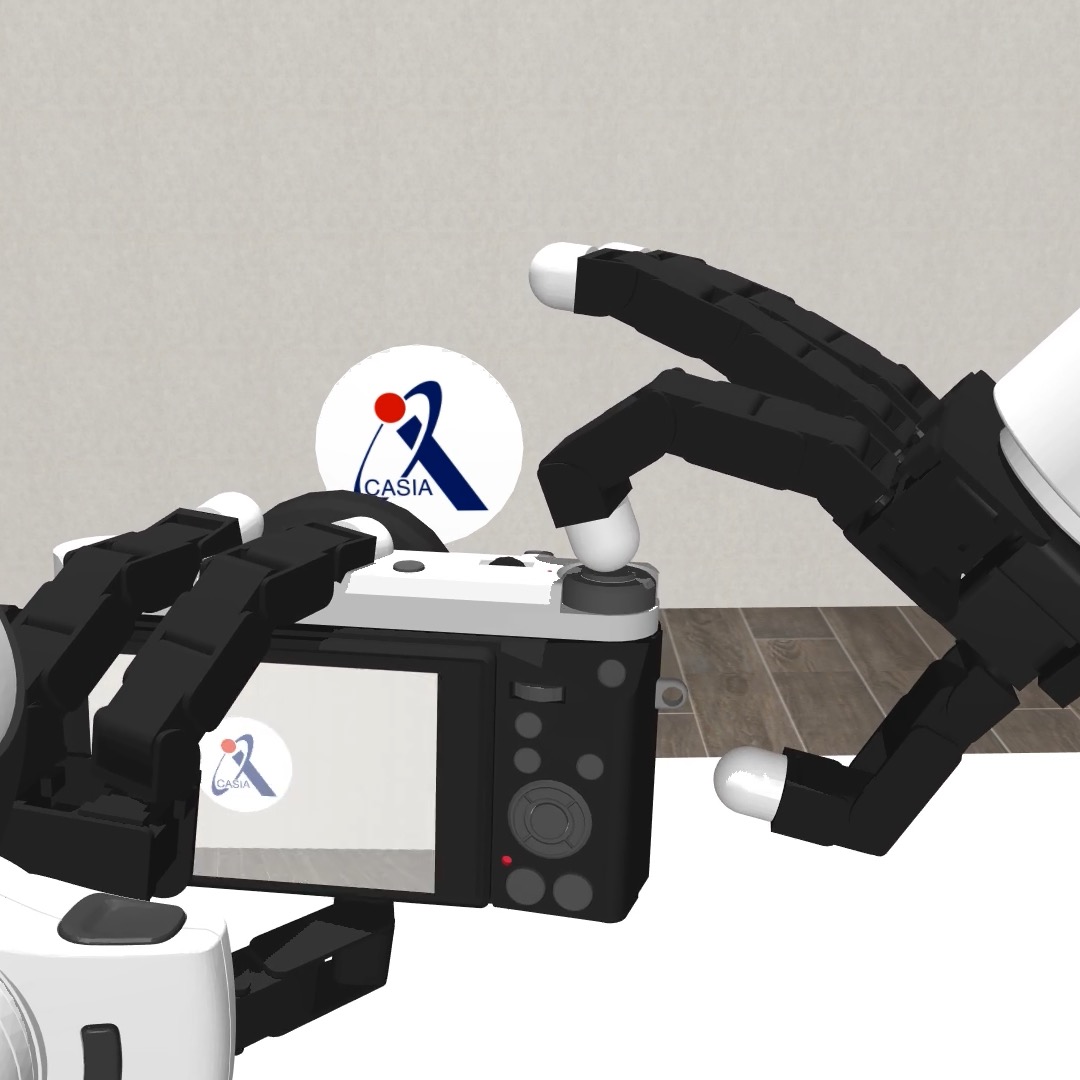} &
		Grasp the camera with the left hand, align it with the logo, and press the shutter button with the right hand.\\
	\end{longtable}
}
\twocolumn

\end{document}